\DocumentMetadata{}  
\documentclass[acmlarge,screen,8pt]{acmart} 

\AtBeginDocument{%
  \providecommand\BibTeX{{%
    \normalfont B\kern-0.5em{\scshape i\kern-0.25em b}\kern-0.8em\TeX}}}

\setcopyright{acmcopyright}
\acmYear{2023}
\acmDOI{XXXXXXX.XXXXXXX}

\usepackage{multirow}
\usepackage{booktabs}
\usepackage{wrapfig}
\usepackage{amsthm}
\usepackage{xcolor}
\usepackage{hyperref} 

\begin{document}

\newcommand{\lgg}[1]{\textcolor[rgb]{0.04, 0.04, 0.98}{\textbf{#1}}}

\newcommand{\edited}[1]{\textcolor[rgb]{0.04, 0.04, 0.98}{#1}}

\title{AI-Generated Content (AIGC) for Various Data Modalities: A Survey}


\author{Lin Geng Foo}
\affiliation{%
  \institution{Singapore University of Technology and Design}
  \country{Singapore}
  \postcode{487372}
}

\author{Hossein Rahmani}
\affiliation{%
  \institution{Lancaster University}
  \city{Lancaster}
  \country{United Kingdom}
  \postcode{LA1 4YW}  
}


\author{Jun Liu}
\authornote{Corresponding author.}
\affiliation{%
  \institution{Lancaster University}
  \city{Lancaster}
  \country{United Kingdom}
  \postcode{LA1 4YW}  
}

\renewcommand{\shortauthors}{Foo, et al.}

\begin{abstract}
AI-generated content (AIGC) methods aim to produce text, images, videos, 3D assets, and other media using AI algorithms. 
Due to its wide range of applications and the potential of recent works, AIGC developments -- especially in Machine Learning (ML) and Deep Learning (DL) -- have been attracting significant attention, and this survey focuses on comprehensively reviewing such advancements in ML/DL. AIGC methods have been developed for various data modalities, such as image, video, text, 3D shape, 3D scene, 3D human avatar, 3D motion, and audio -- each presenting unique characteristics and challenges.
Furthermore, there have been significant developments in cross-modality AIGC methods, where generative methods receive conditioning input in one modality and produce outputs in another. Examples include going from various modalities to image, video, 3D, and audio. This paper provides a comprehensive review of AIGC methods across different data modalities, including both single-modality and cross-modality methods, highlighting the various challenges, representative works, and recent technical directions in each setting. 
We also survey the representative datasets throughout the modalities, and present comparative results for various modalities. 
Moreover, we discuss the typical applications of AIGC methods in various domains, challenges, and future research directions.
\end{abstract}




\maketitle

\section{Introduction}\label{sec:introduction}

Amidst the rapid advancement of artificial intelligence (AI), the development of content generation techniques stands out as one of the most captivating and widely discussed topic in the field.
AI-generated content (AIGC) encompasses the production of text, images, videos, 3D assets, and other media through AI algorithms, which enables automation of the content creation process.
Despite the broad term of ``AI'' used in ``AIGC'', the focus and progress of the AIGC field in recent years lies mainly on Machine Learning (ML) and Deep Learning (DL) techniques, which is also the focus in this survey. Such automation facilitated by AIGC significantly
{{\parfillskip500pt\par}}
\begin{wrapfigure}{r}{0.45\textwidth}
\vspace{-8mm}
    \center
    \tiny
    \includegraphics[width=0.44\textwidth]{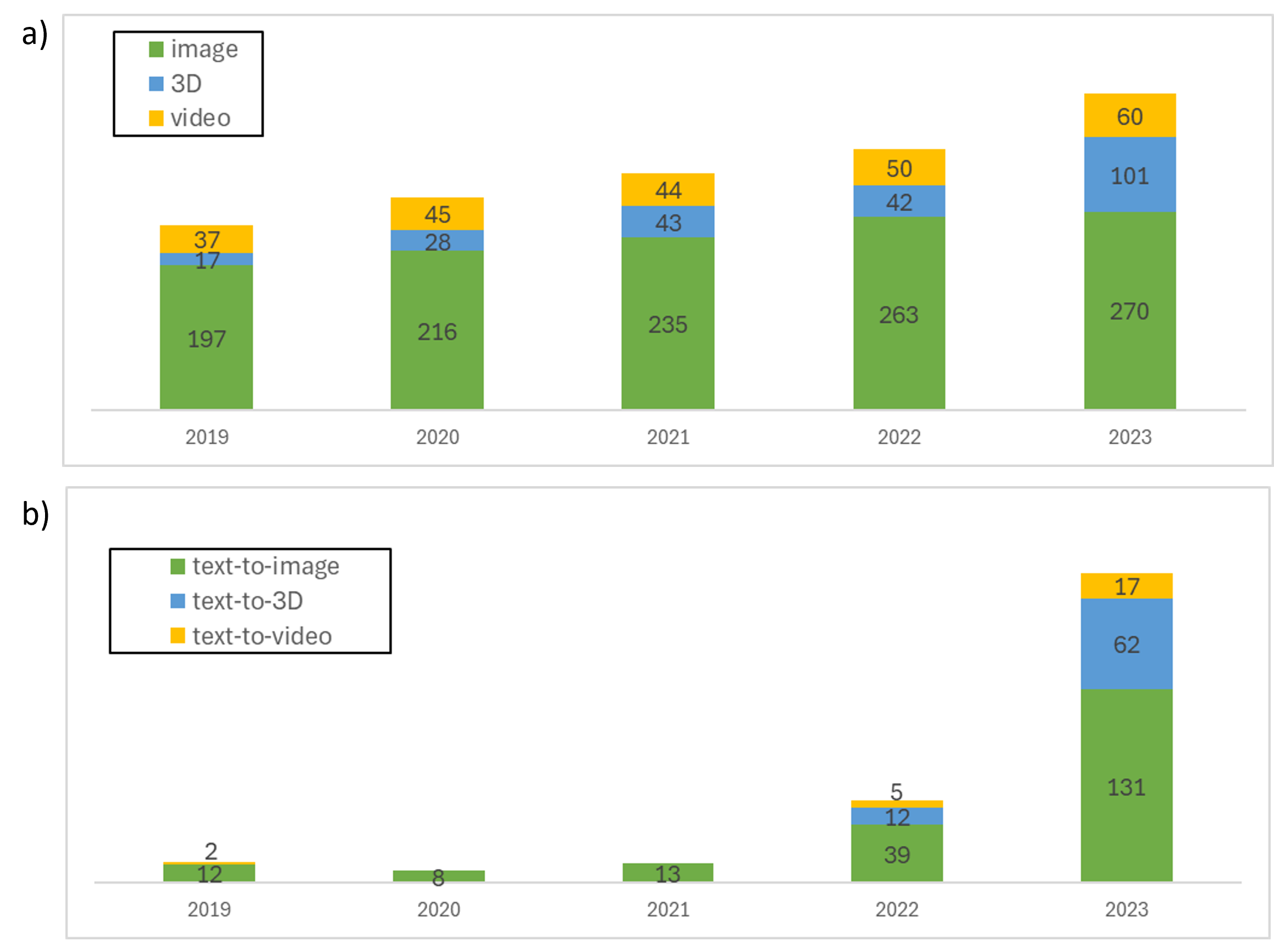}  
    \vspace{-2.5mm}
    \caption{
    General trend of the number of papers published regarding a) single-modality and b) cross-modality generation/editing every year over the past five years in six top CV and ML conferences (CVPR, ICCV, ECCV, NeurIPS, ICML, and ICLR), as well as three related AI conferences (AAAI, IJCAI, ECAI). 
    a) There is an increasing trend over the years for AIGC papers published regarding image, video and 3D (shape, human and scene) generation for a single modality.     
    b) There is an observable spike over the last 2 years for papers published regarding cross-modality (text-to-image, text-to-video and text-to-3D) generation. 
    }
    \label{figure:paper_trends}
\vspace{-4mm}    
\end{wrapfigure}
\noindent
reduces the need for human labor and lowers the costs of content creation,  fundamentally reshaping industries such as advertising \cite{guo2021vinci}, education \cite{baidoo2023education}, code development \cite{chen2021evaluating}, and media/entertainment \cite{park2019generating}.
These generative methods across the various modalities have an incredible variety of applications. For instance, text generation models have driven the development of powerful chatbots (e.g., ChatGPT), which have found widespread applications across domains such as education \cite{baidoo2023education}, gaming \cite{lanzi2023chatgpt}, and code development \cite{chen2021evaluating}. Meanwhile, image and video generation models can be applied to industries such as media/entertainment \cite{park2019generating}, advertising \cite{guo2021vinci}, and film production \cite{singh2023artificial}, while generation of 3D modalities can be adopted in AR/VR applications (e.g., metaverse) \cite{ratican2023proposed}, e-commerce and fashion \cite{gong2024laga}, as well as architecture and interior design \cite{gao2023scenehgn}.

In the earlier days of AIGC, developments and approaches mainly involved a single modality, where the inputs (if any) and outputs of the generation model both share the same modality.
The seminal work \cite{goodfellow2014generative} by Goodfellow et al. first introduced Generative Adversarial Networks (GANs) that were in principle capable of training deep neural networks to generate images that were difficult to be distinguished from images in the training dataset.
This demonstration of generative capabilities by deep neural networks led to extensive developments on single-modality generation for images \cite{goodfellow2014generative,kingma2014auto,ho2020denoising}, as well as various other modalities, such as videos \cite{vondrick2016generating,tulyakov2018mocogan}, text \cite{radford2018improving,radford2019language,brown2020language}, 3D shapes (as voxels \cite{wu20153d,riegler2017octnet}, point clouds \cite{achlioptas2018learning,yang2019pointflow}, meshes \cite{sinha2017surfnet,vahdat2022lion} and neural implicit fields \cite{chen2019learning,park2019deepsdf}), 3D scenes \cite{nguyen2019hologan,mildenhall2020nerf}, 3D avatars (full bodies \cite{chen2022gdna,zhang2022avatargen} and heads \cite{yenamandra2021i3dmm,hong2022headnerf}), 3D motions \cite{holden2016deep,raab2023modi}, audio \cite{oord2016wavenet,kalchbrenner2018efficient}, and so on.
Moreover, such developments have continued consistently over the years, where the number of works published in the field every year has been steadily increasing over the past years, as shown in Figure \ref{figure:paper_trends}(a).
Although generative models for each modality share some similar approaches and principles, they also encounter unique challenges.
Consequently, the methods and designs for generative models of each modality are specially dedicated to address these distinct challenges.

Recently, there has also been a rapid development of AIGC involving multiple modalities, where the input and output modalities differ.
This cross-combination of modalities grants users greater control over what outputs to generate, e.g., generating desired images with input text descriptions, or generating a personalized 3D human avatar from RGB images or videos.
However, such cross-modality methods are generally more challenging to train as there can be a large gap between the representations of different modalities.
Moreover, they often require larger datasets that include paired data from multiple modalities in order to effectively capture the diverse relationships between them.
Notably, recent works such as Stable Diffusion \cite{rombach2022high}, Make-A-Video \cite{singer2023make}, and DreamFusion \cite{poole2023dreamfusion} have further demonstrated the impressive capabilities of AIGC in receiving text prompts and delivering remarkable outputs in various modalities that can rival human craftsmanship, which have inspired a large increase in the number of works in the field, as shown in Figure \ref{figure:paper_trends}(b).
These recent advancements exhibit the potential of AIGC in various modalities, while simultaneously also open new and exciting avenues for cross-modality content generation.

\begin{figure*}[t]
    \center
    \includegraphics[width=0.97\textwidth]{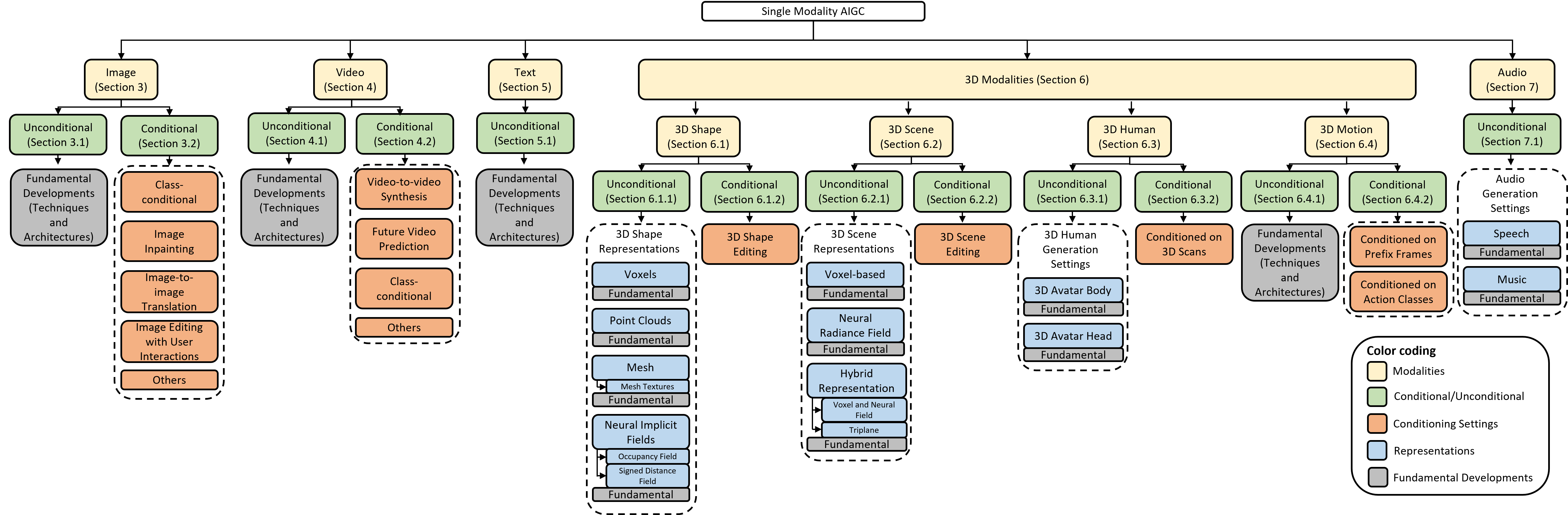}
    \vspace{-3mm}
    \caption{
    Taxonomy of single-modality AIGC methods in this survey. We organize the taxonomy according to the various generated modalities at the top level (in yellow). Then, each modality is further split into unconditional generation methods and conditional generation methods (in green). 
    Specifically, the unconditional methods we discuss are often the fundamental techniques and architectures (in grey) for generating each modality.
    For the discussion of certain modalities' unconditional methods (e.g., 3D modalities), we further categorize them according to the different representations or settings of the modality (in blue), before going in-depth into their respective fundamental techniques (in grey) for each representation or setting. To save space, we label them as ``Fundamental'' in the figure to represent ``Fundamental Developments (Techniques and Architectures)''. Furthermore, when discussing the conditional methods of each modality, we categorize them according to the different conditioning scenarios and settings (in orange).
    }    
    \label{figure:taxonomy_combined}
    \vspace{-7mm}
\end{figure*}

Therefore, recognizing the diverse nature of generative models across different modalities and the timely significance of cross-modality generation in light of recent advances, we review existing AIGC methods from this perspective.
Specifically, we comprehensively review the single-modality methods across a broad range of modalities, while also reviewing the latest cross-modality methods which lay the foundation for future works.
We discuss the challenges in each modality and setting, as well as the representative works and recent technical directions.
The main contributions are summarized as follows:

\begin{itemize}
    \item  To the best of our knowledge, this is the first survey paper that comprehensively reviews AIGC methods from the perspective of modalities, including: image, video, text, 3D (shape, scene, human, motion), and audio modality.    
    Since we focus on modalities, we further categorize the settings in each modality based on the input conditioning information.

    \item We comprehensively review the cross-modality AIGC methods, including cross-modality image, video, 3D (shape, scene, human, motion) and audio generation.    
    
    \item We focus on reviewing the most recent and advanced AIGC methods to provide readers with the state-of-the-art approaches.  
\end{itemize}

\begin{figure*}[t]
    \center
    \includegraphics[width=0.97\textwidth]{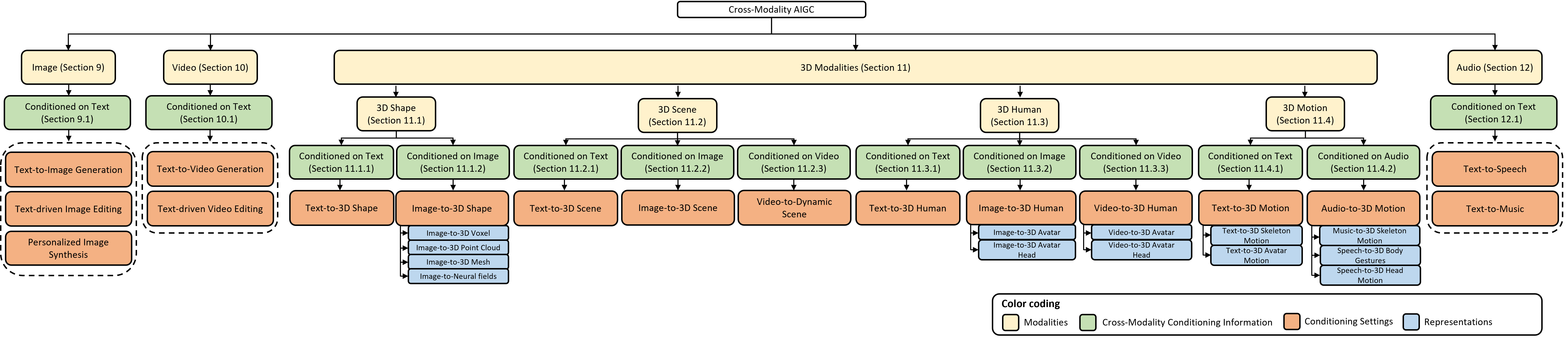} 
    \vspace{-4mm}
    \caption{
    Taxonomy of cross-modality AIGC methods. At the top level, we organize the taxonomy according to the various generated modalities (i.e., output modalities). Then, for each modality (in yellow), we categorize the works according to the modality of the input conditioning information (in green). Moreover, we further categorize them according to the different conditioning settings (in orange) and representations (in blue).    
    }
    \label{figure:taxonomy_combined2}
    \vspace{-5mm}
\end{figure*}

\noindent
\textbf{Paper Organization.}
The paper is organized as follows.
First, we compare our survey with existing works.
Next, we introduce the generative models involving a single modality, by introducing each modality separately.
We first present the image modality where we cover in-depth many of the foundational methods that are core to all the modalities, before presenting the other modalities.
Since this survey focuses on modalities, we further categorize the methods in each modality according to whether they are an unconditional approach (where no constraints are provided regarding what images to generate), or according to the type of conditioning information required.
Note that the unconditional approaches tend to be foundational generative techniques for the modality. 
On the other hand, the single-modality conditional approaches tend to involve more specific scenarios (e.g., various aspects of editing and customization), and we cover the common conditional scenarios in our work.
A taxonomy of these single-modality methods can be found in Figure \ref{figure:taxonomy_combined}.
Then, we introduce the cross-modality AIGC approaches, and a taxonomy for these methods can be found in Figure \ref{figure:taxonomy_combined2}.
Next, we survey the datasets and benchmarks throughout various modalities.
Following that, we present the challenges of existing AIGC methods, some typical applications for them, and possible future directions.
Lastly, we also discuss some observed trends in the surveyed papers.

\section{Comparison with existing surveys}

Due to the importance and popularity of AIGC, there have been several previous attempts at surveying AIGC from various perspectives.
Wu et al. \cite{wu2023ai} surveyed the pros and cons of AIGC as a whole, as well as some potential challenges and applications.
A few works \cite{ray2023chatgpt,yang2023harnessing,zhang2023one} discuss specifically about ChatGPT and large language models and their potential impacts and challenges.
Gozalo et al. \cite{gozalo2023chatgpt} introduce a few selected works across several modalities.
Xu et al. \cite{xu2023unleashing} surveys the AIGC works in the context of mobile and cloud computing.
Abdollahzadeh et al. \cite{abdollahzadeh2023survey} focus on generative techniques using less data.
Several works \cite{zhou2023comprehensive,zhang2023complete,cao2023comprehensive} focus mainly on surveying image and text generation methods.

Differently, we discuss AIGC over a broad range of modalities, for many single-modality settings as well as for many cross-modality settings.
Our main focus and key contributions lie in comprehensively introducing the unique challenges, representative works and technical directions of various settings and tasks within a broad range of single-modality and cross-modality scenarios.
We organize and catalog these single-modality and cross-modality settings in a unified consistency -- first according to their generated output, then according to the type of conditioning input.
For the first time, these sections and subsections across the various modalities are arranged hierarchically in an organized framework.
We select the most representative works, i.e., the works from the top journals and conferences, as well as highly cited works that are not from these venues, and we we also cite the pioneering works and milestone works for each representative technical line. 
We present these works in a chronological order as best as we can, to provide a comprehensive history and background of the subject.
Furthermore, we provide standardized comparisons and tables of results for AIGC methods throughout many modalities, which is much more extensive than previous works.
In addition, we also provide analysis of the trends in AIGC across the modalities, such as the number of AIGC papers over time in Figure \ref{figure:paper_trends}, and some notable trends in Figure \ref{figure:affilation_counts}, which are also meaningful contributions to the field.

\section{Image Modality}

\label{sec:image}

The image modality was the earliest to undergo developments for deep generative modeling, and often forms the testbed for many foundational techniques such as Generative Adversarial Networks (GANs), Variational Auto-Encoders (VAEs), Normalizing Flows (NFs) and Denoising Diffusion Models (DMs).
This is due to several reasons, including the ready availability of image data, the relative simplicity of images compared to other modalities (e.g., videos or 3D data), and the ease and efficiency of modeling the grid-like RGB data with Convolutional Neural Networks (CNNs).
Therefore, even though we comprehensively explore these foundational techniques (GANs, VAEs, NFs, DMs) in this section, they are also the core techniques which are applied in other modalities as well, albeit with modality-specific modifications.

Initial attempts at image generation with deep learning faced a myriad of difficulties. For instance, many methods faced training instability, which is particularly evident in GANs with the risk of mode collapse. 
Additionally, modeling long-range dependencies and efficiently scaling up image resolution posed significant difficulties.
Besides, generating diverse images was also challenging.
However, the progress over the years have mostly overcome these issues, making it relatively easy to train image generation models to produce diverse and high-quality images that are often difficult to tell apart from real images with the naked eye.
Below, we first discuss the unconditional methods, followed by the conditional methods where various constraints are applied to the generation process.

\subsection{Unconditional Image Generation}
\label{sec:image_unconditional}

\begin{wrapfigure}{r}{0.30\textwidth}
\vspace{-5mm}
    \center
    \includegraphics[width=0.31\textwidth]{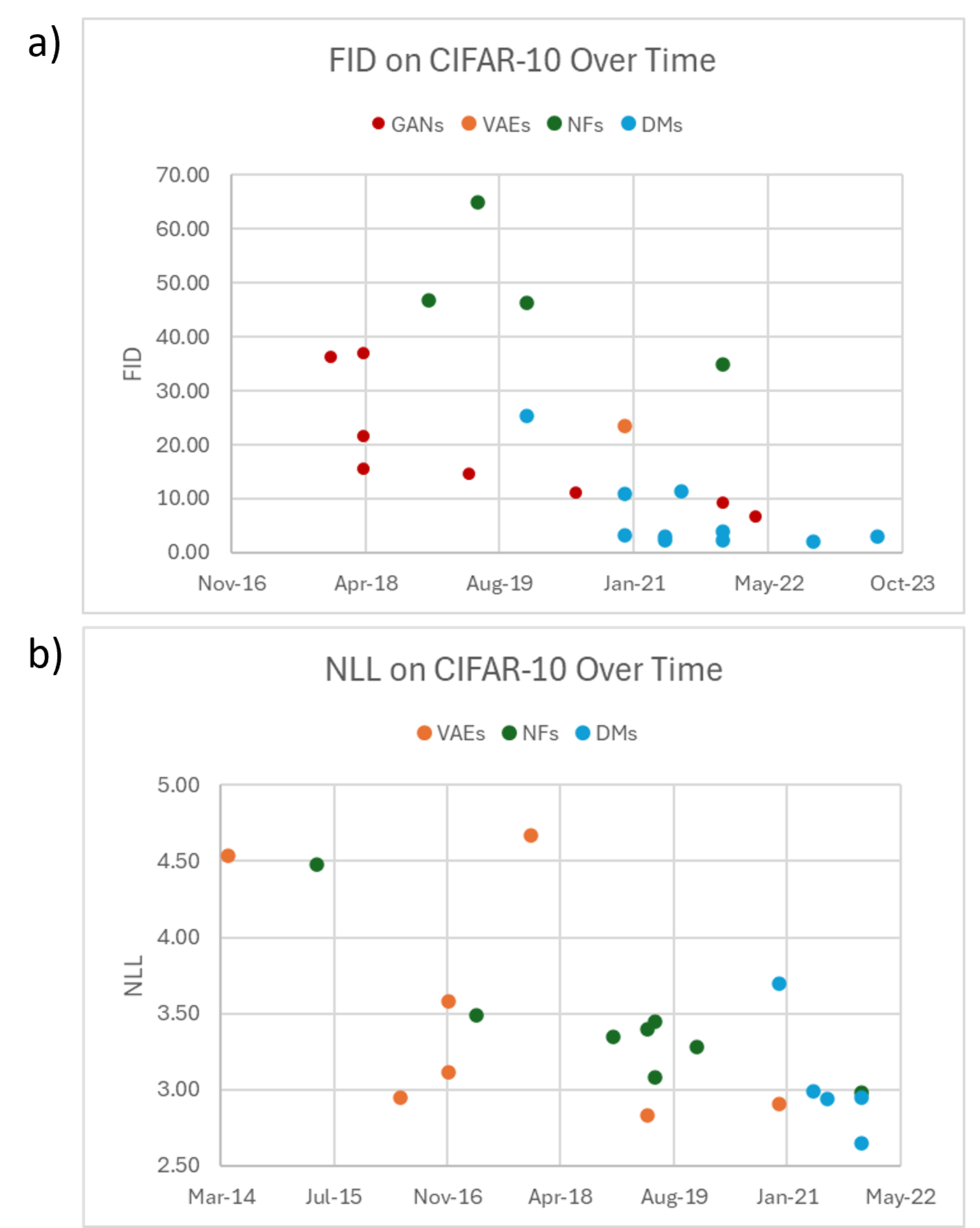}  
    \vspace{-4.5mm}
    \caption{
    Visualization of the trend of improvement of image generation metrics (FID and NLL) on CIFAR-10 dataset. Each point corresponds to a result reported in Table \ref{table:image_methods}, and the points are color-coded according to the type of method.
    }
    \vspace{-4mm}
    \label{figure:image_metrics}
\end{wrapfigure}

Here, we discuss the representative methods for unconditional image generation. 
In general, these unconditional methods are the foundational techniques behind image generation which seek to improve quality of generated images, training stability and efficiency. 
As these methods tend to represent foundational improvements, they are often applicable to conditional generation as well, with only minor modifications.
Below, we split the methods into four categories: Generative Adversarial Networks (GANs), Variational Autoencoders (VAEs), Normalizing Flows (NFs) and Denoising Diffusion Models (DMs).
Table \ref{table:image_methods} shows a summary and comparison of the representative methods in each category.
We also visualize the progression of these image generation methods on the CIFAR-10 dataset in Figure \ref{figure:image_metrics}, where we observe the FID and NLL metrics gradually improving over time. We can also observe the superiority of diffusion models over older methods on these metrics.

\noindent
\textbf{Generative Adversarial Networks} (GANs) have been a very popular choice for image generation since their introduction by Goodfellow et al. \cite{goodfellow2014generative}. 
GANs are trained in an adversarial manner, where a generator generates synthetic images and gets gradient updates from a discriminator that tries to determine if the images are real or fake. 
One major drawback of GANs is their instability during training, and the tendency for mode collapse to occur, where the generators stick to generating only one or a limited set of images.
Since the introduction of GANs \cite{goodfellow2014generative}, many improvements have been proposed \cite{salimans2016improved}, and here we discuss some of the representative methods.

\begin{wraptable}{r}{0.45\textwidth}
\vspace{-2mm}
   \tiny
  \centering
  \caption{\tiny
  Comparison between the quality of images generated by various deep generative models.
  Results are reported on CIFAR-10 ($32\times32$) \cite{Krizhevsky09cifar10}, CelebA-HQ ($1024\times1024$) \cite{karras2018progressive}, FFHQ ($1024\times1024$) \cite{karras2019style} datasets.  
  We report the Fr\'echet Inception Distance (FID) metric \cite{heusel2017gans} for all datasets and the Negative Log-Likelihood (NLL) metric (measured in terms of bits/dimension) for CIFAR-10, where lower is better for both metrics. 
  FID measures the similarity between real and generated images, while NLL measures the ability of the generative model to represent the data distribution.
  A dash `-' indicates that the metric was not reported for that work. The best result in each column is in bold.
  }
  \vspace{-0.35cm}
  \label{table:image_methods}
  \scalebox{0.95}{
  \hspace{-0.4cm}
  \begin{tabular}{lcccccccc}
    \toprule
     & & \multicolumn{2}{c}{CIFAR-10}  & CelebA-HQ & FFHQ \\
    Type & Method & FID($\downarrow$) & NLL($\downarrow$) & FID($\downarrow$) & FID($\downarrow$)  \\        
    \midrule
    \multirow{14}{*}{GANs} & DCGAN \cite{radford2016unsupervised} & 37.11  & - & - & - \\
       & WGAN-GP \cite{gulrajani2017improved} & 36.40  &  - & - & -\\
       & ProGAN \cite{karras2018progressive} & 15.52 &  -   & 7.30 & 8.04 \\
       & SNGAN \cite{miyato2018spectral} & 21.7 & -  & - & -\\    
       & StyleGAN \cite{karras2019style} & - & -  & 5.17 & 4.40 \\
       & SAGAN \cite{zhang2019self} & - & -  & - & -\\
       & BigGAN \cite{brock2019large} & 14.73 & -   & - & -\\
       & StyleGAN2 \cite{karras2020analyzing} & 11.1 & -  &  - & 2.84  \\
       & StyleGAN3-T \cite{karras2021alias} & - & -   &  - & \textbf{2.79} \\
       & GANsformer \cite{hudson2021generative} & -  & -  &  - & 7.42 \\
       &  TransGAN \cite{jiang2021transgan} & 9.26 & -   &  - & - \\  
       &  HiT \cite{zhao2021improved} & -  & - &  8.83 & 6.37 \\    
       &  ViTGAN \cite{lee2022vitgan} & 6.66 & -  & - & - \\     
       &  StyleSwin \cite{zhang2022styleswin} & -  & - &  \textbf{4.43} & 5.07 \\     
    \midrule
    \multirow{7}{*}{VAEs} & VAE \cite{kingma2014auto} & - & $\leq$ 4.54  & - & -   \\
       & VLAE \cite{chen2016variational} & - & $\leq$ 2.95  & - & - \\ 
       & IAF-VAE \cite{kingma2016improved} & -  & $\leq$ 3.11  & - & -  \\       
       & Conv Draw \cite{gregor2016towards} & -  & $\leq$ 3.58   & - & -   \\
       & VQ-VAE \cite{van2017neural} & -  & $\leq$ 4.67  & - & -   \\
      & $\delta$-VAE \cite{razavi2019preventing} &  - & $\leq$ 2.83  & - & -  \\  
       & NVAE \cite{vahdat2020nvae} &  23.49 & $\leq$ 2.91  & - & -  \\
    \midrule   
    \multirow{8}{*}{NFs} & NICE \cite{dinh2015nice} & - & 4.48  &  - & - \\
       & RealNVP \cite{dinh2017density} & - & 3.49   &  - & - \\
       & Glow \cite{kingma2018glow} & 46.90 & 3.35  &  - & -  \\
       & i-Resnet \cite{behrmann2019invertible} &  65.01 & 3.45  &  - & -  \\
       & FFJORD \cite{grathwohl2019ffjord} &  - & 3.40   &  - & -  \\
       & Residual Flow \cite{chen2019residual} &  46.37 & 3.28   &  - & -  \\
       & Flow++ \cite{ho2019flow++} &  - & 3.08  &  - & -  \\
       & DenseFlow \cite{grcic2021densely} &  34.90 & 2.98   &  - & - \\    
    \midrule
    \multirow{10}{*}{DMs} & DDPM \cite{ho2020denoising} &  3.17 & $\leq$ 3.70  & 4.90 & -  \\
       & NCSN \cite{song2019generative} & 25.32 &  -  & -  & - \\
       & NCSNv2 \cite{song2020improved} & 10.87 &  -  & - & - \\
       & Improved DDPM \cite{nichol2021improved} & 11.47  & $\leq$ 2.94    &  - & -  \\
       & VDM  \cite{kingma2021variational}  & 4.00 & \textbf{$\leq$ 2.65}   & - & -  \\          
       & Score SDE (NCSN++) \cite{song2021score} & 2.20 &  - & -  & - \\
       & Score SDE (DDPM++) \cite{song2021score} & 2.92 &  $\leq$ 2.99   & - & - \\        
       & LSGM (balanced)  \cite{vahdat2021score} & 2.17 & $\leq$ 2.95   & -  & - \\ 
       & EDM  \cite{karras2022elucidating} & \textbf{1.97} & -  & - & - & -  \\ 
       & Consistency Distillation \cite{song2023consistency} & 2.93 &  - & - & -  \\  
    \bottomrule
  \end{tabular}
  }
\vspace{-5mm}
\end{wraptable}

Many prominent approaches improve the architecture and design of the generator and discriminator, which are important for smoothly generating high-resolution images.
DCGAN \cite{radford2016unsupervised} is a generative model with a deep convolutional structure, which follows a set of proposed architectural guidelines for better stability.
ProGAN \cite{karras2018progressive} proposes a training methodology for GANs where both the generator and discriminator are progressively grown, which improves quality, stability and variation of outputs.
BigGAN \cite{brock2019large} explores large-scale training of GANs and proposes a suite of tricks to improve the generator and discriminator.
SAGAN \cite{zhang2019self} proposes a self-attention module to model long range, multi-level dependencies across image regions.
StyleGAN \cite{karras2019style} proposes a generator architecture that allows it to learn to separate high-level attributes in the generated images, which gives users more control over the generated image.
StyleGAN2 \cite{karras2020analyzing} re-designs the generator of StyleGAN and proposes an alternative to progressive growing \cite{karras2018progressive}, leading to significant quality improvements.
Subsequently, StyleGAN3 \cite{karras2021alias} proposes small architectural changes to tackle aliasing and obtain a rotation equivariant generator.
More recently, several works explore Transformer architectures \cite{vaswani2017attention} for GANs.
GANsformer \cite{hudson2021generative} explores an architecture combining bipartite attention with convolutions, which enables long-range interactions across the image.
TransGAN \cite{jiang2021transgan} explores a pure transformer-based architecture for GANs.
ViTGAN \cite{lee2022vitgan} integrates the Vision Transformer (ViT) architecture \cite{dosovitskiy2021an} into GANs.
HiT \cite{zhao2021improved} improves the scaling of its Transformer architecture, synthesizing high-definition images.
StyleSwin \cite{zhang2022styleswin} proposes to leverage Swin Transformers \cite{liu2021swin} in a style-based architecture, which is more scalable.
Another orthogonal line of works explores GAN loss functions and regularization. 
For instance, 
WGANs \cite{arjovsky2017wasserstein} minimizes an efficient approximation of the Earth Mover's distance, and is further improved in WGAN-GP \cite{gulrajani2017improved} with a gradient penalty.
SNGAN \cite{miyato2018spectral} proposes spectral normalization, providing regularization that stabilizes training.

\noindent
\textbf{Variational Autoencoders} (VAEs) \cite{kingma2014auto,rezende2014stochastic} are a variational Bayesian approach to image generation, which learn to encode images to a probabilistic latent space and reconstruct images from that latent space. 
Then, new images can be generated by sampling from the probabilistic latent space to produce images.
VAEs are generally more stable to train than GANs, but struggle to achieve high resolutions and tend to display blurry artifacts.

Many works have been proposed to improve VAEs. They mostly aim to either improve the quality of generated images (e.g., via tackling the posterior collapse issue) or to improve the learning of high-level concepts with VAEs.
An early work, IWAE \cite{burda2015importance} explores a strictly tighter log-likelihood lower bound that is derived from importance weighting, which greatly improves visual quality.
Subsequently, several representative works aim to explore the \textit{learning of high-level concepts} in VAEs, which are discussed next.
VLAE \cite{chen2016variational} proposes to learn global representations by combining VAEs with neural autoregressive models. allowing them to better uncover the ``global structure'' of images with VAEs.
Conv Draw \cite{gregor2016towards} adopts a hierarchical latent space and a fully convolutional architecture for better compression of high-level concepts, with the vision that the generation process can probabilistically ``fill in'' the lower-level information during decoding. 
Building upon previous works in learning high-level concepts, $\beta$-VAE \cite{higgins2017beta} aims to discover interpretable and factorised latent representations from unsupervised training with raw image data. Specifically, $\beta$-VAE \cite{higgins2017beta} introduces a hypermaraeter $\beta$ that can emphasize the learning of statistically independent latent factors, which greatly improves the disentanglement of concepts in the learned latent space.
This is further extended in \cite{burgess2018understanding} with a progressive training paradigm for better reconstruction quality.
Besides, to further improve reconstruction quality of VAEs, a key direction has been to \textit{tackle the issue of posterior collapse} -- where the the VAE's latent codes are ``ignored'' when the VAE is paired with a powerful autoregressive decoder.
The representative work in this area, VQ-VAE \cite{van2017neural} proposes to learn discrete latent representations by incorporating ideas from vector quantisation. 
A key insight is that using discrete latent codes instead of continuous ones makes it difficult for the autoregressive decoder to ignore the latents, thus allowing the VQ-VAE to circumvent the posterior collapse issue.
VQ-VAE 2 \cite{razavi2019generating} further scales and enhances the autoregressive priors used in VQ-VAE, which generates higher quality samples.
Another approach to overcome posterior collapse has been $\delta$-VAE \cite{razavi2019preventing}, which proposes to constrain the latent space by ensuring there is an adequate distance between the prior and the posterior (i.e., the latents are making a difference and are not being ignored).
Besides exploring high-level concept learning or tackling the posterior collapse issue, more recent works \cite{Ghosh2020From,vahdat2020nvae} also investigate orthogonal directions to improve the reconstruction quality of VAEs.
RAE \cite{Ghosh2020From} investigates the noise injection in VAEs, proposing regularization schemes and an ex-post density estimation scheme as substitutes for the proposed deterministic RAE. These proposed schemes can be a drop-in replacement for many existing VAE architectures, for improved generation quality. 
Moreover, in another orthogonal direction, NVAE \cite{vahdat2020nvae} proposes a deep hierarchical VAE with a carefully designed network architecture involving depth-wise separable convolutions and batch normalization, achieving good results for image generation.

\noindent
\textbf{Normalizing Flows} (NFs) \cite{rezende2015variational,dinh2015nice}
transform a simple probability distribution into a richer and more complex distribution (e.g., image data distribution) via a series of invertible and differentiable transformations. 
After learning the data distribution, images can be generated by sampling from the initial density and applying the learned transformations.
NFs hold the advantage of being able to produce tractable distributions due to their invertibility, which allows for efficient image sampling and exact likelihood evaluation.
However, it can be difficult for NFs to achieve high resolutions and high image quality.

Rezende et al. \cite{rezende2015variational} use NFs for variational inference and also develop categories of finite and infinitesimal flows. 
NICE \cite{dinh2015nice} proposes to learn complex non-linear transformations via deep neural networks.
IAF \cite{kingma2016improved} proposes a new type of autoregressive NF that scales well to high-dimensional latent spaces.
RealNVP \cite{dinh2017density} defines a class of invertible functions with tractable Jacobian determinant, by making the Jacobian triangular.
Glow \cite{kingma2018glow} defines a generative flow using an invertible $1 \times 1$ convolution.
Neural ODEs \cite{chen2018neural} introduced continuous normalizing flows, which is extended in FFJORD \cite{grathwohl2019ffjord} with an unbiased stochastic estimator of the likelihood that allows completely unrestricted architectures.
i-Resnets \cite{behrmann2019invertible} introduce a variant of ResNet with a different normalization scheme that is invertible, and use them to define a NF for image generation.
Some other recent advancements in NF methods include Flow++ \cite{ho2019flow++}, Residual Flow \cite{chen2019residual} and DenseFlow \cite{grcic2021densely} which further improve the quality of the generated image.

\noindent
\textbf{Denoising Diffusion Models} (DMs) \cite{sohl2015deep,ho2020denoising} generate images by iteratively ``denoising'' random noise to produce an image from the desired data distribution.
Such an approach can also be seen as estimating the gradients (i.e., score function) of the data distribution \cite{song2019generative}.
DMs have become very popular recently as they tend to be more stable during training and avoid issues like mode collapse of GANs.
Furthermore, DMs also tend to produce high-quality samples and scale well to higher resolutions, while also being able to sample diverse images.
However, due to their iterative approach, DMs may require longer training and inference times as compared to other generative models.

DDPM \cite{ho2020denoising} designs a denoising diffusion probabilistic model which builds upon diffusion probabilistic models \cite{sohl2015deep} and incorporates the ideas of denoising over multiple steps from NCSN \cite{song2019generative}.
NCSN v2 \cite{song2020improved} introduces a way to scale noise at each step for more effective score-based modeling, which leads to improvements in image quality.
Score-SDE \cite{song2021score} proposes a stochastic differential equation (SDE) to gradually remove the noise, and use numerical SDE solvers to generate image samples.
DDIM \cite{song2021denoising} significantly improves efficiency by accelerating the sampling process, introducing a class of iterative implicit probabilistic models that can use significantly less sampling steps.
Improved DDPM \cite{nichol2021improved} proposes to learn the variance used in the reverse process and design a cosine noise schedule to improve upon DDPM \cite{ho2020denoising}.
D3PM \cite{austin2021structured} explores discrete diffusion models for quantized images.
Karras et al. \cite{karras2022elucidating} investigate the sampling processes and the networks' inputs, outputs and loss functions, which enable significant improvements in quality of synthesized image.
Progressive Distillation \cite{salimans2022progressive} reduces the sampling time of DMs by progressively distilling a pre-trained DM into a smaller model requiring less steps.
Besides, some works explore combining multiple DMs for better performance, e.g., Cascaded DMs \cite{ho2022cascaded}.
Another line of work investigates integrating DMs with ideas from other types of generative models.
VDMs \cite{kingma2021variational} propose variational diffusion models which optimize
the diffusion process parameters jointly with the rest of the model, which makes it a type of VAE.
LSGM \cite{vahdat2021score} leverages a VAE framework to map the image data into latent space, before applying a score-based generative model in the latent space.
Denoising Diffusion GANs \cite{xiao2022tackling} models the denoising distributions
with conditional GANs, achieving speed-up compared to current diffusion models while outperforming GANs in terms of sample diversity.
DiffFlow \cite{zhang2021diffusion} is proposed to extend NFs by adding noise to the sampling trajectories, while also making the forward diffusion process trainable.
Consistency models \cite{song2023consistency} have been proposed as a new family of models to directly generate images from noise in a single step, which borrows ideas from both DMs \cite{karras2022elucidating} and NF-based models \cite{chen2018neural}.

\subsection{Conditional Image Generation}
\label{sec:image_conditional}

\begin{wrapfigure}{r}{0.45\textwidth}
\vspace{-5mm}
    \center
    \includegraphics[width=0.46\textwidth]{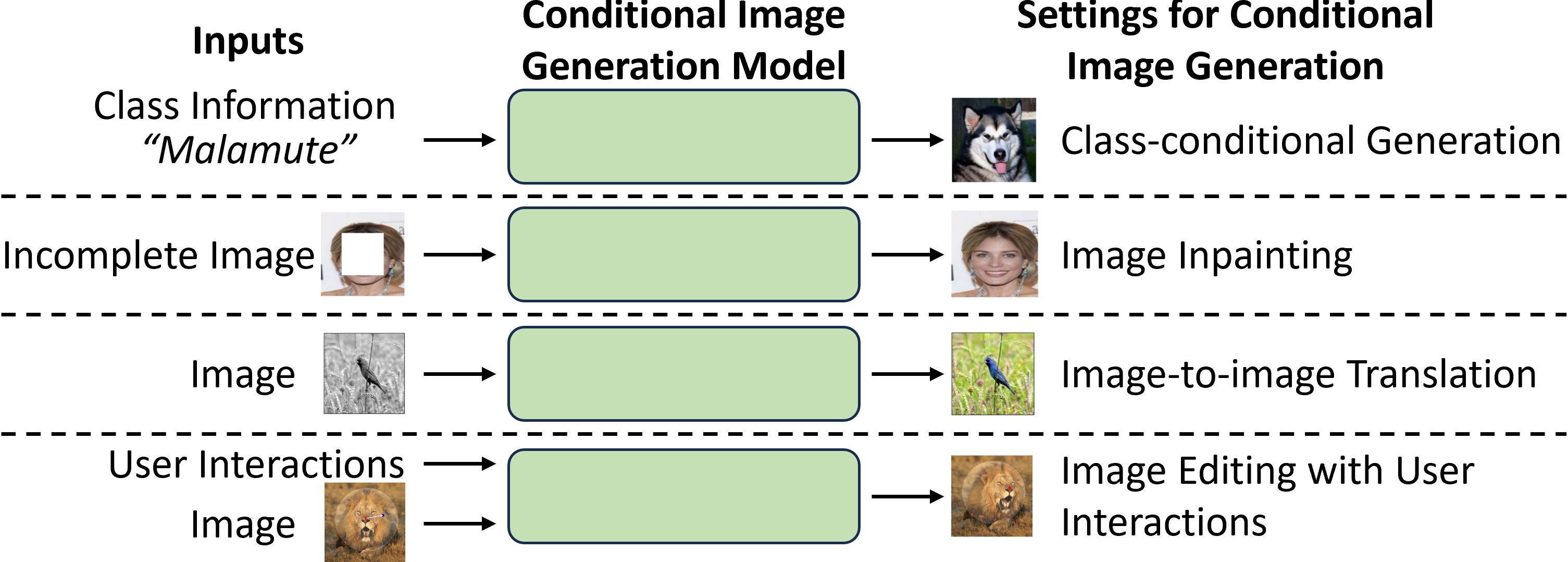}  
    \vspace{-5.5mm}
    \caption{
    Illustration of various conditional image generation settings.
    Examples obtained from \cite{ho2021classifier,yu2018generative,saharia2022palette,pan2023draggan}. 
    }
    \vspace{-3mm}
    \label{figure:image_conditional}
\end{wrapfigure}

In order to control the content of the generated image, various types of conditioning information can be provided as input to the generative models.
Below, we first discuss the methods that generate images based on class information, which is one of the basic ways to control the generated content.
Then, we discuss the usage of input images as conditioning information (for image inpainting or image-to-image translation), as well as with user inputs.
A summary of these settings is shown in Figure \ref{figure:image_conditional}.
Note that, the conditioning with other modalities (e.g., text) have been left to Section \ref{sec:crossmodal_image}.

\noindent
\textbf{Class-conditional Generation} aims to produce images containing the specified class by conditioning the generative model on the class labels (e.g., ``dog'' or ``cat''), and is one of the basic approaches for conditional image generation. 
Below we discuss some representative works, including some foundational developments in conditional generation.

Conditional GANs (cGANs) \cite{mirza2014conditional} are the first to extend GANs \cite{goodfellow2014generative} to a conditional setting, where both the generator and discriminator are conditioned on additional information (e.g., class labels that are encoded as one-hot vectors).
AC-GAN \cite{odena2017conditional} introduces an auxiliary classifier to add more structure to the GAN latent space.
To improve the flexibility of AC-GANs, PPGN \cite{nguyen2017plug} aims to generate images at test time according to a replaceable condition classifier network in a plug-and-play manner.
Then, to better modulate the intermediate feature representations based on the conditioning information, Dumoulin et al. \cite{dumoulin2017a} proposes conditional normalization mechanisms in the architecture.
Following the development of DM-based approaches, Score-SDE \cite{song2021score} explores a score-based generative approach for conditional generation, where the conditional reverse-time SDE can be estimated from unconditional scores.
To guide the DM's generation process without heavy re-training, classifier guided diffusion sampling \cite{dhariwal2021diffusion} proposes to use the gradients of a pre-trained image classifier.
Then, to remove the need for a classifier, classifier-free guided sampling \cite{ho2021classifier} is proposed as a way to train a conditional and unconditional model by training with the class information as inputs and randomly dropping the class information during training, which provides control over the generation process.
Subsequently, Self-guided Diffusion Models \cite{hu2023selfguided} aim to eliminate the need for labeled image data by creating self-annotations with a self-supervised feature extractor.

\noindent
\textbf{Image Inpainting} is where a model takes as input an image that is incomplete in some way (e.g., missing pixels), and tries to complete the image.
Many approaches are designed to better leverage the context surrounding the missing patch, and to handle patches of various shapes.

Pathak et al. \cite{pathak2016context} are the first to adopt GANs for image inpainting, and introduce a Context Encoder to generate the contents of a region based on the surrounding context.
Iizuka et al. \cite{iizuka2017globally} propose to use a combination of global and local context discriminators to complete the image with better local and global consistency.
Shift-Net \cite{yan2018shift} aims to use the encoder features of the known regions in the missing regions, and introduces a shift-connection layer to its model architecture for feature rearrangement.
Yu et al. \cite{yu2018generative} present a contextual attention layer with channel-wise weights to encourage coherency between the existing and generated pixels.
Partial Convolutions \cite{liu2018image} are convolutions that are masked and renormalized such that the convolutional results depend only on the non-missing regions.
Gate Convolutions \cite{yu2019free} learn a dynamic feature gating mechanism which performs well when the missing regions have arbitrary shapes, i.e., free-form inpainting.
Repaint \cite{lugmayr2022repaint} proposes a DM for inpainting, which leverages a pre-trained unconditional DM and does not require training on the inpainting task itself, allowing it to generalize to any mask shape given during testing while having powerful semantic generation capabilities.

\noindent
\textbf{Image-to-Image translation between different visual domains} is an important task, where the model takes an image as input and produces an image belonging to another domain.
Popular applications include style transfer (e.g., sketch to photo, synthetic to real), pose transfer (e.g., modifying a human to have a given pose), attribute transfer (e.g., dog breed translation), colorization (e.g., RGB from grayscale images), etc.
Although these applications are very different, they can often be tackled with similar approaches (as introduced below), but with image data from the corresponding domains.
Research in this field often aim to learn without the strong constraint of requiring paired images from both domains. Many works also aim to improve the diversity of generated images while preserving key attributes.

Pix2Pix \cite{isola2017image} investigates using cGANS \cite{mirza2014conditional} to tackle image-to-image translation problems in a common framework.
To reduce the requirement of paired data, CycleGAN \cite{zhu2017unpaired} designs an algorithm that can translate between image domains without paired examples between domains in the training data, through training with a cycle consistency loss.
UNIT \cite{liu2017unsupervised} also tackles the case where no paired examples exist between domains, and approaches the task by enforcing a shared VAE latent space between domains with an adversarial training objective.
DiscoGAN \cite{kim2017learning} aims to discover cross-domain relations when given unpaired data, and can transfer styles while preserving key attributes in the output image.
DRIT \cite{lee2018diverse} proposes to generate diverse outputs with unpaired training data via disentangling the latent space into a domain-invariant content space and a domain-specific attribute space, together with a cross-cycle consistency loss.
StarGAN \cite{choi2018stargan} explores handling more than two domains with a single model by introducing the target domain label as an input to the model, such that users can control the translation of the image into the desired domain.
FUNIT \cite{liu2019few} explores the few-shot setting, where only a few images of the target class are needed at test time.
Recently, Palette \cite{saharia2022palette} adopts a DM for image-to-image translation which achieves good performance.

\noindent
\textbf{Image editing and manipulation with user interaction} aims to control the image synthesis process via human input (e.g., scribbles and sketches).
Methods in this category explore different ways to incorporate user inputs into deep image generation pipelines.

Zhu et al. \cite{zhu2016generative} explore manipulating the image's shape and colour through the users' scribbles, and relies on the learned manifold of natural images in GANs to constrain the generated image to lie on the learned manifold, i.e., a GAN inversion approach.
SketchyGAN \cite{chen2018sketchygan} aims to synthesize images from many diverse class categories given a sketch.
Ghosh et al. \cite{ghosh2019interactive} also use a single model to generate a wide range of object classes, and introduce a feedback loop where the user can interactively edit the sketch based on the network's recommendations.
Paint2pix \cite{singh2022paint2pix} explores a progressive image synthesis pipeline, which allows a user to progressively synthesize the images from scratch via user interactions at each step.
Recently, DragGAN \cite{pan2023draggan} proposes to allow users to flexibly edit the image by dragging points of the image toward target areas.
A similar approach is adopted by DragDiffusion \cite{shi2023dragdiffusion}, but a DM is used instead.

\noindent
\textbf{Others.}
Various works also explore the conditional image generation process in other directions.
Some works \cite{takagi2023high} take in human brain activity (e.g., fMRI signals) to generate the corresponding visual images.
Another research direction involves layout generation \cite{chai2023layoutdm}, where the layout of the room is generated conditioned on the given elements with user-specified attributes.

\subsection{Summary of Image Modality}
In summary, in this section we first introduced the unconditional image generative methods in Section \ref{sec:image_unconditional}, including GANs \cite{goodfellow2014generative}, VAEs \cite{kingma2014auto}, NFs \cite{dinh2015nice}, and DMs \cite{ho2020denoising}.
Notably, these are foundational techniques for AIGC methods in other modalities as well.
These methods have been applied in various conditional settings (in Section \ref{sec:image_conditional}), such as class-conditional image generation \cite{mirza2014conditional}, image inpainting \cite{pathak2016context}, image-to-image translation between different visual domains \cite{isola2017image}, and image manipulation with user interactions \cite{zhu2016generative}.
As these conditional scenarios each transform images towards different objectives, they explore different architectures and designs to achieve their objectives.

\vspace{2mm}

\vspace{-1mm}

\section{Video modality}

\begin{wraptable}{r}{0.5\textwidth}
\vspace{-4mm}
  \tiny
  \centering
  \caption{\tiny
  Comparison between representative video generative models.
  Results are reported on the UCF-101 \cite{soomro2012ucf101} and Sky Timelapse \cite{xiong2018learning} datasets.  
  We report the Fr\'echet Video Distance (FVD) evaluation metric \cite{unterthiner2019fvd} based on the standardized implementation by \cite{skorokhodov2022styleganv} (lower is better), where FVD\textsubscript{16} and FVD\textsubscript{128} refer to evaluation of clips with 16 and 128 frames respectively. 
  FVD measures the similarity between generated videos and real videos.
  We also report the Inception Score (IS) \cite{salimans2016improved,saito2017temporal} on UCF-101, which measures the class diversity over generated videos and how clearly each generated video corresponds to a class.
  Besides, we also report the resolution (Res.) of each video frame for each evaluation result.
  An asterisk (*) indicates that methods were trained on both training and testing videos of the dataset.
  A dash `-' indicates that the metric was not reported for that work. The best result in each column is in bold.
  }
  \vspace{-3mm}
  \label{table:video_methods}
  \scalebox{0.95}{
  \hspace{-0.4cm}
  \begin{tabular}{l|c|ccc|ccc}
    \toprule 
     &  & \multicolumn{3}{c}{UCF-101} & \multicolumn{2}{|c}{Sky Timelapse}  \\    
    Method & Res. & IS($\uparrow$) & FVD\textsubscript{16}($\downarrow$)  &  FVD\textsubscript{128}($\downarrow$) &  FVD\textsubscript{16}($\downarrow$) & FVD\textsubscript{128}($\downarrow$)  \\        
    \midrule
    VGAN \cite{vondrick2016generating} & 64$\times$64 & 8.31$\pm$0.09 & - & - & - & - \\
    TGAN \cite{saito2017temporal} & 64$\times$64 & 11.85$\pm$0.07 & - & - & - & - \\
    MoCoGAN \cite{tulyakov2018mocogan} & 64$\times$64 & 12.42$\pm$0.07 & - & - & - & - \\
    MoCoGAN \cite{tulyakov2018mocogan} &  256$\times$256  & 10.09$\pm$0.30 & 2886.8 & 3679.0 & 206.6 & 575.9 \\
    DVD-GAN \cite{clark2019adversarial} & 128$\times$128   &  32.97$\pm$1.7 &  - &  - &  - &  - \\
    MoCoGAN+SG2 \cite{karras2020analyzing,skorokhodov2022styleganv} &  256$\times$256  & 15.26$\pm$0.95 & 1821.4 & 2311.3 & 85.88 & 272.8 \\
    MoCoGAN-HD \cite{tian2021good} & 256$\times$256  & 23.39$\pm$1.48 & 1729.6 & 2606.5 & 164.1 & 878.1 \\
    VideoGPT \cite{yan2021videogpt} & 256$\times$256  & 12.61$\pm$0.33 & 2880.6 &  - & 222.7 & -  \\
    VideoGPT \cite{yan2021videogpt} & 128$\times$128 & 24.69$\pm$0.30 & - & - & - & - \\ 
    DIGAN \cite{yu2022generating} & 128$\times$128 & 29.71$\pm$0.53 & - & - & - & - \\     
    DIGAN \cite{yu2022generating} & 256$\times$256  & 23.16$\pm$1.13 & 1630.2 & 2293.7 & 83.11 & 196.7 \\
    Video Diffusion* \cite{ho2022video} & 64$\times$64 & 57.00$\pm$0.62 & - & - & - & - \\ 
    TATS \cite{ge2022long} &  128$\times$128  &  57.63$\pm$0.24 &  - &  - &  - &  - \\
    StyleGAN-V* \cite{skorokhodov2022styleganv} & 256$\times$256  & 23.94$\pm$0.73  & 1431.0 & 1773.4 & 79.52 & 197.0 \\  
    VideoFusion \cite{luo2023videofusion} & 128$\times$128 & 72.22 & - & - & - & - \\ 
    PVDM \cite{yu2023video} & 256$\times$256  & \textbf{74.40$\pm$1.25} & \textbf{343.6} & \textbf{648.4} & \textbf{55.41} & \textbf{125.2} \\    
    \bottomrule
  \end{tabular}
  }
\vspace{-8mm}
\end{wraptable}

Following the developments in image-based AIGC, there has also been much attention on video-based AIGC, which has many applications in the advertising and entertainment industries.
However, video generation remains a very challenging task.
Beyond the difficulties in generating individual frames/images, the generated videos must also be temporally consistent to ensure coherence between frames, which can be extremely challenging for longer videos.
Moreover, it is also difficult to produce realistic motions.
Due to the much larger size of the output, it is also challenging to generate videos quickly and efficiently.
We discuss some approaches to overcome these challenges below, starting with the unconditional methods.

\vspace{-1mm}

\subsection{Unconditional Video Generation}
\label{sec:video_unconditional}

Here, we discuss the methods for unconditional generation of videos, which generally aim to foundationally improve the video generation quality, diversity and stability.
They are mostly applicable to the conditional setting as well, with some modifications.
Table \ref{table:video_methods} shows a summary and comparison between some representative methods.

VGAN \cite{vondrick2016generating} proposes a GAN for video generation with a spatial-temporal convolutional architecture.
TGAN \cite{saito2017temporal} uses two separate generators -- a temporal generator to output a set of latent variables and an image generator that takes in these latent variables to produce frames of the video, and adopts a WGAN \cite{arjovsky2017wasserstein} model for more stable training.
MoCoGAN \cite{tulyakov2018mocogan} learns to decompose motion and content by utilizing both video and image discriminators, allowing it to generate videos with the same content and varying motion, and vice versa.
DVD-GAN \cite{clark2019adversarial} decomposes the discriminator into spatial and temporal components to improve efficiency.
MoCoGAN-HD \cite{tian2021good} improves upon MoCoGAN by introducing a motion generator in the latent space and also a pre-trained and fixed image generator, which enables generation of high-quality videos.
VideoGPT \cite{yan2021videogpt} and Video VQ-VAE \cite{walker2021predicting} both explore using VQ-VAE \cite{van2017neural} to generate videos via discrete latent variables.
DIGAN \cite{yu2022generating} designs a dynamics-aware approach leveraging a neural implicit representation-based GAN with temporal dynamics.
StyleGAN-V \cite{skorokhodov2022styleganv} also proposes a continuous-time video generator based on implicit neural representations, and is built upon StyleGAN2 \cite{karras2020analyzing}.
Following the development of DMs, Video Diffusion \cite{ho2022video} explores video generation through a DM with a video-based 3D U-Net architecture.
Subsequent developments for video diffusion models employ latent diffusion models (LDMs) for video generation, such as Video LDM \cite{blattmann2023align}, PVDM \cite{yu2023video} and VideoFusion \cite{luo2023videofusion}, which show a strong ability to generate coherent and long videos at high resolutions.
Specifically, Video LDM \cite{blattmann2023align} first pre-trains an LDM on images only, then fine-tunes the LDM on videos.
PVDM \cite{yu2023video} introduces a projected latent space for the video diffusion model, which factorizes the complex cubic array structure of videos into three latent vectors, thus facilitating computational efficiency. 
To improve the diffusion process, VideoFusion \cite{luo2023videofusion} introduce a decomposed diffusion process which uses noise that is correlated between the various frames of the video, leading to more consistent generated content across the video frames.
At the same time as these developments in diffusion models, another string of recent works \cite{harvey2022flexible,ge2022long,yin2023nuwa} also aim to improve the generation of long videos.
For instance, to generate long videos, TATS \cite{ge2022long} introduces a hierarchical process with  an autoregressive Transformer to generate sparse latent frames to form a global structure, followed by an interpolation Transformer to interpolate between these sparse frames.
Another approach by FDM \cite{harvey2022flexible} explores a flexible diffusion model that can be flexibly conditioned on any number of frames, allowing various sampling schemes to be executed to generate long videos through its generation and infilling capability.
NUWA-XL \cite{yin2023nuwa} further develops a coarse-to-fine DM architecture that is directly trained on long videos, thus eliminating the training-inference gap where most previous works train/fine-tune their method on short videos but expect them to produce long videos at test time.

\subsection{Conditional Video Generation}
\label{sec:video_conditional}

\begin{wrapfigure}{r}{0.45\textwidth}
\vspace{-5mm}
    \center
    \includegraphics[width=0.46\textwidth]{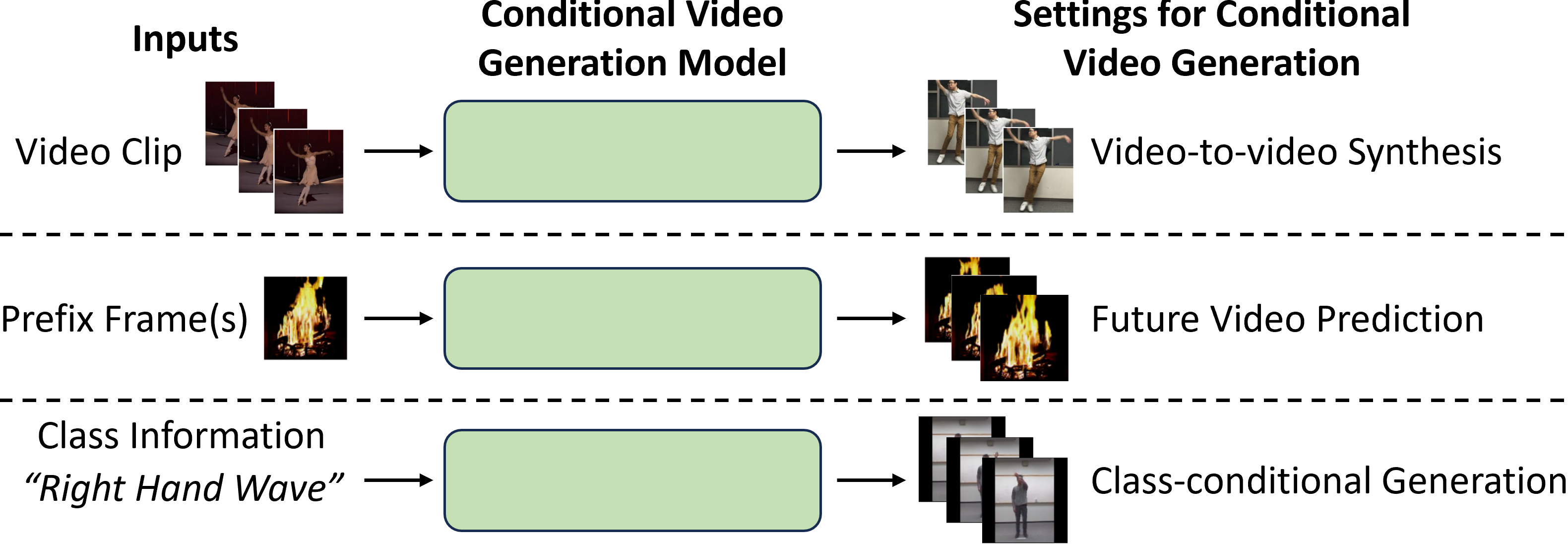}     
    \vspace{-5.5mm}
    \caption{Illustration of various conditional video generation inputs.
    Examples obtained from \cite{chan2019everybody,dorkenwald2021stochastic,ni2023conditional}.
    }
    \label{figure:video_conditional}
\vspace{-4mm}
\end{wrapfigure}

In conditional video generation, users can control the content of the generated video through various inputs. Here, we present the scenarios where the input information can be a collection of video frames (i.e., including single images), or a class.
Figure \ref{figure:video_conditional} summarizes these settings.

\noindent
\textbf{Video-to-Video Synthesis} aims to generate a video conditioned on an input video clip.
There are many applications for this task, including motion transfer and synthetic-to-real synthesis.
Most approaches identify ways to transfer information (e.g., motion information) to the generated video while maintaining consistency in other aspects (e.g., identities remain the same).
Vid2vid \cite{wang2018video} learns a cGAN using paired input and output videos with a spatio-temporal learning objective to learn to map videos from one visual domain to another.
Chan et al. \cite{chan2019everybody} extract the pose sequence from a subject and transfers it to a target subject, which allows users to synthesize people dancing by transferring the motions from a dancer.
Wang et al. \cite{wang2019few} proposes a few-shot approach for video-to-video synthesis which requires only a few videos of a new person to perform the transfer of motions between subjects.
LIA \cite{wang2022latent} approaches the video-to-video task without explicit structure representations, and achieve it purely by manipulating the latent space of a GAN \cite{goodfellow2014generative} for motion transfer.

\noindent
\textbf{Future Video Prediction} is where a video generative model takes in some prefix frames (i.e., an image or video clip), and aim to generate the future frames.
Srivastava et al. \cite{srivastava2015unsupervised} proposes to learn video representations with LSTMs in an unsupervised manner via future prediction, which allows it to generate future frames of a given input video clip.
Walker et al. \cite{walker2016uncertain} adopts a conditional VAE-based approach for self-supervised learning via future prediction, which can produce multiple predictions for an ambiguous future.
PredNet \cite{lotter2017deep} designs an architecture where only the deviations of predictions from early network layers are fed to subsequent network layers, which learns better representations for motion prediction.
SV2P \cite{babaeizadeh2018stochastic} aims to provide effective stochastic multi-frame predictions for real-world videos via an approach based on stochastic inference.
MD-GAN \cite{xiong2018learning} presents a GAN-based approach to generate realistic time-lapse videos given a photo, and produce the videos via a multi-stage approach for more realistic modeling.
Dorkenwald et al. \cite{dorkenwald2021stochastic} uses a conditional invertible neural network to map between the image and video domain, allowing more control over the video synthesis task.

\noindent
\textbf{Class-conditional Generation} aims to generate videos containing activities according to a given class label. 
A prominent example is the action generation task, where action classes are fed into the video generative model.
CDNA \cite{finn2016unsupervised} proposes an unsupervised approach for learning to generate the motion of physical objects while conditioned on the action via pixel advection.
PSGAN \cite{yang2018pose} leverages human pose as an intermediate representation to guide the generation of videos conditioned on extracted pose from the input image and an action label.
GameGAN \cite{kim2020learning} trains a GAN to generate the next video frame of a graphics game based on the key pressed by the agent.
ImaGINator \cite{wang2020imaginator} introduces a spatio-temporal cGAN architecture that decomposes appearance and motion for better generation quality.
Recently, LFDM \cite{ni2023conditional} proposes a DM for temporal latent flow generation based on class information, which is more efficient than generating in the pixel space.

\noindent
\textbf{Others.}
Some works \cite{blattmann2021ipoke,hao2018controllable} propose pipelines for users to interact with the video generation process, e.g., via pokes \cite{blattmann2021ipoke} or a user-specified trajectory \cite{hao2018controllable}.
Besides, Mahapatra and Kulkarni \cite{mahapatra2022controllable} aim toward controlling the animation of fluids from images via various user inputs such as arrows and masks, enabling the video generation of fluids.

\subsection{Summary of Video Modality}
To recap, in this section we discussed the unconditional video generative methods in Section \ref{sec:video_unconditional}, highlighting the progress of video generative models from predominantly GAN-based methods \cite{vondrick2016generating} to DM-based methods \cite{ho2022video} over the years. This development trend closely follows that of the AIGC methods in the image modality (Section \ref{sec:image}).
These foundational generation techniques have been extended to various conditional settings (in Section \ref{sec:video_conditional}), including video-to-video synthesis \cite{wang2018video}, future video prediction \cite{srivastava2015unsupervised}, and class-conditional video generation \cite{finn2016unsupervised}.

\vspace{2mm}

\section{Text Modality}
\label{sec:text_modality}

Another field which has received a lot of attention is text generation, which has gained more widespread interest with well-known chatbots such as ChatGPT.
Text generation is a challenging task due to several reasons.
Initial approaches found it challenging to adopt generative methods such as GANs for text representations which are discrete, which also led to issues with training stability.
It is also challenging to maintain coherence and consistently keep track of the context over longer passages of text. 
Moreover, it is difficult to apply deep generative models to generate text that adhere to grammatical rules while also capturing the intended tone, style and level of formality.

Text generation models are mostly trained to produce text while conditioned on a text input (e.g., input question, preceding text, or text to be translated).
Therefore, due to the way they are trained, even though text models can often generate text unconditionally, but the same models can also be conditioned on input text.
In this section, to maintain the consistency of the taxonomy levels and the terminology with other sections, we categorize these text generation methods as unconditional methods.
This is because they are able to generate text unconditionally and are also the foundational techniques in this field.
Table \ref{table:text_methods} reports the performance of some recent text generation methods.

\vspace{2mm}

\subsection{Unconditional Text Generation}

We categorize the text generation methods based on their techniques, namely: VAEs, GANs, and Autoregressive Transformers, DMs.

\begin{wraptable}{r}{0.6\textwidth}
\vspace{-2mm}
  \tiny
  \centering
  \caption{\tiny
  Comparison between recent representative text generative models.
  Results are reported on common sense reasoning benchmarks BoolQ \cite{clark2019boolq}, WinoGrande \cite{sakaguchi2021winogrande}, ARC easy and ARC challenge (ARC-e and ARC-c) \cite{clark2018think}, as well as closed book question answering benchmarks Natural Questions \cite{kwiatkowski2019natural} and TriviaQA \cite{joshi2017triviaqa}.
  For all datasets, we report accuracy as the evaluation metric (higher is better).
  We report results for the fine-tuned setting where the pre-trained model is fine-tuned on the dataset as a downstream task, and also the zero-shot setting where the model does not get any additional task-specific training.
  We also report the size of each model, i.e., number of parameters.
  A dash `-' indicates that the metric was not reported for that work.
  }
  \vspace{-0.25cm}
  \label{table:text_methods}
  \scalebox{0.95}{
  \hspace{-0.4cm}
  \begin{tabular}{lcccccc|cc}
    \toprule
    Setting & Method & Size & BoolQ & WinoGrande & ARC-e & ARC-c & NaturalQuestions & TriviaQA  \\    
    \midrule
    \multirow{4}{*}{Fine-tuned} & T5-Base \cite{raffel2020exploring} & 223M & 81.4 & 66.6 & 56.7 & 35.5 & 25.8 & 24.5 \\
    &  T5-Large \cite{raffel2020exploring} & 770M & 85.4  & 79.1 & 68.8 &  35.5 & 27.6 & 29.5 \\    
    &  Switch-Base \cite{fedus2022switch} & 7B & -  & 73.3 & 61.3 &  32.8 & 26.8 & 30.7 \\      
    &  Switch-Large \cite{fedus2022switch} & 26B & -  & 83.0 & 66.0 &  35.5 & 29.5 & 36.9 \\  
    \midrule
    \multirow{7}{*}{Zero-shot} & GPT-3 \cite{brown2020language} & 175B & 60.5 & 70.2 & 68.8 & 51.4 & 14.6 & -\\
    &  Gopher \cite{rae2021scaling} & 280B & 79.3  & 70.1 & - &  - & 10.1 & 43.5 \\
    &  Chinchilla \cite{hoffmann2022training} & 70B & 83.7  & 74.9 & - &  - & 16.6 & 55.4 \\
    &  PaLM \cite{chowdhery2022palm} & 62B & 84.8  & 77.0 & 75.2 &  52.5 & 18.1 & - \\
    &  PaLM \cite{chowdhery2022palm} & 540B & 88.0  & 81.1 & 76.6 &  53.0  & 21.2 & - \\
    &  LLaMA \cite{touvron2023llama} & 7B & 76.5  & 70.1 & 72.8 &  47.6 & - & - \\
    &  LLaMA \cite{touvron2023llama} & 33B & 83.1  & 76.0 & 80.0 &  57.8  & 24.9 & 65.1 \\   
    \bottomrule
  \end{tabular}
  }
\vspace{-2mm}
\end{wraptable}

\noindent
\textbf{VAEs.}
Bowman et al. \cite{bowman2016generating} first explores a VAE-based approach for text generation, which overcomes the discrete representation of text by learning a continuous latent representation through a VAE.
Subsequent works \cite{yang2017improved,dieng2019avoiding} have explored ways to improve the training stability and avoid the problem of KL collapse where the VAE's latent representation does not properly encode the input text.
However, these methods generally still do not achieve high levels of performance, stability and diversity.

\noindent
\textbf{GANs.}
Training text-based GANs poses a challenge due to the non-differentiability of the discrete text representation.
Thus, building upon earlier works \cite{li2016deep} that learn to generate text via reinforcement learning (RL), many GAN-based approaches leverage RL to overcome the differentiability issue.
SeqGAN \cite{yu2017seqgan} adopts a policy gradient RL approach to train a GAN, where the real-or-fake prediction score of the discriminator is used as a reward signal to train the generator.
However, this approach faces considerable training instability.
To alleviate the training instability, some works improve the guidance signals from the discriminator \cite{guo2018long},
the architecture design \cite{zhang2017adversarial}, or explore various training techniques such as RL-based text infilling in \cite{fedus2018maskgan}.
However, using GANs for text generation has not been very successful, due to the instability of GANs, coupled with the complications brought by the non-differentiable discrete text representations.

\noindent
\textbf{Autoregressive Transformers.}
GPT \cite{radford2018improving} is a seminal work that proposes a generative pre-training approach to train a Transformer-based language model, where a Transformer learns to autoregressively predict the next tokens during pre-training.
Then, GPT-2 \cite{radford2019language} shows that GPT's approach can be effective in even in a zero-shot setting.
XLNet \cite{yang2019xlnet} is designed to enable learning of bidirectional contexts during pretraining.
BART \cite{lewis2020bart} presents a bidirectional encoder with an autoregressive decoder, enabling pre-training with an arbitrary corruption of the original text.
T5 \cite{raffel2020exploring} proposes to treat every text processing task as a text generation task and investigates the transfer learning capabilities of the model in such a setting.
GPT-3 \cite{brown2020language} significantly scales up the size of language models, showing large improvements overall, and achieving good performance even in few-shot and zero-shot settings.
Notably, the original ChatGPT chatbot is mainly based on GPT-3, but with further fine-tuning via reinforcement learning from human feedback (RLHF) as performed in InstructGPT \cite{ouyang2022training}.
Gopher \cite{rae2021scaling} investigates the scaling of Transformer-based language models (up to a 280 billion parameter model) by evaluating across many tasks.
The training of Chinchilla \cite{hoffmann2022training} is based on an estimated compute-optimal frontier, where a better performance is achieved when the model is 4 times smaller than Gopher, while training on 4 times more data.
PaLM \cite{chowdhery2022palm} proposes a large and densely activated language model, allowing for efficient training across TPU pods.
Switch Transformer \cite{fedus2022switch} is a model with a large number of parameters, yet is sparsely activated due to its Mixture of Experts design where different parameters are activated for each input sample, and is thus computationally efficient despite its size.
LLaMA \cite{touvron2023llama} is trained by only using publicly available data, making it compatible with open-sourcing, and it is also efficient, outperforming GPT-3 on many benchmarks despite being much smaller.

\noindent
\textbf{Diffusion Models.}
Several works also explore adopting DMs to generate text, which holds the advantage of generating more diverse text samples, 
Diffusion-LM \cite{li2022diffusion} introduces a diffusion-based approach for text generation, which denoises random noise into word vectors in a non-autoregressive manner.
DiffuSeq \cite{gong2023diffuseq} also adopts a diffusion model for text generation, which shows high diversity during generation.

\vspace{-1.5mm}

\section{3D Modality} 
\label{sec:3D}

The 3D modality is another popular modality for generative methods.
The 3D generative methods generally fall into one of the following categories: 3D shapes, 3D scenes, 3D humans, or 3D motions. Since the challenges in each of these categories tend to be different, their methods and research directions tend to focus on different aspects as well. Therefore, we discuss each of these lines of works separately.
In this section, we first describe the AIGC methods for the 3D modality in this order: 3D shapes, 3D scenes, 3D humans, and 3D motions.

\subsection{3D Shapes Generation} 
\label{sec:3D_shapes}

3D shape generation methods aims to generate novel 3D shapes and objects.
The ability to swiftly generate 3D assets can be very useful, particularly in sectors like entertainment (e.g., VR/AR applications) or manufacturing, where it aids in rapid prototyping. 
Below, we first discuss the unconditional methods, followed by the conditional methods.

\subsubsection{Unconditional 3D Shapes Generation}

\begin{wrapfigure}{r}{0.30\textwidth}
\vspace{-5.5mm}
    \center
    \includegraphics[width=0.29\textwidth]{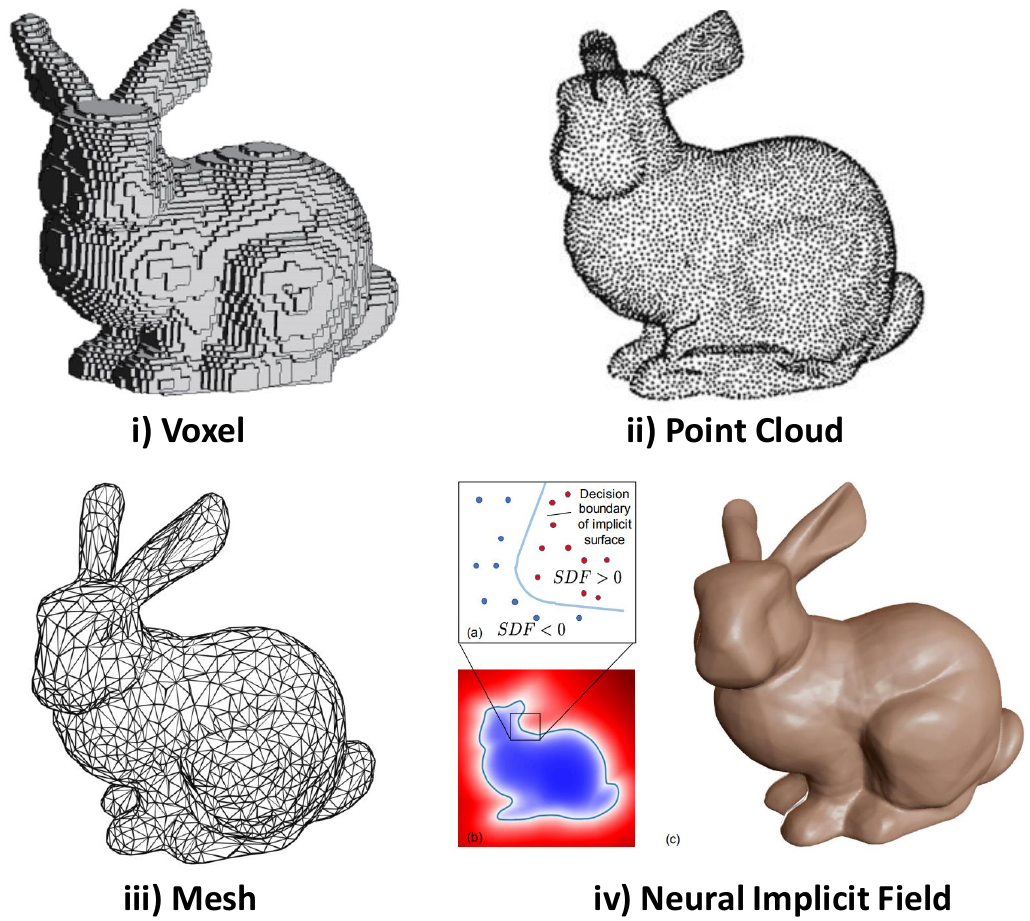} 
    \vspace{-4mm}
    \caption{\tiny
    Visualization of the Stanford Bunny as represented in (i) voxel; (ii) point cloud; (iii) mesh; (iv) neural implicit field. Images are obtained from \cite{camuffo2022recent,agarwal2009robust,rossi2021robust,park2019deepsdf}.
    }
    \label{figure:3d_bunny}
\vspace{-4mm}
\end{wrapfigure}

When generating 3D shapes, users can choose to generate the 3D shape in various 3D representations: voxel, point cloud, mesh, or neural implicit fields.
The choice of 3D representation is crucial as each 3D representation generally adopts different settings
and backbones, and each have their own characteristics, advantages and disadvantages.
A visualization of the various 3D representations is shown in Figure \ref{figure:3d_bunny}.
In practice, for many tasks a specific representation can be more suited than the other, where considerations can include the memory efficiency, ease of handling the representation, and the cost of obtaining supervision signals.
Below, we further categorize the unconditional 3D shape generation methods based on the generated output 3D data representation of each method.
Table \ref{table:3D_methods} reports the performance of some representative 3D shape methods.

\noindent
\textbf{Voxels.}
The voxel representation is a natural extension of the 2D image pixel representation into 3D, where the 3D space is divided into grids and each voxel stores some values (e.g., occupancy \cite{wu20153d} or signed distance values \cite{dai2017shape}).
As the voxel representation is the natural extension of the 2D pixel space, they can conveniently be processed with 3D CNNs, which is why earlier 3D generative works \cite{wu20153d,wu2016learning} tend to leverage the voxel representation.
However, generation via the 3D voxel representation can be computationally costly with poor memory efficiency, due to the cubic growth of the volume with increasing resolution.

3D ShapeNet \cite{wu20153d} explores using 3D voxel grids to represent 3D shapes which allows for 3D shape completion, and introduces ModelNet (a large dataset with 3D CAD models) to train it.
3D-GAN \cite{wu2016learning} proposes to leverage GANs to model and generate 3D objects, and designs a 3D GAN architecture to produce objects in a 3D voxel representation.
PlatonicGAN \cite{henzler2019escaping} aims to train a 3D generative model from a collection of 2D images, by introducing a 2D discriminator and rendering layers that connect between the 3D generator and 2D discriminator.
Some works propose to incorporate autoencoders into their work to have a better grasp of the high-level 3D structure through the autoencoder's latent code \cite{li2017grass,wu2019sagnet}, or to refine parts of the shape \cite{wang2018global}.
Yet, storing 3D voxel representation in memory can be inefficient, since the memory and computational requirements of handling the 3D voxel grid grow cubically with the resolution, which limits the 3D voxel output to low resolutions.
To overcome this, OctNet \cite{riegler2017octnet} leverages the octree data structure \cite{meagher1982geometric} -- a data structure with adaptive cell sizes -- and hierarchically partitions the 3D space into a set of unbalanced octrees, which allows for more memory and computations to be allocated to the regions that require higher resolutions.
OGN \cite{tatarchenko2017octree} presents a convolutional decoder architecture to operate on octrees, where intermediate convolutional layers can predict the occupancy value of each cell at different granularity levels, which is thus able to flexibly predict the octree's structure and does not need to know it in advance.
More recently, Octree Transformer \cite{ibing2023octree} introduces a Transformer-based architecture to effectively generate octrees via an autoregressive sequence generation approach.

In order to achieve better precision for 3D voxel representations efficiently, another approach is to use the 3D voxel grid to represent a signed distance function (SDF). In this representation, instead of storing occupancy values, each voxel instead stores the signed distance to the nearest 3D surface point, where the inside of the object has negative distance values and the outside has positive distance values.
3D-EPN \cite{dai2017shape} represents the shape's surface by storing the signed distances in the 3D voxel grid instead of occupancy values to perform 3D shape completion from partial, low-resolution scans.
OctNetFusion \cite{riegler2017octnetfusion} proposes a deep 3D convolutional architecture based on OctNets \cite{riegler2017octnet}, that can fuse depth information from different viewpoints to reconstruct 3D objects via estimation of the signed distance.
AutoSDF \cite{mittal2022autosdf} adopts a VAE-based approach and stores features in a 3D voxel grid, reconstructing the object by referring to a codebook learnt using the VAE which maps the features in each locality into more precise shapes.
This SDF approach is further extended into the continuous space with neural implicit functions in other works, and we discuss this later on in this section.

\begin{wraptable}{r}{0.6\textwidth}
\vspace{-4mm}
  \tiny
  \centering
  \caption{\tiny
  Comparison between representative 3D shape generative models on the chair and airplane shapes from ShapeNet \cite{chang2015shapenet}.    
  We report the Minimum Matching Distance (MMD) and Coverage (COV) scores \cite{achlioptas2018learning} based on Earth Mover's Distance (EMD) and Light Field Descriptor (LFD) \cite{chen2003visual} and 1-Nearest Neighbour Accuracy (1-NNA) metric \cite{lopez2017revisiting,yang2019pointflow} based on LFD.
  MMD is an indication of the fidelity of the generated samples (lower is better), COV measures the coverage and diversity of the generated samples (higher is better), while 1-NNA measures the similarity between the distribution of generated samples and the data distribution (lower is better).
  On the other hand, EMD and LFD are different ways of computing distances for these metrics.
  Note that MMD (EMD) scores are multiplied by 10\textsuperscript{2}.
  As there have been several different data pre-processing, post-processing approaches and implementations of the evaluation metrics (which lead to varying results), we also indicate which protocol each of the reported results use.
  We denote each protocol with a symbol, where each protocol comes from a specific paper, as follows: \dag \cite{shim2023diffusion}, \ddag \cite{luo2021diffusion}, * \cite{achlioptas2018learning}, \S \cite{zhou20213d}, \P \cite{ibing2023octree}, ** \cite{vahdat2022lion}, \dag\dag \cite{lyu2023controllable}, \S\S \cite{ibing2023octree}, \P\P \cite{gao2019sdm}. 
    A dash `-' indicates that the metric was not reported for that work.
  }
  \vspace{-0.25cm}
  \label{table:3D_methods}
  \scalebox{0.8}{
  \hspace{-0.4cm}
  \begin{tabular}{lc|ccccc|ccccc}    
    \toprule
    & & \multicolumn{5}{c|}{Chair}   & \multicolumn{5}{c}{Airplane}   \\    
    \cline{3-12}    
    Type & Method &  \multicolumn{2}{c|}{MMD ($\downarrow$)} & \multicolumn{2}{c|}{COV ($\uparrow$)} & 1-NNA ($\downarrow$)  &  \multicolumn{2}{c|}{MMD ($\downarrow$)} & \multicolumn{2}{c|}{COV ($\uparrow$)} & 1-NNA ($\downarrow$)  \\     
    \cline{3-12}
    & & EMD & \multicolumn{1}{c|}{LFD} & EMD & \multicolumn{1}{c|}{LFD} & EMD  & EMD & \multicolumn{1}{c|}{LFD} & EMD & \multicolumn{1}{c|}{LFD} & EMD \\
    \midrule
    \multirow{2}{*}{Voxel} & 3D-GAN \cite{wu2016learning} *  & 9.1 &  -  & 22.4 &  - &  - &  - &  -  & - & - &  - \\
    & 3D-GAN \cite{wu2016learning} \S\S  & - &  4365  & - &  25.07 &  - &  - &  -  & - & - &  - \\
    & GRASS \cite{li2017grass} \P\P   & 7.44 & - & 44.5 & - & -  & -  & -  & -  & -  & -  \\
    & G2L \cite{wang2018global} \P\P   & 6.82 & - & 83.4 & - & -  & -  & -  & -  & -  & -  \\
    & SAGNet \cite{wu2019sagnet} \P\P   & 6.08 & - & 74.3 & - & -  & -  & -  & -  & -  & -  \\
    & Octree Transformer \cite{ibing2023octree} \S\S  & - &  2958  & - &  76.47 &  - &  - &  3664  & - & 73.05 &  - \\
    \midrule
    \multirow{2}{*}{Point Cloud} & r-GAN \cite{achlioptas2018learning} *  & 12.3 &  - & 19.0 & -  & - & - & -  &  - &  - &  -  \\    
    & l-GAN (EMD) \cite{achlioptas2018learning} *  & 6.9 &  - & 57.1 & -  & - & - & - &  - &  - &  -  \\
    & PC-GAN \cite{li2019point} \S\S & - &  3143  & - &  70.06 &  - &  - &  3737  & - & 73.55 &  - \\
    &  PC-GAN \cite{li2019point} \ddag  & 31.04 &  - & 22.14 &  - & 100.00 &  18.10 &  - & 13.84 &  - & 98.52 \\ 
    &  GCN-GAN \cite{valsesia2019learning} \ddag  & 22.13 &  - & 35.09&  -  & 95.80 &  16.50 &  - & 18.62 &  - & 98.60 \\      
    &  Tree-GAN \cite{shu20193d} \ddag & 36.13 &  - & 6.77 &  - & 100.00 & 19.53 &  - & 8.40 &  - & 99.67 \\   
    &  PointFlow \cite{yang2019pointflow} \ddag  & 18.56 &  - & 43.38 &  - & 68.40 &  10.90 &  - & 44.65 &  - & 69.36 \\   
    &  ShapeGF \cite{cai2020learning} \ddag & 17.85 &  - & 46.71 &  - & 62.69 &   10.27 &  - & 47.12 &  - & 70.51 \\ 
    &  Luo et al. \cite{luo2021diffusion} \ddag  & 17.84 &  - & 47.52 &  - & 69.06 &   10.61 &  - & 45.47 &  - & 75.12 \\
    &  r-GAN \cite{achlioptas2018learning} \S   & - &  - & - &  - & 99.70   & - &  - & - &  - &   96.79 \\
    &  l-GAN (EMD) \cite{achlioptas2018learning} \S   & - &  - & - &  - & 64.65  & - &  - & - &  - &   76.91 \\
    &  PointFlow \cite{yang2019pointflow} \S & - &  - &  - & - & 60.57   & - & - &  - &  - &   70.74 \\
    &  SoftFlow \cite{kim2020softflow} \S   & - &  - &  - & - & 60.05   & - &  - &  - & - &   65.80 \\
    &  DPF-Net \cite{klokov2020discrete} \S  & - &  - &  - & - &  58.53   & - &  - &  - & - &   65.55 \\
    &  ShapeGF \cite{cai2020learning} \S  & - &  - &  - & - &  65.48 & - &  - &  - & - &   76.17 \\
    &  Luo et al. \cite{luo2021diffusion} \S  & - &  - &  - & - &  74.77  & - &  - &  - & - &   86.91 \\ 
    &  PVD \cite{zhou20213d} \S  & - & - &  - &  - &  53.32  & - &  - &  - & - &   64.81 \\
    &  Luo et al. \cite{luo2021diffusion} **  & - & - &  - &  - & 74.96  & - & - &  - &  - & 96.04   \\
    &  PVD \cite{zhou20213d} **  & - & -&  - &  -  & 57.90  & - & - &  - &  - & 56.06  \\
    &  PointFlow \cite{yang2019pointflow} \dag  & 4.21 &  - & 49.63 &  - & 74.74   & 1.67 &  - & 52.97 &  - & 62.50  \\
    &  Luo et al. \cite{luo2021diffusion} \dag  & 4.08 &  - & 41.65 &  - & 76.66   & 1.64 &  - & 52.23&  - &  63.37  \\
    &  PVD \cite{zhou20213d} \dag  & 3.56 &  - & 50.37 &  - &  53.03  & 1.55 &  - & 53.96 &  - &   52.72 \\
    &  Tree-GAN \cite{shu20193d}  \dag\dag  & 9.02 &  - & 48.89 &  - & 62.04  & 4.04 &  - & 43.81 &  -  & 71.78  \\
    &  ShapeGF \cite{cai2020learning} \dag\dag & 8.53 &  - & 50.96 &  - & 54.14 & 4.03 &  - & 41.58  &  - & 71.29  \\
    &  PVD \cite{zhou20213d} \dag\dag  & 8.38 &  - & 50.52 &  - & 52.36  & 4.29 &  - & 34.9 &  -  & 83.91  \\
    &  Luo et al. \cite{luo2021diffusion} \dag\dag  & 9.51 &  - & 47.27 &  - & 69.50 & 4.00 &  - & 48.76 &  -  & 68.81  \\
    &  SP-GAN \cite{li2021sp} \dag\dag & 9.67 &  - & 31.91 &  - & 75.85 & 4.03 &  - & 46.29  &  - & 70.42  \\    
     \midrule
    \multirow{2}{*}{Mesh} & SDM-NET \cite{gao2019sdm} \P\P   & 0.671 & - & 84.1 & - & -  & -  & -  & -  & -  & -  \\
    & LION \cite{vahdat2022lion} \S & - &  - &  - &  - & 52.34  &  - &  - &  -  & - & 53.70 \\
    & LION \cite{vahdat2022lion} ** & - &  - &  - &  - & 48.67 &  -  &  - &  - & - & 53.84 \\ 
    & SLIDE (centroid) \cite{lyu2023controllable} \dag\dag  &  8.49 &  - & 49.63 &  - & 51.18  & 3.77&  -  & 46.29 &  -  & 65.84  \\
    & SLIDE (random) \cite{lyu2023controllable} \dag\dag  &  8.63 &  - & 50.37&  -  & 53.10  & 3.81 &  - & 47.03  &  - & 67.82  \\    
     \midrule
    \multirow{2}{*}{Neural Fields} & IM-GAN \cite{chen2019learning} **  & - & - &  - &  - & 58.20  & - & - &  - &  - & 77.85\\
    & IM-GAN \cite{chen2019learning} \S\S  & - &  2893  & - &  75.44 &  - &  - &  3689  & - & 70.33 &  - \\
    & Grid IM-GAN \cite{ibing20213d} \S\S  & - &  2768  & - &  82.08 &  - &  - &  3226  & - & 81.58 &  - \\ 
    &  SDF-Diffusion \cite{shim2023diffusion} \dag  & 3.61 &  - & 49.31 &  - & 51.77   & 1.49  &  - & 55.20 &  - &  48.14 \\
    \bottomrule
  \end{tabular}
  }
\vspace{-3mm}
\end{wraptable}

\noindent
\textbf{Point Cloud.}
Point clouds are unordered sets of 3D points that represent a surface of a 3D object or scene.
One advantage of point clouds is that they are relatively scalable as compared to voxels, since they only explicitly encode the surface.
Furthermore, 3D point clouds are a popular representation because they can be conveniently obtained via depth sensors (e.g., LiDAR or the Kinect).
On the other hand, point clouds do not have a regular grid-like structure, which makes them challenging to be processed by CNNs.
Additionally, due to the point clouds being unordered, the proposed models that handle point clouds need to be permutation-invariant, i.e., invariant to the ordering between the points.
Another disadvantage is that point clouds do not have surfaces, so it can be difficult to directly add textures and lighting to the 3D object or scene.
To overcome this, 3D point clouds can be transformed to meshes through surface reconstruction techniques \cite{peng2021shape}.

Some works generate 3D point clouds via GANs.
Achlioptas et al. \cite{achlioptas2018learning} explore GANs for point cloud generation and improve the performance by training them in the fixed latent space of an autoencoder network.
PC-GAN \cite{li2019point} improves the sampling process and proposes a sandwiching objective.
GCN-GAN \cite{valsesia2019learning} investigates a GAN architecture with graph convolutions to effectively generate 3D point clouds by exploiting local topology.
Tree-GAN \cite{shu20193d} introduces a tree-structured graph convolutional network (GCN) which is more accurate and can also generate more diverse object categories.
However, the above GAN-based approaches tend to generate a fixed number of points and lack flexibility.
Therefore, PointFlow \cite{yang2019pointflow} explores an approach based on NFs, which can sample an arbitrary number of points for the generation of the 3D object.
DPF-Nets \cite{klokov2020discrete} propose a discrete alternative to PointFlow \cite{yang2019pointflow} which significantly improves computational efficiency.
SoftFlow \cite{kim2020softflow} estimates the conditional distribution of perturbed input data instead of directly learning the data distribution, which reduces the difficulty of generating thin structures.
ShapeGF \cite{cai2020learning} adopts a score-based approach, learning gradient fields to move an arbitrary number of sampled points to form the shape's surface.
More recently, Luo et al. \cite{luo2021diffusion} tackle 3D point cloud generation via DMs, which are simpler to train than GANs and continuous-flow-based models, and removes the requirement of invertibility for flow-based models.
PVD \cite{zhou20213d} is also a DM-based method, but adopts a point-voxel representation, which provides point clouds with a structured locality and spatial correlations that can be exploited.

\noindent
\textbf{Mesh.}
Polygonal meshes depict surfaces of 3D shapes by utilizing a set of vertices, edges, and faces.
Meshes are more memory-efficient and scalable than voxels as they only encode the surfaces of scenes, and are also more efficient than point clouds, where the mesh only requires three vertices and one face to represent a large triangular surface but many points need to be sampled for point clouds.
Furthermore, meshes offer an additional benefit by being well-suited for geometric transformations such as translation, rotation, and scaling of objects, which can be easily accomplished through straightforward vertex operations.
Besides, meshes can also encode textures more conveniently than voxels or point clouds.
Thus, meshes are commonly used in traditional computer graphics and related applications.
Nevertheless, dealing with and generating 3D meshes can present challenges, and the level of difficulty surpasses that of generating point clouds.
Similar to point clouds, the irregular data structure of meshes presents handling challenges. 
However, predicting both vertex positions and topology for meshes introduces further complexity, making it challenging to synthesize meshes with plausible connections between mesh vertices.
Thus, many works use point cloud as an intermediate representation, and transform point clouds to mesh via surface reconstruction techniques \cite{peng2021shape}.

SurfNet \cite{sinha2017surfnet} explores an approach and a network architecture to generate a 3D mesh represented by surface points, to reduce the computational burden from voxel representations.
SDM-NET \cite{gao2019sdm} introduces a network architecture to produce structured deformable meshes, that decomposes the overall shape into a set of parts and recovers the finer-scale geometry of each part by deforming a box, which is an approach that facilitates the editing and interpolation of the generated shapes. 
PolyGen \cite{nash2020polygen} presents an autoregressive approach with a Transformer-based architecture, which directly predicts vertices and faces sequentially while taking into account the long-range dependencies in the mesh.
LION \cite{vahdat2022lion} uses a hierarchical VAE with two diffusion models in the latent space to produce the overall shapes and the latent points, before using them to reconstruct a mesh.
SLIDE \cite{lyu2023controllable} designs a diffusion-based approach for mesh generation which employs point clouds as an intermediate representation via a cascaded approach, where one diffusion model learns the distribution of the sparse latent points (in a 3D point cloud format) and another diffusion model learns to model the features at latent points.

\textit{Mesh Textures.}
As 3D mesh is capable of representing surface textures, a line of work focuses on synthesizing textures of 3D surfaces.
Texture Fields \cite{oechsle2019texture} proposes a texture representation which allows for regressing a continuous function (parameterized by a neural network), and is trained by minimizing the difference between the pixel colors predicted by the Texture Field and the ground truth pixel colors.
AUV-Net \cite{chen2022auv} learns to embed 3D surfaces into a 2D aligned UV space, which allows textures to be aligned across objects and facilitates swapping or transfer of textures.
Texturify \cite{siddiqui2022texturify} generates geometry-aware textures for untextured collections of 3D objects without any explicit 3D color supervision.
GET3D \cite{gao2022get3d} aims to generate 3D meshes that are practical for real-world applications, which have detailed geometry and arbitrary topology, is a textured mesh, and is trained from 2D image collections.

\noindent
\textbf{Neural Implicit Fields.}
Neural implicit fields (or neural fields) are an implicit representation of 3D shapes where a neural network ``represents'' the 3D shape.
In order to observe the 3D shape, the 3D point coordinates are fed as inputs into the neural network, and the properties pertaining to each point can be predicted.
The properties predicted by the neural field tend to be either occupancy values \cite{chen2019learning,mescheder2019occupancy} or signed distance values \cite{park2019deepsdf}, from which we can infer the 3D shape.
For example, by querying the 3D space for occupancy values, we can infer which regions of the 3D space are occupied by the 3D shape.
Neural fields hold the advantage of being a continuous representation, which can potentially encode 3D shapes in high resolutions with high memory efficiency.
They are also flexible in encoding various topologies.
However, extracting the 3D surfaces from neural fields is typically slow, since it requires a dense evaluation of 3D points with a neural network (usually an MLP).
Furthermore, it can be challenging to train neural fields as there is a lack of ground truth data to effectively supervise the training, and it is also not straightforward to train generative models such as GANs to produce a neural field representation.
Moreover, since each neural field only captures one 3D shape, more effort is required to learn the ability to generalize and generate new shapes.
It can also be difficult to edit the 3D shape, as the information is stored implicitly as neural network weights.

\textit{Occupancy Fields.}
IM-GAN \cite{chen2019learning} and ONet \cite{mescheder2019occupancy} are the first to perform 3D shape generation while modeling the implicit field with a neural network, which assigns an occupancy value to each point.
LDIF \cite{genova2020local} decomposes a shape into a set of overlapping regions according to a structured implicit function template, improving the efficiency and generalizability of the neural field.
Peng et al. \cite{peng2020convolutional} introduce occupancy networks that leverage convolutions, which are more scalable than the fully-connected architecture often used in previous works.

\textit{Signed Distance Fields.}
However, predicting only occupancy values in an occupancy field (as described above) provides limited utility when compared to predicting the metric signed distance to the surface of the 3D object via a signed distance field (SDF).
For example, the signed distance values of the SDF can be used to raycast against surfaces to render the model, and its gradients can be used to compute surface normals.
DeepSDF \cite{park2019deepsdf} is the first work to use a neural field to predict the signed distance values to the surface of the 3D shape at each 3D point.
However, ReLU-based neural networks tend to face difficulties encoding high frequency signals such as textures, thus SIRENs \cite{sitzmann2020implicit} use periodic activation functions instead, in order to better handle high-frequency details.
Unlike previous methods that depend on a low-dimensional latent code to generalize across various shapes, MetaSDF \cite{sitzmann2020metasdf} aims to improve the generalization ability of neural SDFs via meta-learning.
Following developments in DMs, SDF-Diffusion \cite{shim2023diffusion} proposes a diffusion-based approach for generating high-resolution 3D shapes in the form of an SDF through iteratively increasing the resolution of the SDF via a denoising diffusion process.

\subsubsection{Conditional 3D Shapes Generation}
\noindent
\textbf{3D Shape Editing} is where a model aims to edit a 3D shape conditioned on the user inputs.
Some works explore the manipulation and editing of 3D shapes through adjusting values of primitives \cite{hao2020dualsdf}.
Other works also explore texture transfer through generating dense correspondence among shapes with neural fields \cite{deng2021deformed}.
NeuMesh \cite{yang2022neumesh} uses a mesh-based representation for more fine-grained editing of the mesh's geometry and texture, avoiding the flexibility issues associated with neural fields.

\vspace{2mm}

\subsection{Novel View Synthesis for 3D Scenes}
\label{sec:3D_scene_NVS}

Along with the progress in 3D shape reconstruction, there has also been more attention and interest on 3D scenes, which can involve one or multiple objects and the background.
The mainstream generative approach involving 3D scenes is to explicitly or implicitly encode a 3D scene representation (i.e., via a voxel-based representation or a neural implicit representation), which allows for synthesis of images from novel views when required.
Due to the need for encoding complete scenes, this task tends to be much more challenging than 3D shape generation.
Another difficulty arises with how to implicitly encode the 3D scene, since the rendering of 3D scenes involves the generation of color, texture and lighting, which are challenging elements that now need to be encoded, while occupancy fields and signed distance fields as introduced in Section \ref{sec:3D_shapes} for 3D shape representations do not naturally encode for colour.

\subsubsection{Unconditional Novel View Synthesis for 3D Scene.}
There are several lines of work for novel view synthesis, and they can be categorized according to the representations they operate on, which we discuss below.
Table \ref{table:novelview_methods} shows a quantitative comparison  of these  methods.

\begin{wraptable}{r}{0.4\textwidth}
\vspace{-5mm}
  \footnotesize
  \centering
  \caption{
  Comparison between 3D scene novel view synthesis models.
  Results are reported on FFHQ \cite{karras2019style} at $256\times256$ and $512\times512$, Cats \cite{zhang2008cat} at $256\times256$ and CARLA \cite{dosovitskiy2017carla,schwarz2020graf} at $128\times128$.  
  We report FID scores \cite{heusel2017gans} to measure the quality of images rendered from the scene (lower is better).  
  We report results from various protocols (each denoted with a symbol), where each protocol comes from a specific paper, as follows:   
    \S \cite{schwarz2022voxgraf}, \ddag \cite{or2022stylesdf}, * \cite{deng2022gram}.
    A dash `-' indicates that the metric was not reported for that work.
  }
  \vspace{-0.35cm}
  \label{table:novelview_methods}
  \scalebox{0.95}{
  \hspace{-0.4cm}
  \begin{tabular}{lcccccccc}
    \toprule
    Method & FFHQ 256\textsuperscript{2} & FFHQ 512\textsuperscript{2} & Cat 256\textsuperscript{2} & Carla 128\textsuperscript{2}\\
    \midrule
    GIRAFFE \cite{niemeyer2021giraffe} & 32 & - &  33.39 & -  \\
    $\pi$-GAN \cite{chan2021pi}  & - & - & - & 29.2  \\  
    StyleNeRF \cite{gu2022stylenerf}  & 8.00 & 7.8  & 5.91 & - \\  
    StyleSDF \cite{or2022stylesdf} & 11.5 & 11.19 & - & -  \\
    EG3D \cite{chan2022efficient}  & 4.8  & 4.7  & - &  - \\
    VolumeGAN \cite{xu20223d} & 9.1 & - & - & 7.9 \\
    MVCGAN \cite{zhang2022multi} & 13.7 & 13.4 & 39.16 & -  \\
    GIRAFFE-HD \cite{xue2022giraffe}  & 11.93 & - & 12.36 & -  \\
    \midrule
    GRAF \cite{schwarz2020graf} * &  73.0 & - & 59.5 & 32.1  \\
    $\pi$-GAN \cite{chan2021pi} * &  55.2 & - & 53.7 & 36.0  \\
    GIRAFFE \cite{niemeyer2021giraffe} * &  32.6 & - & 20.7 & 105  \\
    GRAM \cite{deng2022gram} * &  17.9 & - & 14.6 & 26.3  \\
    \midrule
    HoloGAN \cite{nguyen2019hologan} \ddag & 90.9 & - &  - & -  \\
    GRAF \cite{schwarz2020graf} \ddag & 79.2 & - &  - & -  \\
    $\pi$-GAN \cite{chan2021pi} \ddag & 83.0 & - &  - & -  \\
    GIRAFFE \cite{niemeyer2021giraffe} \ddag & 31.2 & - &  - & -  \\
    StyleSDF \cite{or2022stylesdf} \ddag & 11.5 & - &  - & -  \\
    \midrule
    GRAF \cite{schwarz2020graf} \S & 71 & - & -  & 41  \\  
    GIRAFFE \cite{niemeyer2021giraffe} \S & 31.5 & - & - &  -  \\
    $\pi$-GAN \cite{chan2021pi} \S & 85 & - & - & 29.2  \\
    GOF \cite{xu2021generative} \S & 69.2 & - & - & 29.3  \\
    GRAM \cite{deng2022gram} \S & 17.9  & - & - & 26.3  \\
    VoxGRAF\cite{schwarz2022voxgraf} \S & 9.6 & - & - &  6.7 \\
    \bottomrule
  \end{tabular}
  }
\vspace{-4mm}
\end{wraptable}

\noindent
\textbf{Voxel-based Representations.}
One approach for novel view synthesis stores the 3D scene information (geometry and appearance) as features in a voxel grid, which can be rendered into images.
DeepVoxels \cite{sitzmann2019deepvoxels} introduces a deep 3D voxel representation where features are stored in a small 3D voxel grid, which can be rendered by a neural network (i.e., neural rendering) to produce 3D shapes.
HoloGAN \cite{nguyen2019hologan} has an architecture that enables direct manipulation of view, shape and appearance of 3D objects and can be trained using unlabeled 2D images only, while leveraging a deep 3D voxel representation \cite{sitzmann2019deepvoxels}.
BlockGAN \cite{nguyen2020blockgan} also generates 3D scene representations by learning from unlabeled 2D images, while adding compositionality where the individual objects can be added or edited independently.
HoloDiffusion \cite{karnewar2023holodiffusion} introduces a DM-based approach that can be trained on posed images without access to 3D ground truth, where the DM generates a 3D feature voxel grid, which is rendered by a rendering function (MLP) to produce the 2D images.
Besides, sparse voxel-based representations \cite{schwarz2022voxgraf} have been proposed to improve the optimization efficiency.
For instance, VoxGRAF \cite{schwarz2022voxgraf} proposes a 3D-aware GAN which represents the scene with a sparse voxel grid to generate novel views.

\noindent
\textbf{Neural Radiance Fields (NeRFs).}
Another approach involves implicitly encoding the 3D scene as a neural implicit field, i.e., where a neural network represents the 3D scene.
In order to additionally output appearance on top of shapes, NeRF \cite{mildenhall2020nerf} presents neural radiance fields that encodes color and density given each point and the viewing angle, which can produce a 3D shape with color and texture after volume rendering.
NeRF-based scene representations have become popular as they are memory efficient and scalable to higher resolutions.

GRAF \cite{schwarz2020graf} is the first work to generate radiance fields via an adversarial approach, where the discriminator is trained to predict if the rendered 2D images from the radiance fields are real or fake.
$\pi$-GAN \cite{chan2021pi} improves the expressivity of the generated object by adopting a network based on SIREN \cite{sitzmann2020implicit} for the neural radiance field, while leveraging a progressive growing strategy.
GOF \cite{xu2021generative} combines NeRFs and occupancy networks to ensure compactness of learned object surfaces.
Mip-NeRF \cite{barron2021mip} incorporates a mipmapping from computer graphics rendering pipelines to improve efficiency and quality of rendering novel views.
GRAM \cite{deng2022gram} confines the sampling and radiance learning to a reduced space (a set of implicit surfaces), which improves the learning of fine details.
MVCGAN \cite{zhang2022multi} incorporates geometry constraints between viewsn.
StyleNeRF \cite{gu2022stylenerf} and StyleSDF \cite{or2022stylesdf} incorporate a style-based generator \cite{karras2020analyzing} to improve the rendering efficiency while generating high-resolution images.
Mip-NeRF 360 \cite{barron2022mip} builds upon Mip-NeRF \cite{barron2021mip} to extend to unbounded scenes, where the camera can rotate 360 degress around a point.
TensoRF \cite{chen2022tensorf} models the radiance field of a scene as a 4D tensor and factorizes the tensors, leading to significant gains in reconstruction speed and rendering quality.
An orthogonal line of work explores the compositional generation of scenes for more control over the generation process.
GIRAFFE \cite{niemeyer2021giraffe} incorporates a compositional 3D scene representation to control the image formation process with respect to the camera pose, object poses and objects' shapes and appearances.
GIRAFFE HD \cite{xue2022giraffe} extends GIRAFFE to generate high-quality high-resolution images by leveraging a style-based \cite{karras2020analyzing} neural renderer while generating the foreground and background independently to enforce disentanglement.

\noindent
\textbf{Hybrid 3D Representations.}
Hybrid representations often combine explicit representations (e.g., voxels, point clouds, meshes) with implicit representations (e.g., neural fields).
Such hybrid representations aim to capitalize on the strengths of each type of representation, e.g., explicit control over geometry afforded by explicit representations, as well as memory efficiency and flexibility of the implicit representations.

\textit{Voxels and Neural Fields.}
One line of works combine voxels with neural fields.
NSVF \cite{liu2020neural} defines a set of neural fields which are organized in a sparse voxel-based structure, where the neural field in each cell models the local properties in that cell.
SNeRG \cite{hedman2021baking} is a representation with a sparse voxel grid of diffuse color, volume density and features, where images can be rendered by an MLP in real-time.
VolumeGAN \cite{xu20223d} represents 3D scenes in a hybrid fashion, with explicit voxel grids and NeRF-like implicit feature fields.
DiffRF \cite{muller2023diffrf} generates the neural radiance fields via a DM on a voxel grid and a volumetric rendering loss, which leads to efficient, realistic and view consistent renderings.

\textit{Triplane.}
Another promising hybrid approach is the triplane representation introduced by EG3D \cite{chan2022efficient}, where the scene information is stored in three axis-aligned orthogonal feature planes, and a decoder network takes in aggregated 3D features from the three planes to predict the color and density values at each point.
This representation is efficient, as it can keep the decoder small by leveraging explicit features from the triplanes, while the triplanes scale quadratically instead of cubically (as compared to dense voxels) in terms of the memory requirement.
Shue et al. \cite{shue20233d} generate neural fields in a triplane representation with a DM for generation of high-fidelity and diverse scenes.

\subsubsection{Conditional Novel View Synthesis for 3D Scene.}
\textbf{3D Scene Editing.}
A major line of work aims to edit 3D scene radiance fields.
Editing-NeRF \cite{liu2021editing} proposes to edit the NeRF based on a user's coarse scribbles, which allows for color modification or removing of certain parts of the shape.
ObjectNeRF \cite{yang2021learning} learns an object-compositional NeRF to duplicate, move, or rotate objects in the scene.
NeRF-editing \cite{yuan2022nerf} explores a method to edit a static NeRF to perform user-controlled shape deformation, where the user can edit a reconstructed mesh.
Deforming-NeRF \cite{xu2022deforming} enables free-form radiance field deformation by extending cage-based deformation of meshes to radiance field deformation, which allows for explicit object-level scene deformation or animation.

\subsection{3D Human Generation}
\label{sec:3D_human}

Besides generation of 3D shapes and scenes, generation of 3D humans is also an important task.
Different from generation of shapes which tend to be static and inanimate, the ultimate goal of 3D human generation is to synthesize animatable humans with movable joints and non-rigid components (such as hair and clothing), which further increases the difficulty of the task.
Below, we first discuss methods for unconditional 3D human generation, followed by the conditional methods.

\subsubsection{Unconditional 3D Human Generation}
In general, the work on generating 3D human avatars can be split into the full body and head, which adopt different approaches. We discuss them separately below.

\noindent
\textbf{3D Avatar Body.}
3D human avatar generation aims to generate diverse 3D virtual humans with different identities and shapes, which can take arbitrary poses.
This task can be very challenging, as there can be many variations in clothed body shapes and their articulations can be very complex.
Furthermore, since 3D avatars should ideally be animatable, 3D shape generation methods are typically not easily extended to non-rigid clothed humans.
Early works (e.g., SMPL \cite{loper2015smpl}) explore generating 3D human mesh via human parametric models, which express human body shapes in terms of a relatively small set of parameters that deform a template human mesh.
These models conveniently synthesize human shapes as users only need to fit/regress values for a small set of parameters.
Some works also explore a similar approach for modeling clothed humans \cite{ma2020learning}.
However, such human parametric models are not able to capture many finer and personalized details due to the fixed mesh topology and the bounded resolution.
To overcome these limitations, a line of works propose to generate 3D human avatars by adopting implicit non-rigid representations \cite{saito2020pifuhd}.
gDNA \cite{chen2022gdna} proposes a method to generate novel detailed 3D humans with a variety of garments, by predicting a skinning field and a normal field (i.e., using an MLP to predict the surface normals), and is trained using raw posed 3D scans of humans.
AvatarGen \cite{zhang2022avatargen} leverages an SDF on top of a SMPL human prior, and enables disentangled control over the human model's geometry and appearance.
EVA3D \cite{hong2023eva3d} is a compositional human NeRF representation that can be trained using sparse 2D human image collections, which is efficient and can render in high-resolution.

\noindent
\textbf{3D Avatar Head.}
On the other hand, 3D avatar head generation aims to generate a 3D morphable face model with fine-grained control over facial expressions.
It is challenging to produce realistic 3D avatar heads, and it is even more difficult to model the complex parts such as hair.
Traditional approaches model facial appearance and geometry based on the 3D Morphable Models (3DMM) \cite{blanz1999morphable} which is a parametric model that simplifies the face modeling to fitting values in a linear subspace.
Subsequently, many variants have been proposed, including multilinear models \cite{vlasic2006face}, full-head PCA models \cite{li2017learning} and fully articulated head models \cite{paysan20093d}.
Several works \cite{tran2019learning} have explored deep generative methods to generate an explicit 3D face model, however these methods tend to produce 3D avatar heads that lack realism.
To improve the realism, many recent methods proposed implicit methods for unconditional generation of head avatars, with some exploring an SDF-based approach \cite{yenamandra2021i3dmm}, or a NeRF-based approach \cite{hong2022headnerf}.
By leveraging upon its implicit representation, \cite{yenamandra2021i3dmm} is also able to model hair.

\subsubsection{Conditional 3D Human Generation}
\textbf{3D Avatar Body Generation Conditioned on 3D scans.}
Many works aim to produce an animatable 3D avatar from 3D scans.
Traditionally, articulated deformation of a given 3D human mesh is often performed with the classic linear blend skinning algorithm.
However, the deformation is simple and cannot produce pose-dependent variations, which leads to various artifacts \cite{lewis2000pose}.
Many methods have been proposed to tackle these issues, such as dual quaternion blend skinning \cite{kavan2008geometric} and multi-weight enveloping \cite{wang2002multi}.
Recently, many works aim to produce an animatable 3D avatar from 3D scans via implicit representations of the human body, which are resolution-independent, smooth and continuous.
NASA \cite{deng2020nasa} learns an occupancy field to model humans as a collection of deformable components.
Some works such as SNARF \cite{chen2021snarf} also leverage occupancy field representations, and can generate 3D animated human characters which generalize well to unseen poses.
Some other works \cite{saito2021scanimate} explore an SDF-based implicit representation for improved generation quality.
Some works (e.g., Peng et al. \cite{peng2021animatable}, Neural Actor \cite{liu2021neural}) also propose deformable and animatable NeRFs for synthesizing humans from novel poses in novel views, which handle colors and textures effectively with the NeRF.

\subsection{3D Motion Generation}
\label{sec:3D_motion}

Besides generating 3D humans, many AIGC works also generate 3D human motions to drive the movements of 3D humans.
The most common representation used for 3D motion generation is the 3D skeleton pose -- is a simple yet effective representation that captures the human joint locations.
After generating the 3D skeleton pose sequences, they can be utilized to generate simple moving human meshes, or to animate a rigged avatar (which can be generated by methods in Section \ref{sec:3D_human}).
In 3D motion generation, it can be challenging to synthesise long-term motions that are realistic and coherent. It is also difficult to capture multiple diverse yet plausible motions from a starting point.

\subsubsection{Unconditional 3D Motion Generation}

For unconditional synthesis of human motions,
Holden et al. \cite{holden2016deep} introduce a deep learning framework for motion synthesis with an autoencoder that learns the motion manifold.
CSGN \cite{yan2019convolutional} introduces a GAN-based approach to generate long motion sequences with meaningful and continuous actions.
MoDi \cite{raab2023modi} proposes an encoder architecture to learn a structured, highly semantic and clustered latent space, which facilitates motion generation via a GAN-based approach.
MDM \cite{tevet2023human} adopts a DM for generating human motion, which frees the assumption of the one-to-one mapping of autoencoders and can express many plausible outcomes.

\subsubsection{Conditional 3D Motion Generation}
3D motions are often generated with two types of conditional information: prefix frames and on action classes. We discuss both scenarios below.

\noindent
\textbf{Conditioned on prefix frames.} 
In conditional 3D motion generation, one line of works aims to generate the motion following a given set of prefix frames, which is also known as motion prediction.
These prefix 3D skeleton poses can be obtained using depth sensors such as the Kinect, or extracted from an input image or video with a 3D pose estimation algorithm \cite{gong2023diffpose,foo2023unified}.
Earlier works rely on recurrent neural networks (RNNs) \cite{fragkiadaki2015recurrent} for learning to autoregressively predict and generate future motion.
In order to improve the modeling of spatio-temporal information and long-term motion generation, other architectures have also been investigated, such as CNNs \cite{li2018convolutional}, GCNs \cite{mao2019learning}, Transformers \cite{cai2020learningprog} and MLPs \cite{guo2022back}.
Furthermore, in order to predict multiple diverse yet plausible future poses, 
some other works also adopt probabilistic elements such as VAEs \cite{cai2021unified}, GANs \cite{barsoum2018hp}, or NFs \cite{yuan2020dlow}.

\noindent
\textbf{Conditioned on actions.}
Another line of works focus on generating motions of a specified action class, where an initial pose or sequence is not required.
Generally, this can be achieved by feeding the target action class as input to the motion generation model during training, and have class-annotated clips as supervision.
Action2Motion \cite{guo2020action2motion} first introduces a Lie Algebra-based VAE framework to generate motions from desired action categories to cover a broad range of motions.
ACTOR \cite{petrovich2021action} is a Transformer VAE that can synthesize variable-length motion sequences of a given action.
Some approaches based on GANs have also been proposed to generate more diverse motions \cite{degardin2022generative}.
Furthermore, some works \cite{maheshwari2022mugl} also perform multi-person motion generation.

\subsection{Summary of 3D Modality}
In summary, in this section we reviewed the AIGC methods for the 3D modality.
Since the 3D field is broad and contains many subfields, we first categorize the works according to the various subfields -- 3D shapes, 3D scene novel view synthesis, 3D humans, 3D motion -- where each subfield tends to have a different focus.
First, for 3D shape generation (Section \ref{sec:3D_shapes}), we first introduce the unconditional generation methods across representations such as voxels \cite{wu20153d}, point clouds \cite{achlioptas2018learning}, mesh \cite{sinha2017surfnet}, and neural implicit fields \cite{chen2019learning}, followed by discussing some conditional 3D shape generation works. 
Next, we discuss 3D scene novel view synthesis methods (Section \ref{sec:3D_scene_NVS}), including voxel-based approaches \cite{sitzmann2019deepvoxels}, NeRF-based approaches \cite{mildenhall2020nerf}, and hybrid approaches \cite{liu2020neural}.
We also further discuss editing of these 3D scene representations.
Then, we discuss about 3D humans (Section \ref{sec:3D_human}), where most methods target the 3D avatar body \cite{zhang2022avatargen} or head \cite{cheng2019meshgan}. 
We also present the conditional generation of avatars from 3D scans of real people \cite{mihajlovic2021leap}. 
Lastly, we review 3D motion generation (Section \ref{sec:3D_motion}) that can animate 3D human avatars. Besides covering the unconditional approaches \cite{holden2016deep}, we also explore the generation of 3D motion conditioned on prefix frames \cite{fragkiadaki2015recurrent} and conditioned on actions \cite{guo2020action2motion}.

\vspace{2mm}

\section{Audio Modality}
\label{sec:audio}

\begin{wraptable}{r}{0.45\textwidth}
\vspace{-6mm}
  \tiny
  \centering
  \caption{\tiny
  Comparison between audio generation models.
  We report results for unconditional generation based on raw waveform on the Speech Commands \cite{warden2018speech} dataset.
  For this task, we report the Fr\'echet Inception Distance (FID), Inception Score (IS) and 5-scale Mean Opinion Score (MOS) metrics.
  FID measures the similarity between real and generated audio, IS measures the diversity of generated audio and whether it can be clearly determined by a classifier, while MOS evaluates the quality according to human subjects.
  Results for neural vocoding (where mel-spectrograms are given) are reported on proprietary dataset of \cite{chen2021wavegrad} and the LJ speech \cite{ito2017lj} dataset, where we report the 5-scale Mean Opinion Score (MOS) metric.
   We report the results from two different protocols, which have been denoted with different symbols: \dag \cite{chen2021wavegrad}, \ddag \cite{kong2021diffwave}.
  }
  \vspace{-0.35cm}
  \label{table:audio_methods}
  \scalebox{0.95}{
  \hspace{-0.4cm}
  \begin{tabular}{lcccccccc}
    \toprule
    Method & \multicolumn{3}{c}{Speech Commands} & LJ & \cite{chen2021wavegrad}'s dataset  \\ 
     & FID ($\downarrow$) & IS ($\uparrow$) & MOS ($\uparrow$) &  MOS ($\uparrow$) &  MOS ($\uparrow$) \\
    \midrule
    WaveRNN \cite{kalchbrenner2018efficient} \dag & - & - & - & 4.49$\pm$0.05  & 4.49$\pm$0.04   \\
    Parallel WaveGAN \cite{yamamoto2020parallel} \dag &  - & - & - &-
 & 3.92$\pm$0.05   \\
    MelGAN \cite{kumar2019melgan} \dag & - & - & - &-  &  3.95$\pm$0.06   \\
    Multi-band MelGAN \cite{yang2021multi} \dag & - & - & - & - & 4.10$\pm$0.05   \\
    GAN-TTS \cite{binkowski2020high} \dag &  - & - & - &-  &4.34$\pm$0.04   \\
    WaveGrad Base \cite{chen2021wavegrad} \dag & - & - & - &4.35$\pm$0.05  &  4.41$\pm$0.03   \\
    WaveGrad Large \cite{chen2021wavegrad} \dag & - & - & - &4.55$\pm$0.05  &  4.51$\pm$0.04   \\
    \midrule
    WaveNet-128 \cite{oord2016wavenet} \ddag & 3.279 & 2.54 & 1.34$\pm$0.29 & 4.43$\pm$0.10 &  - \\
    WaveNet-256 \cite{oord2016wavenet} \ddag & 2.947 & 2.84 & 1.43$\pm$0.30 & - &  - \\
    WaveGAN \cite{donahue2018adversarial} \ddag & 1.349 & 4.53 & 2.03$\pm$0.33 & - & - \\
    ClariNet \cite{ping2019clarinet} \ddag & - & - & - & 4.27$\pm$0.09 & - \\
    WaveGlow \cite{prenger2019waveglow} \ddag & - & - & - & 4.33$\pm$0.12 & - \\
    WaveFlow-64 \cite{ping2020waveflow} \ddag & - & - & - & 4.30$\pm$0.11 & - \\
    WaveFlow-128 \cite{ping2020waveflow} \ddag & - & - & - & 4.40$\pm$0.07 & - \\
    DiffWave Base \cite{kong2021diffwave} \ddag & - & - & - & 4.38$\pm$0.08 & - \\
    DiffWave Large \cite{kong2021diffwave} \ddag & - & - & - & 4.44$\pm$0.07 & - \\
    DiffWave \cite{kong2021diffwave} \ddag & 1.287 & 5.30 & 3.39$\pm$0.32 & - & - \\
    \bottomrule
  \end{tabular}
  }
\vspace{-9mm}
\end{wraptable}

Many AIGC methods also aim to generate audio, which facilitates the creation of voiceovers, music, and other sound effects. They can also be useful in text-to-speech applications, e.g., assistive technology and entertainment purposes.
However, it can be challenging to generate audio realistically, including the pitch and timbre variations.
In the context of speech, it can be difficult to capture the emotional and tonal variations, and it is also difficult to generate speech based on a given identity.
In various applications, inference speed of generating the audio is also important.
Furthermore, the lack of large-scale open-source audio datasets remains an issue for the audio modality, which may explain why there seems to be fewer representative works in the audio modality as compared to other modalities (see discussion in Section \ref{sec:papers_discussion}).

Audio generation models are mostly trained to generate audio while conditioned on audio inputs (e.g., speech sample of a target).
Therefore, similar to the text modality (Section \ref{sec:text_modality}), in this section we categorize the audio AIGC methods as unconditional methods to maintain the consistency of the taxonomy levels and the terminology with other sections.
Yet, we note that these can often be used as conditional methods as well.

\subsection{Unconditional Audio Generation}
Audio generation works are split into two main lines of work: speech generation and audio generation. We discuss them separately below.

\noindent
\textbf{Speech.}
Here, we discuss the unconditional speech generation works.
In general, these methods can also be slightly modified to apply to the conditional settings (e.g., text-to-speech that is discussed in Section \ref{sec:crossmodal_audio}).
WaveNet \cite{oord2016wavenet} is the first deep generative model for audio generation.
It is an autoregressive model that predicts the next audio sample conditioned on previous audio samples, which are built based on causal convolutional layers.
Subsequently, some other works such as SampleRNN \cite{mehri2017samplernn} and WaveRNN \cite{kalchbrenner2018efficient} also adopt an autoregressive approach while improving the modeling of long-term dependencies \cite{mehri2017samplernn} and reducing the sampling time \cite{kalchbrenner2018efficient}.
VQ-VAE \cite{van2017neural} leverages a VAE to encode a latent representation for audio synthesis.
Some works \cite{rezende2015variational,kingma2018glow} also explore a NF-based approach to generate audio. 
Notably, Parallel WaveNet \cite{oord2018parallel} distills a trained WaveNet model into a
flow-based IAF \cite{kingma2016improved} model for the sake of efficient training and sampling.
WaveGlow \cite{prenger2019waveglow} extends the flow-based Glow \cite{kingma2018glow} with WaveNet for efficient and high-quality audio synthesis.
WaveFlow \cite{ping2020waveflow} builds upon WaveGlow, further improving the fidelity and synthesis speed.
ClariNet \cite{ping2019clarinet} distills a flow-based Gaussian IAF model from the autoregressive WaveNet.
WaveGAN \cite{donahue2018adversarial} and MelGAN \cite{kumar2019melgan} adopt a GAN-based approach to generating audio, with MelGAN providing significant improvements in generation speed.
Parallel WaveGAN \cite{yamamoto2020parallel} further improves the speed of audio waveform generation.
Diffusion-based audio generation has also been explored in DiffWave \cite{kong2021diffwave} and WaveGrad \cite{chen2021wavegrad}, which show strong performance.

\noindent
\textbf{Music.}
Music generation is another widely explored audio generation task.
Different from speech, music can have multiple tracks which represent different instruments.
There are also many styles of music (e.g., pop, classical) and different emotions/themes, which can be challenging to keep consistent over long ranges.
Although there are earlier attempts at music generation \cite{eck2002first}, producing complex multi-track music has largely been facilitated by deep generative methods.
The first attempts \cite{chu2016song,huang2016deep} to generate music with deep learning adopted recurrent neural networks (RNNs) to recurrently generate new notes.
Subsequently, Engel et al. \cite{engel2017neural} approach music synthesis with a WaveNet-based autoencoder, and also introduce the NSynth dataset with annotated notes of various instruments for training, which enables expressive and realistic generation of instrument sounds.
MiDiNet \cite{yang2017midinet} aims to generate melodies from scratch or from a few given bars of music by taking a GAN-based approach, and can be expanded to generate music with multiple tracks.
DeepBach \cite{hadjeres2017deepbach} aims to generate polyphonic music, with a specific focus on hymn-like pieces.
Dieleman et al. \cite{dieleman2018challenge} do not handle the symbolic representations of music (e.g., scores, MIDI sequences) and handle raw audio instead, which improves generality and is more able to capture the precise timing, timbre and volume of the notes.
Jukebox \cite{dhariwal2020jukebox} generates music with singing with a multi-scale VQ-VAE encoder and an autoregressive Transformer decoder, which can generate high-fidelity and diverse songs that are coherent for up to several minutes.
It can also be conditioned on the artist and genre to control the vocal style.

\section{Summary of Single-Modality AIGC Methods}
In summary, in Sections \ref{sec:image}-\ref{sec:audio}, we have covered the single-modality AIGC methods across modalities such as image, video, text, 3D, and audio modalities. 
For each modality, we discussed the unconditional setting, which generally involves the foundational techniques of the modality.
We also covered the mainstream and common conditional generation settings, which tend to focus on how to best incorporate conditional information to achieve the tasks, e.g., editing.
Specifically, we also discuss in-depth the foundational advancement in generative techniques (e.g., GANs, VAEs, NFs and DMs) under the image modality. 
This is because they are often developed by running experiments with images, before being adopted across other modalities.
Under the large 3D modality section, we also discussed various 3D modalities: 3D shape, 3D scene, 3D human, and 3D motion, where we also go in-depth in each of the respective sections in terms of their representations and settings, which are quite varied.
See Figure \ref{figure:taxonomy_combined} to view the broad organization and taxonomy of the single-modality methods discussed in this paper.

\section{Cross-Modality Image Generation}
\label{sec:crossmodal_image}

In this section, we discuss cross-modality image generation, where images are generated using conditioning information from other modalities. 
In particular, text-conditioned image generation is a fundamental task which is the setting for many foundational developments in cross-modality generation. 
Thus, the works reviewed in this section also often the foundation upon which other cross-modality methods are built.

Here, we briefly discuss the structure and focus of the sections discussing cross-modality methods (Section \ref{sec:crossmodal_image} to Section \ref{sec:crossmodal_audio}).
In the cross-modality setting, it is crucial that we learn the interaction between multiple modalities, which enables us to control the generation process via the inputs from another modality.
To this end, throughout this section and the sections discussing cross-modality generation, we focus on describing how the different approaches bridge the two different modalities, e.g., whether they tend to adopt pre-trained text encoders or whether they operate upon a shared cross-modality embedding space.
Throughout this section and subsequent cross-modality sections, the general organization is as such: in each section we discuss the cross-modality generation methods for generating a particular modality, where the methods are further categorized according to the type of input conditioning modality.
Then, we discuss the various approaches while focusing on how they effectively bridge the cross-modality gap.
Please refer to the taxonomy in Figure \ref{figure:taxonomy_combined2} for more information.

\subsection{Conditioned on Text}

\begin{wrapfigure}{r}{0.52\textwidth}
    \center
    \includegraphics[width=0.53\textwidth]{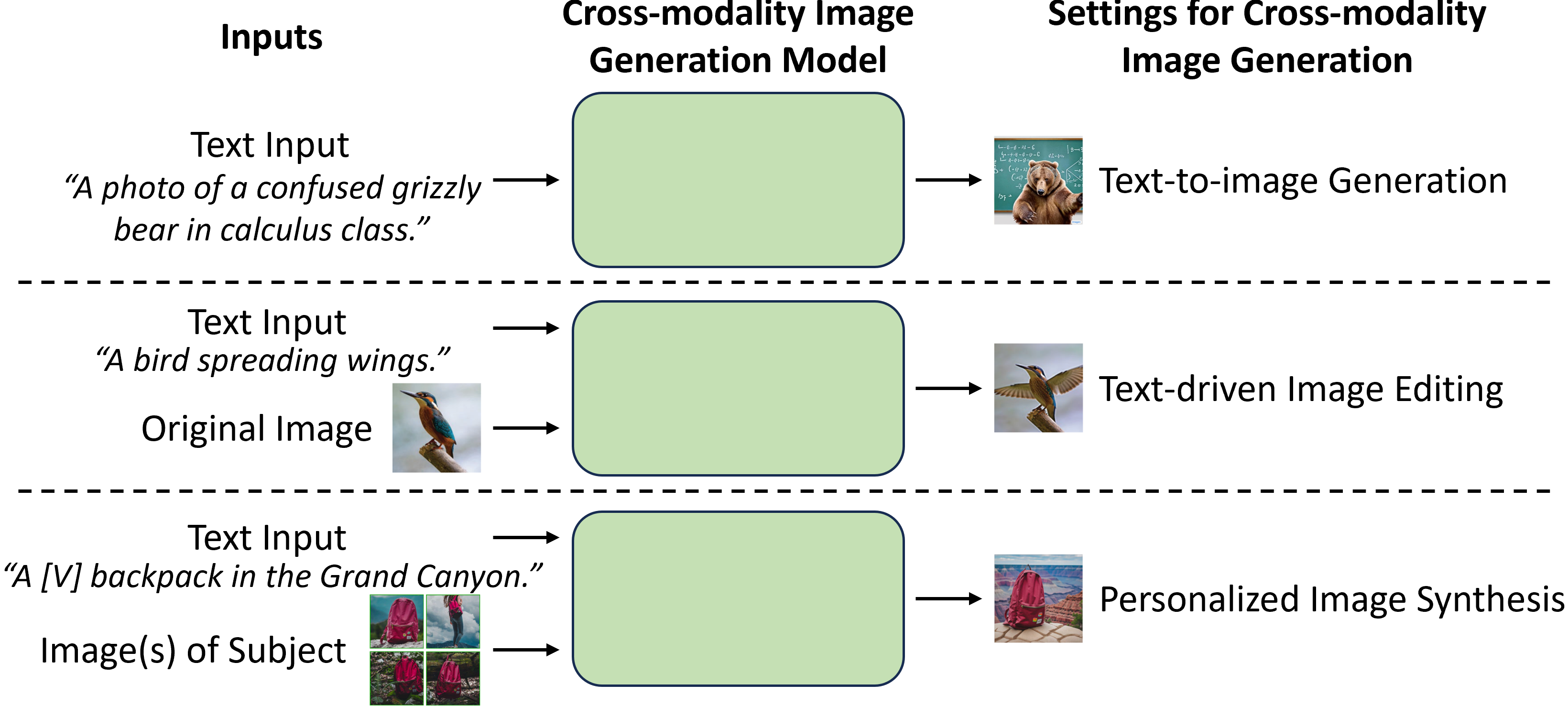}
    \vspace{-4mm}
    \caption{
    Illustration of various cross-modality image generation settings.    
    Note that ``[V]'' refers to the subject's unique identifier.
    Examples obtained from \cite{saharia2022photorealistic,ruiz2023dreambooth,kawar2023imagic}.
    }
    \label{figure:text_to_image}
\vspace{-3.5mm}    
\end{wrapfigure}

Text-to-image has several mainstream settings: generation, editing, and personalized image synthesis, which we discuss below. A visual summary is shown in Figure \ref{figure:text_to_image}.

\noindent
\textbf{Text-to-Image Generation.}
In the text-to-image task, the goal is to generate images corresponding to a given text description. 
Text-to-image is a popular direction that has attracted a lot of attention, and many developments have been made over the years.
In general, these developments can be categorized as: GANs, Autoregressive Transformers, and DMs. 
We discuss these developments, especially in terms of

\noindent
how they bridge the modalities. Note that these techniques are often foundational works in cross-modality generation, and are often the foundation for other non-image-based cross-modality generation as well.
A comparison between representative works can be found in Table \ref{table:text2image_methods}.

\textit{GANs.}
Most GAN-based works for text-to-image generation incorporate a \textit{pre-trained text encoder} (e.g., an RNN) to encode text embeddings which are fed to the GAN.
Notably, by using a pre-trained-text encoder trained only on text inputs, we can get meaningful text embeddings as inputs to the GAN, even while using only a relatively small image-text dataset.
Therefore, through the use of a pre-trained text encoder, early works were able to bridge the cross-modality gap for GANs, and generate images corresponding to the input text embeddings.
As the first work exploring this direction, AlignDRAW \cite{mansimov2015generating} generates images mainly by using RNNs to encode the text inputs, and generates images  which learns an alignment between the input captions and the generating canvas, but employ GANs for post-processing to improve the image quality.
Then, Reed et al. \cite{reed2016generative} develops a fully end-to-end differentiable deep convolutional GAN architecture with a convolutional-RNN, which significantly improves the realism and resolution of the generated images (from $28 \times 28$ of AlignDraw to $64 \times 64$ images).
GAWWN \cite{reed2016learning} builds upon \cite{reed2016generative} and finds that image quality can be improved by providing additional information to the model, such as bounding boxes that instruct the model where to generate the objects.
StackGAN \cite{zhang2017stackgan} and StackGAN++ \cite{zhang2018stackgan++} adopt the text encoder of \cite{reed2016generative}, and present a coarse-to-fine refinement approach and stack multiple GANs to improve the image resolution, yielding images of resolution $256 \times 256$.
AttnGAN \cite{xu2018attngan} proposes an attention-driven architecture that allows subregions of the image to be drawn by focusing on the most relevant words of the caption, providing significant improvements to the image quality.
Other GAN-based text-to-image methods have been proposed for better resolution (e.g., DM-GAN \cite{zhu2019dm}), or better semantic alignment through training losses (e.g., XMC-GAN \cite{zhang2021cross}) and text-image fusion architectures (DF-GAN \cite{tao2022df}).

\begin{wraptable}{r}{0.51\textwidth}
\vspace{-2.5mm}
  \tiny
  \centering
  \caption{\tiny
  Comparison between representative text-to-image generative models.
  Results are reported on MS-COCO \cite{lin2014microsoft} and CUB \cite{wah2011caltech,reed2016learningdeep}.
  For both datasets, we report the Fr\'echet Inception Distance (FID) and Inception Score (IS), where FID measures the similarity between the generated images and reference ground truth images with the same captions, while IS measures the diversity of generated images and whether it can be clearly determined by a classifier.
  On MS-COCO, we also report the zero-shot FID metric, where models are tasked to generate images for the MS-COCO captions without dataset-specific tuning, which is a metric that recent works tend to focus on.
   A dash `-' indicates that the metric was not reported for that work. The best result in each column is in bold.
  }
  \vspace{-0.25cm}
  \label{table:text2image_methods}
  \scalebox{0.95}{
  \hspace{-0.4cm}
  \begin{tabular}{lc|ccc|cccc}
    \toprule
     &  & \multicolumn{3}{c}{MS-COCO} & \multicolumn{2}{|c}{CUB}  \\   
    Type & Method & FID$(\downarrow)$ & Zero-shot FID$(\downarrow)$ & IS$(\uparrow)$ & FID$(\downarrow)$ & IS$(\uparrow)$ \\    
    \midrule
    \multirow{2}{*}{GANs} & GAN-INT-CLS \cite{reed2016generative} & -  &  - & 7.88 & - & 2.88   \\
    &  GAWWN \cite{reed2016learning}  &  -  &  -  & -  &  -  &  3.62  \\
    & StackGAN \cite{zhang2017stackgan} &  74.05  & -   & 8.45  & 51.89 & 3.70  \\
    & StackGAN++ \cite{zhang2018stackgan++} &  81.59  & -   & 8.30  & 15.30 & 4.04  \\
    & PPGN \cite{nguyen2017plug}  & - & - & 9.58 & - & - \\
    & AttnGAN \cite{xu2018attngan} &  35.49  &  - & 25.89   &  23.98  &  4.36  \\
    & DM-GAN \cite{zhu2019dm} &  32.64  & -  &   \textbf{30.49}  &  16.09  &  4.75 \\
    & DF-GAN \cite{tao2022df} & 19.32  &  - &  -  & 14.81  & \textbf{5.10} \\
    & XMC-GAN \cite{zhang2021cross} & 9.33 &  -  & 30.45 & - & -  \\
    \midrule
    \multirow{2}{*}{Autoregressive} & DALL-E \cite{ramesh2021zero} & - &   27.50 & - & - & - \\
    & CogView \cite{ding2021cogview} &  - & 27.10  & -  &  -  & - \\
    & CogView2 \cite{ding2022cogview2} & 17.5  &  24.0 & 25.2 & - & - \\
    & N\"UWA \cite{wu2022nuwa} & 12.9 & -  & 27.2 & - &  -  \\
    & Make-A-Scene \cite{gafni2022make} &  7.55 &  11.84 - & - & - \\
    & Parti \cite{yu2022scaling} &  \textbf{3.22}  &  7.23 &  - &  - &  -  \\
    \midrule
    \multirow{2}{*}{DMs} & Stable Diffusion \cite{rombach2022high} &  -  & 12.61 & -  &  -  &  -  \\     
    & GLIDE \cite{nichol2022glide} &  - &   12.24 &  - &  - &  -   \\
    &  VQ-Diffusion \cite{gu2022vector} & 13.86  &  -  &  -  & \textbf{10.32}  &  -  \\
    & DALL-E 2 \cite{ramesh2022hierarchical} &  - &  10.39 &  - &  - &  -  \\
    & Imagen \cite{saharia2022photorealistic} & -  &  7.27 &  - &  - &  -   \\
    & eDiff-i \cite{balaji2022ediffi} &  -  & \textbf{6.95}  &  - &  - &  -  \\
    & GLIGEN \cite{li2023gligen} & 5.61 & - & - & - & -  \\
    \bottomrule
  \end{tabular}
  }
\vspace{-3mm}
\end{wraptable}

\textit{Autoregressive Transformers.}
With the rise of Autoregressive Transformers, many works explored a straightforward approach to use an \textit{Autoregressive Transformer to autoregressively model text and image tokens} (i.e., inputs) as a single stream of data, allowing them to take text tokens as input while producing image tokens as output. 
As they are able to treat input image and text tokens as a single stream of data, it can be said that these autoregressive methods aim to model the joint distribution of image and text tokens.
Yet, to train these Autoregressive Transformer models, typically without pre-trained text encoders, often requires large amounts of image-text paired data for training.
These models also often require the images to be discretized into discrete tokens, so that they have a similar format to text tokens which are discrete, meaning that they also often require support from other encoder models. 
Therefore, many Autoregressive Transformers employ VAEs to encode the images into image tokens, together with text tokenization methods to get text tokens. 
DALL-E \cite{ramesh2021zero} is the first to leverage autoregressive Transformers \cite{vaswani2017attention} for text-to-image generation, by using them to generate a sequence of image tokens after taking the text tokens as input.
The image tokens are then decoded into a high-quality images via a trained discrete VAE \cite{van2017neural}.
CogView \cite{ding2021cogview} also proposes a similar approach, but with significant improvements in image quality, and also investigates fine-tuning for downstream tasks.
CogView 2 \cite{ding2022cogview2} introduces a cross-modality pre-training method (CogLM) which facilitates the conditioning on both image and text tokens to perform various tasks such as image captioning or image editing, on top of text-to-image generation.
N\"UWA \cite{wu2022nuwa} trains an encoder that takes text or visual inputs, and an autoregressive decoder with an efficient attention mechanism that is shared among 8 visual synthesis tasks, e.g., text-to-image generation and manipulation. 
Make-A-Scene \cite{gafni2022make} improves the controllability of text-to-image generation by allowing an optional segmentation map as input, while also increasing the quality and resolution of the generated images by improving the tokenization process.
Parti \cite{yu2022scaling} further scales up the model size to improve image quality.

\textit{Diffusion Models.}
More recently, DMs have shown to be a very effective approach for text-to-image generation, and have attracted much attention.
Even though diffusion models \cite{saharia2022photorealistic} mostly only require a text encoder that is pre-trained only on text  (much like GANs), many works have found advantages with using text embeddings from a joint text-image embeddings space (e.g,, from CLIP \cite{radford2021learning}). 
For instance, CLIP \cite{radford2021learning} is trained in a contrastive manner on a large image-text dataset, and has a joint image-text embedding space that greatly facilitates the bridging of modalities for the diffusion model.
By leveraging the joint image-text space, one notable advantage is the ability train text-to-image diffusion models while using only an image-only dataset \cite{sheynin2023knn}.
Another notable advantage includes the ability to perform zero-shot language-guided editing \cite{ramesh2022hierarchical}. 
One of the earliest works that explores DMs for text-to-image generation, GLIDE \cite{nichol2022glide} investigates using guidance from CLIP \cite{radford2021learning} or classifier-free guidance \cite{ho2021classifier}, where the latter produces photorealistic images that are consistent with the captions. 
Imagen \cite{saharia2022photorealistic} leverages a generic language model that is pre-trained only on text (T5 \cite{raffel2020exploring}) to encode text and a dynamic thresholding technique for diffusion sampling, which provide improved image quality and image-text alignment.
Different from previous methods that generate in the pixel space, Stable Diffusion \cite{rombach2022high} introduces Latent Diffusion Models that perform diffusion in the latent space.
Specifically, an autoencoder is first trained, and the diffusion model is then trained to generate features in the latent space of the autoencoder, which greatly reduces training costs and inference costs. 
DALL-E 2 \cite{ramesh2022hierarchical} also performs diffusion in the latent space, but instead use the joint text and image latent space of CLIP \cite{radford2021learning}, which has the advantage of being able to semantically modify images by moving in the direction of any encoded text vector.
VQ-Diffusion \cite{gu2022vector} performs diffusion in the latent space of a VQ-VAE \cite{van2017neural} while encoding text inputs using CLIP.
Some other developments include having a database of reference images (e.g., RDM \cite{blattmann2022retrieval}), or training an ensemble of text-to-image diffusion models (e.g., eDiff-I \cite{balaji2022ediffi}).

\noindent
\textbf{Text-driven Image Editing.}
In text-driven image editing, input images are manipulated according to given text descriptions to obtain an edited version of the original image.
In this task, one main challenge lies in how to obtain representations of the input image, such that it can be edited through text prompts.
Another difficulty lies in how to have better control over the edits such that the edits are closer to what the user desires, while maintaining consistency over the other elements in the original image.
Earlier works on text-driven editing are based on GANs, where a text encoder is added to the GAN and the GAN is trained on image-text pairs (e.g., ManiGAN \cite{li2020manigan}. 
For better zero-shot performance with pre-trained image GANs, GAN inversion techniques are applied to invert images into the StyleGAN latent space \cite{tov2021designing}, or are combined with the CLIP's text-image embedding space for image editing (e.g., StyleCLIP \cite{patashnik2021styleclip}).
Such inversion techniques into meaningful latent spaces allow images to be conveniently edited by text prompts.
Recently, many works focus on editing of images with DMs.
DiffusionCLIP \cite{kim2022diffusionclip} explores the text-driven editing of an image by using a CLIP loss on DMs.
Specifically, an input image (to be edited) is first transformed to the latent space via DMs, and the DM is fine-tuned with the directional CLIP loss (which uses the text prompts to provide a gradient), such that it produces updated samples during the reverse diffusion process.
Blended Diffusion \cite{avrahami2022blended} adopts a similar approach, but also allow for region-based image editing where edits are contained in a local region, and enable this through masking the CLIP loss and enforcing a background preservation loss.
Imagic \cite{kawar2023imagic} leverages a pre-trained text-to-image DM and fine-tunes the DM according to the input image, which enables sophisticated manipulations of a real high-resolution image, including editing of multiple objects
To have more control over the edits, another line of  works explore ways to input additional information (e.g., sketches, segmentation maps) to the generation process.
Voynov et al. \cite{voynov2022sketch} take in an additional input sketch to guide a pre-trained text-to-image DM to generate an image that follows the spatial layout of the sketch.
Make-A-Scene \cite{gafni2022make} explores controllable text-to-image generation, allowing users to input additional sketches or edit extracted segmentation maps to the desired classes.
Then, ControlNet \cite{zhang2023adding} aims to allow pre-trained large text-to-image DMs to support many kinds of additional input conditions, e.g., scribbles, poses, segmentation maps, via a fine-tuning approach involving the zero convolution layer.
To perform grounded text-to-image generation and have localized edits, GLIGEN \cite{li2023gligen} takes input prompts and other additional information (e.g., bounding boxes, human poses, depth maps, semantic maps) into a frozen pre-trained text-to-image DM.

\noindent
\textbf{Personalized Image Synthesis.}
Personalized image synthesis is another notable text-to-image task, where the cross-modality generative models aim to generate novel images of a given subject (e.g., a person or bag).
Despite sharing many similarities with text-based image editing, there are some subtle yet important differences between them.
In personalized image synthesis, the aim is to let the model get an understanding of the specific object or thing (through at least 3-5 images), and the model can then be used to generate the object in many novel scenarios.
In contrast, for image editing, only one image is input to the model, and the focus of the model is often to simply change objects or the background; it can be hard to make overly large edits to this image, e.g., it is often difficult to output an image of an object from a novel angle with a novel background.
Hence, since their aims are different, there exists important technical differences between works in personalized image synthesis vs image editing. For instance, methods for personalized image synthesis often represent the personalized concepts as ``words'' that can be flexibly used to generate novel images, or may require fine-tuning on the user-provided images of the object. These approaches are different from the text-based editing works.
Many works leverage GANs to perform this task, such as IC-GAN \cite{casanova2021instance} which additionally requires the discriminator to predict instance information, and Pivotal Tuning \cite{roich2022pivotal} which adopts a GAN inversion approach.
Recently, many works leverage pre-trained text-to-image DMs, which have shown promising improvements.
Textual Inversion \cite{gal2023an} uses only a few images (typically 3-5) of a user-provided concept, and represents the concept through new "words" in the embedding space of a text-to-image DM, enabling personalized creation of concepts guided by natural language sentences.
DreamBooth \cite{ruiz2023dreambooth} aims to generate ``personalized'' images for a given subject (e.g., a specific dog or human) and introduce an approach to fine-tune the pre-trained text-to-image diffusion model which result in better performance.
Custom Diffusion \cite{kumari2023multi} further extends this to multiple concepts and their compositions, through jointly training for multiple concepts or combining multiple fine-tuned models into one.
Recent developments \cite{ruiz2023hyperdreambooth} further improve the efficiency and speed of the personalization process.

\vspace{3mm}

\subsection{Summary of Cross-Modality Image Generation}
\vspace{1mm}
To recap, in this section we discussed the cross-modality image generation methods.
Firstly, we discuss text-to-image generation, where we discuss the progression of methods from GANs \cite{mansimov2015generating}, to autoregressive Transformers \cite{ramesh2021zero}, then to DMs \cite{rombach2022high}.
We highlight how different approaches bridge the gap between the two different modalities. As most foundational cross-modality techniques are developed in the text-to-image setting, these methods often form the foundation for methods in the many other cross-modality settings as well.
Then, we also introduce text-driven image editing \cite{li2020manigan}, and personalized image synthesis \cite{roich2022pivotal}, where we discuss methods that want stronger control over the customization of their images, through techniques such as inversion \cite{tov2021designing}, and optimizing new ``words'' for new objects \cite{gal2023an}.

\section{Cross-Modality Video Generation}

Following the tremendous progress in cross-modality image generation, more works have explored the more challenging task of cross-modality video generation.
Cross-modality video generation is mostly conditioned on text, which we discuss in Section \ref{sec:text_to_video}.

\begin{wraptable}{r}{0.44\textwidth}
\vspace{-5mm}
  \tiny
  \centering
  \caption{\tiny
  Comparison between representative text-to-video generative models.
  Results are reported on MSR-VTT \cite{xu2016msr} and UCF-101 \cite{soomro2012ucf101}, on both the fine-tuned setting (where the dataset is used during training) and the zero-shot setting (where the dataset is not used during training).
  For the MSR-VTT dataset, we report the Fr\'echet Inception Distance (FID) \cite{parmar2022aliased} and CLIP similarity (CLIPSIM) \cite{wu2021godiva}, where CLIPSIM measures the average CLIP similarity between the generated video frames and text which evaluates the semantic match between them (higher is better).
  For the UCF-101 dataset, we report the FVD and IS metrics (refer to Table \ref{table:video_methods} for more details).
  Note that, for the UCF-101 dataset in the text-to-video setting, class names are provided directly as text conditioning.
  A dash `-' indicates that the metric was not reported for that work.
  }
  \vspace{-0.35cm}
  \label{table:text2video_methods}
  \scalebox{0.95}{
  \hspace{-0.4cm}
  \begin{tabular}{lc|cc|cccc}
    \toprule
     &  & \multicolumn{2}{c}{MSR-VTT} & \multicolumn{2}{|c}{UCF-101}  \\   
    Type & Method & FID$(\downarrow)$ & CLIPSIM$(\uparrow)$ & FVD$(\downarrow)$ & IS$(\uparrow)$ \\    
    \midrule
    \multirow{2}{*}{Fine-tuned} &  GODIVA \cite{wu2021godiva} &  -  & 0.2402   & -  &  -  \\
    &  N\"UWA \cite{wu2022nuwa} &  47.68  &  0.2439  & -  &  -  &   \\
    & Make-A-Video \cite{singer2023make} &  - &  -  &  81.25 & 82.55 \\
    \midrule
    \multirow{2}{*}{Zero-shot} &  CogVideo (Chinese) \cite{hong2023cogvideo}  &  24.78  &  0.2614  & 751.34  &  23.55    \\
    & CogVideo (English) \cite{hong2023cogvideo}  &  23.59  &  0.2631  & 701.59  &  25.27 \\ 
    &  Make-A-Video \cite{singer2023make}  &  13.17  &  0.3049  & 367.23  &  33.00 \\ 
    &  VideoLDM \cite{blattmann2023align}  &  -  &  0.2929  & 550.61  &  33.45 \\ 
    \bottomrule
  \end{tabular}
  }
\vspace{-2mm}
\end{wraptable}

\subsection{Conditioned on Text}
\label{sec:text_to_video}

We discuss two popular text-to-video settings below: text-to-video generation, and text-driven video editing.

\noindent
\textbf{Text-to-Video generation.}
The text-to-video task aims to generate videos based on the given text captions, and is significantly more challenging than the text-to-image task, since there are more video frames, which all need to be temporally consistent.
It is also challenging to adhere to specified motions or actions in the text prompt, where the video generation model now needs a good understanding of what each motion or action refers to.
The much higher computational requirements also make this task much more difficult.
Therefore, despite the similarities between these methods with text-to-image settings, such as how the methods (e.g., GANs, Autoregressive Transformers, DMs) utilize the embedding space, there are also technical innovations that aim to overcome the challenges in temporal modeling.
Some common themes involve architectural designs to facilitate long-term video generation (e.g., autoregressive Transformers) and ways to incorporate pre-trained image generation models such that less text-video data is required.
A comparison between recent methods is shown in Table \ref{table:text2video_methods}.

\textit{Non-diffusion-based approaches.}
In the earlier days of text-to-video generation, many works explored GAN-based approaches \cite{balaji2019conditional} and VAE-based approaches \cite{marwah2017attentive} or a combination of both \cite{li2018video} to perform the task. 
These works encoded text descriptions into embeddings with a text encoder (e.g., an RNN) and fed the embeddings to their decoder (which is usually an RNN) to generate videos.
However, these approaches tend to focus on simpler settings, and generally only can produce a short clip with low resolution (e.g., $64 \times 64$). The limited video quality of these early explorations serve to highlight the sheer difficulty of text-to-video generation.
Subsequently, with the rise of autoregressive Transformers, researchers explore them for more powerful autoregressive modeling. GODIVA \cite{wu2021godiva} leverages an image-based VQVAE with an autoregressive model for text-to-video generation, where a Transformer takes in text tokens and produces video frames autoregressively.
This approach is extended in N\"UWA \cite{wu2022nuwa} with multi-task pre-training on both images and videos, enabling better performance through an increased dataset size.
N\"UWA-Infinity \cite{liang2022nuwa} further improves the autoregressive generation process by a patch-by-patch generation approach, which improves the synthesis ability on long-duration videos.
CogVideo \cite{hong2023cogvideo} adds temporal attention modules to a frozen CogView 2 \cite{ding2022cogview2} (which is an autoregressive text-to-image Transformer model) to perform text-to-video generation, which significantly reduces the training cost by inheriting knowledge from the pre-trained text-to-image model.

\textit{Diffusion Models.}
Similar to the success found in text-to-image generation, DMs have also found much success in text-to-video generation.
Video Diffusion \cite{ho2022video} is the first to use DMs to present results on a large text-conditioned video generation task, which proposes an architecture for video DMs and trains on captioned videos to perform text-conditioned video generation.
Imagen Video \cite{ho2022imagen} further improves this by using a cascade of video DMs to improve the quality and fidelity of generated videos.
Make-A-Video \cite{singer2023make} aims to perform text-to-video generation without paired text-video data by leveraging text-to-image models for the correspondence between text and visuals.
Story-LDM \cite{rahman2023make} aims to generate stories, where characters and backgrounds are consistent over time.
Overall, although diffusion models are able to produce videos of high quality, they are not typically autogressive, and still struggle with generating long videos of good quality.

\noindent
\textbf{Text-driven Video Editing.}
One research direction also focuses on the editing of videos with text descriptions.
Text2LIVE \cite{bar2022text2live} introduces a general approach without using a pre-trained generator in the loop, which can perform semantic, localized editing of real world videos.
Subsequently, many approaches leverage DMs in their approach.
Subsequent approaches often depend on pre-trained text-to-image generators.
Tune-A-Video \cite{wu2022tune} uses pre-trained image diffusion models to edit the video of a given text-video pair.
Moreover, by adapting an image-based diffusion model, Video-P2P \cite{liu2023videop2p} performs video editing with cross-attention control based on the text inputs.
Gen-1 \cite{esser2023structure} edits videos with a text-conditioned video diffusion model, which is extended from pre-trained text-to-image DMs with temporal layers.

\section{Cross-Modality 3D Generation}
\label{sec:crossmodal_3D}
\noindent
In addition to cross-modality generation of 2D modalities such as image and video, many AIGC methods have also tackled cross-modality 3D generation, including shapes, scenes, humans, and motions.
In this section, we discuss the cross-modal generation of 3D shapes, scenes, humans and motions separately in Sections \ref{sec:crossmodal_3D_shape} - \ref{sec:crossmodal_3D_motion}, and in the same order as in Section \ref{sec:3D}.

\subsection{Cross-Modality 3D Shape Generation}
\label{sec:crossmodal_3D_shape}

Many works have explored generating 3D shapes with input text or image information, which we comprehensively discuss below.

\subsubsection{Conditioned on Text (\textbf{Text-to-3D Shapes})}
In text-to-3D shape generation, the goal is to generate 3D assets corresponding to text descriptions.
This is very challenging, since most 3D shape generative models are trained on datasets of specific object categories like ShapeNet \cite{chang2015shapenet} and struggle to generalize to the zero-shot setting.
Another challenge is the lack of large-scale captioned 3D shape data for training, which can limit the capabilities of trained text-to-3D shape models. 
Earlier approaches train their generative models with text-3D shape pairs.
Text2Shape \cite{chen2019text2shape} collected a dataset of natural language descriptions on the ShapeNet dataset \cite{chang2015shapenet}, and used it to train a 3D GAN for text-to-3D shape generation.
ShapeCrafter \cite{fu2022shapecrafter} collected a larger dataset (Text2Shape++), and used BERT \cite{devlin2019bert}, a pre-trained text encoder, to encode the text information.
However, such approaches are limited by the relatively small number of text-3D shape pairs, and these approaches also cannot perform zero-shot generation (i.e., cannot generalize well).
Therefore, subsequent works in this field overcome these issues with large-scale pre-trained text-image models (e.g., CLIP, DMs) which help to partially overcome the text-3D modality gap and reduce the text-3D pairs required, while also providing prior knowledge to help with generalization,.
Recent works generally rely on one of two approaches: either a CLIP-based approach, or a diffusion-based approach, which we present below.

\textit{CLIP-based Approaches.}
CLIP provides a joint image-text embedding space, which methods exploit to facilitate text-to-3D shape generation.
CLIP-Forge \cite{sanghi2022clip} uses CLIP to train the generative model using images, but use text embeddings at inference time.
Text2Mesh \cite{michel2022text2mesh} leverages CLIP to manipulate the style of 3D meshes, by using CLIP to enforce semantic similarity between the rendered images (from the mesh) and the text prompt.
Dream Fields \cite{jain2022zero} generates 3D models from natural language prompts, while avoiding the use of any 3D training data.
Specifically, Dream Fields optimizes a NeRF (which represents a 3D object) from many camera views such that rendered images score highly with a target caption according to a pre-trained CLIP model.
Dream3D \cite{xu2023dream3d} also performs optimization via the CLIP-based loss (as in Dream Fields), but further initializes a 3D shape prior using Stable Diffusion \cite{rombach2022high}.

\textit{Diffusion-based Approaches.}
DreamFusion \cite{poole2023dreamfusion} adopts a similar approach to Dream Fields, but instead replaces the CLIP-based loss with a loss based on sampling a pre-trained image diffusion model through a proposed Score Distillation Sampling (SDS) approach.
Most subsequent works further develop SDS-based approaches to use text-to-image DMs.
Magic3D \cite{lin2023magic3d} proposes a coarse-to-fine optimization approach with multiple diffusion priors optimized via a SDS-based approach, which eventually produces a textured mesh.
Magic3D significantly improves the speed of DreamFusion (by $2\times$) while achieving higher resolution (by $8\times$).
Latent-NeRF \cite{metzer2023latent} proposes to operate a NeRF in the latent space of a Latent Diffusion Model to improve efficiency, and also introduce shape-guidance to the generation process, which allows users to guide the 3D shape towards a desired shape.
Some recent works also explore SDF-based representations with generative DMs \cite{cheng2023sdfusion}, or focus on generating textures from text input \cite{richardson2023texture}.

\subsubsection{Conditioned on Image (\textbf{Image-to-3D Shapes})}
Besides generating 3D shapes from text input, many works also focus on synthesizing 3D shapes from input images. 
This is very challenging as there can be depth ambiguity in images, and some parts of the 3D shape might also be occluded and are not visible. 
There is also often a lack of 3D data for training.
Thus, researchers often add an image encoder to capture the image information while improving the correspondence between the image and renders of the output 3D shape.
Some works also reduce the reliance on 3D data, by learning 3D shapes completely from images only.
Below, we categorize the methods based on their 3D representation.

\textit{Image-to-3D Voxel.} 
Some earlier works focus on generating 3D shapes as voxels.
3D-R2N2 \cite{choy20163d} is a RNN architecture which takes in one or more images of an object instance and outputs the corresponding 3D voxel occupancy grid, where the output 3D object can be sequentially refined by feeding more observations to the RNN.
TL-embedding network \cite{girdhar2016learning} introduces an autoencoding-based approach for mapping images to a 3D voxel map, which learns a meaningful representation where it can perform both prediction of 3D voxels from images and also conditional generation by combining feature vectors.
To reduce the reliance on 3D data, Rezende et al. \cite{jimenez2016unsupervised} and Yan et al. \cite{yan2016perspective} learn to recover 3D volumetric structures from pixels, using only 2D image data as supervision.

\textit{Image-to-3D Point Cloud.} 
A few works also explore generation of 3D point clouds conditioned on images, which is more scalable than then voxel representation.
Fan et al. \cite{fan2017point} aim to reconstruct a 3D point cloud from a single input image, by designing a point set generation network architecture and exploring loss functions based on the Chamfer distance and Earth Mover's distance.
Recently, PC\textsuperscript{2} \cite{melas2023pc2} introduces a conditional diffusion-based approach to reconstruct a point cloud from an input image.

\textit{Image-to-3D Mesh.}
There has also been much interest in generating 3D meshes from image inputs.
Kato et al. \cite{kato2018neural} propose a neural renderer that can render the mesh output (i.e., project the mesh vertices onto the screen coordinate system and generating the image) as a differentiable operation, enabling generation of 3D meshes from images.
Kanazawa et al. \cite{kanazawa2018learning} presents an approach to generate meshes with only a collection of RGB images, without requiring ground truth 3D data or multi-view images of the object, which enables image to mesh reconstruction.
Pixel2Mesh \cite{wang2018pixel2mesh} also generates 3D meshes from single RGB images with their GCN-based architecture in a coarse-to-fine fashion, which learns to deform ellipsoid meshes into the target shape.
Liao et al. \cite{liao2018deep} design a differentiable Marching Cubes layer which allows learning of an explicit surface mesh representation given raw observations (e.g., images or point clouds) as input.

\textit{Image-to-neural fields.}
To further improve scalability and resolution, some works explore generating neural fields from images.
DISN \cite{xu2019disn} introduces an approach to take a single image as input and predict the SDF which is a continuous field  that represents the 3D shape with arbitrary resolution.
Michalkiewicz et al. \cite{michalkiewicz2019implicit} incorporate level set methods into the architecture and introduces an implicit representation of 3D surfaces as a distinct layer in the architecture of a CNN to improve accuracy on 3D reconstruction from images.
SDFDiff \cite{jiang2020sdfdiff} is a differentiable render based on ray-casting SDFs, which allows 3D reconstruction with a high level of detail and complex topology.

\subsection{Cross-Modality 3D Scene Synthesis}
\label{sec:crossmodal_3D_scene}

With the increasing progress in AIGC methods, more research has focused on generation of 3D scenes instead of 3D shapes only. 
The generation of such scenes can be controlled by other input modalities, such as text, image, and video, which we discuss below.

\subsubsection{Conditioned on Text (\textbf{Text-to-3D Scene})}
Many recent works have explored modelling 3D scenes with text input.
Due to the lack of paired text-3D scene data, most works leverage on CLIP or pre-trained text-to-image models to handle the text input.
Text2Scene \cite{hwang2023text2scene} leverages CLIP to model and stylize 3D scenes with text (or image) inputs, by decomposing the scene into sub-parts for handling.
Text2NeRF \cite{zhang2023text2nerf} generates NeRFs from text with the aid of text-to-image DMs to represent 3D scenes.
Text2Room \cite{hollein2023text2room} further leverages a pre-trained monocular depth prediction model for more geometric consistency, and directly generates the 3D textured mesh representation of the scene.
Po and Wetzstein \cite{po2023compositional} propose a locally conditioned diffusion technique for compositional scene generation with various text captions.
MAV3D \cite{singer2023text} aims to generate 4D dynamic scenes by using a text-to-video diffusion model.

\subsubsection{Conditioned on Image (\textbf{Image-to-3D Scene})}
Unlike using text captions that can be rather vague, some works also synthesize 3D scene representations from image input, where the 3D scene corresponds closely to what is depicted in the image.
PixelNeRF \cite{yu2021pixelnerf} predicts a NeRF based on one or a few posed images via a re-projection approach.
Pix2NeRF \cite{cai2022pix2nerf} learn to output a neural radiance field from a single image by building upon $\pi$-GAN \cite{chan2021pi}.
On the other hand, NeRF-VAE \cite{kosiorek2021nerf} explores a VAE-based approach to learn to generate a NeRF with very few views of an unseen scene.
Recently, Chan et al. \cite{chan2023generative} introduce a diffusion-based approach for generating novel views from a single image, where the latent 3D feature field captures a distribution of scene representations which is sampled from during inference.

\subsubsection{Conditioned on Video (\textbf{Video-to-Dynamic Scene})}
Besides encoding static 3D scenes (with image or text input), another line of works research aim to encode dynamic scenes, where the scene can be changing with respect to time.
This is sometimes also called 4D view synthesis, where a video is taken as input, and a dynamic 3D scene representation is obtained, where users can render images of the scene at any given time instance.
Xian et al. \cite{xian2021space} learn neural irradiance fields (with no view dependency) while estimating the depth to constrain the scene's geometry at any moment.
D-NeRF \cite{pumarola2021d} and Nerfies \cite{park2021nerfies} aim to synthesize novel views of dynamic scenes with complex and non-rigid geometries from a monocular video, through a dynamic NeRF representation that additionally takes the temporal aspect into account to deform the scene.
NR-NeRF \cite{tretschk2021non} implements scene deformation as ray bending which deforms straight rays non-rigidly.
D\textsuperscript{2}NeRF \cite{wu2022dnerf} learns separate radiance fields for the the dynamic and static portions of the scene in a fully self-supervised manner, effectively decoupling the dynamic and static components.

\subsection{Cross-Modality 3D Human Generation}
\label{sec:crossmodal_3D_human}

\begin{wrapfigure}{r}{0.45\textwidth}
\vspace{-7mm}
    \center
    \includegraphics[width=0.44\textwidth]{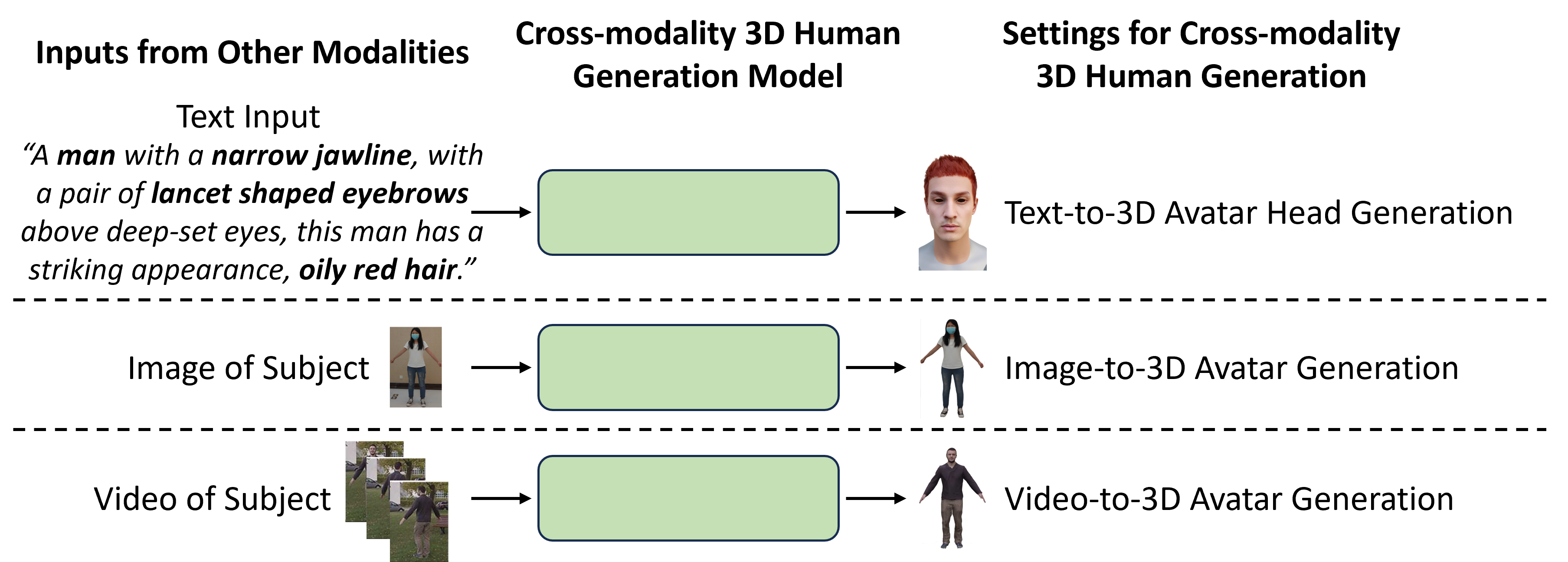}      
    \vspace{-3mm}
    \caption{
    Illustration of various cross-modality 3D human generation settings, where inputs can be text, images or videos. The outputs can often either be avatar heads or full avatars. 
    Pictures obtained from \cite{alldieck2018video,jiang2022selfrecon,zhang2023dreamface}.    
    }
    \label{figure:crossmodal_3dhuman}
\vspace{-6mm}
\end{wrapfigure}

Besides cross-modality generation of 3D shapes and 3D scenes, cross-modality generation of 3D humans has also attracted much attention, where various other modalities are used to condition the generation process to produce the desired 3D human avatar.
Cross-modality generation of 3D humans can also be categorized according to the type of conditioning information: text, image or video, and the various settings are illustrated in Figure \ref{figure:crossmodal_3dhuman}.

\subsubsection{Conditioned on Text (\textbf{Text-to-3D Human})}
Many text-to-3D Human methods often aim to generate avatar heads from text.
Wu et al. \cite{wu2023high} explore a text-to-3D face generation approach, which adopts CLIP to encode text information and generates a 3D face with 3DMM representation.
DreamFace \cite{zhang2023dreamface} also leverages CLIP and DMs in the latent and image space to generate animatable 3D faces via SDS optimization.
ClipFace \cite{aneja2022clipface} uses CLIP to perform text-guided editing of 3DMMs.
Some recent works also generate full 3D human avatars conditioned on text \cite{liu2024humangaussian} via SDS.
Other works aim to edit the clothes of 3D human avatars \cite{gong2024laga}, or control the editing process more precisely \cite{he2024avatar}.

\subsubsection{Conditioned on Image (\textbf{Image-to-3D Human})}
Another line of works generate 3D humans from RGB images. Since the works that generate full avatars and those that generate avatar heads are often different lines of work, we discuss them separately below.

\textit{Image-to-3D Avatar.}
Earlier works in this field aim to recover the human mesh from RGB images, including model-based methods \cite{kanazawa2018end} that build upon human parametric models, as well as model-free methods that directly predict the human mesh via Transformers \cite{lin2021end} or a diffusion process \cite{foo2023distribution}. 
However, this line of works tends to overlook the textures of the human and also clothing aspect, and thus require additional modifications such as external garment layers \cite{bhatnagar2019multi}, displacements \cite{alldieck2019tex2shape}, volumetric methods \cite{varol2018bodynet}.
More recently, PIFu \cite{saito2019pifu} proposed to digitize clothed humans from RGB images via implicit functions (an occupancy field). By aligning the 3D human with pixels of 2D images, PIFu is able to infer 3D surfaces and textures and model intricate shapes such as clothing and hairstyles.
PIFuHD \cite{saito2020pifuhd} further improves the fidelity of PIFu.
The fidelity and quality of reconstructed clothed humans are further improved in more recent works, e.g., by using an SDF to model the human body surface \cite{liao2023high}, or by taking into account albedo and shading information \cite{alldieck2022photorealistic}.
Yet, these above methods produce meshes or models corresponding to an RGB image, but they are not animation ready.
To overcome this, ARCH \cite{huang2020arch} learns to generate detailed 3D rigged human avatars from a single RGB image.
Specifically, ARCH can perform animation of the generated 3D human body by controlling its 3D skeleton pose.
Arch++ \cite{he2021arch++} further improves the reconstruction quality with an end-to-end point-based geometry encoder and a mesh refinement strategy.

\textit{Image-to-3D Avatar Head.}
On the other hand, earlier approaches in image-to-avatar head generation tend to be optimization-based methods, which aim to fit a parametric face model to align with a given image \cite{romdhani2005estimating} or collection of images \cite{roth2016adaptive}. 
However, these methods tend to require iterative optimization, which can be costly.
Recently, many learning-based approaches have also been introduced, where a model (often a deep neural network) directly regresses the 3D face model (usually based on 3DMM \cite{blanz1999morphable} or FLAME \cite{li2017learning}) from input images.
Some earlier works introduce methods that require full supervision \cite{richardson2017learning}, while some methods explore weakly-supervised \cite{deng2019accurate} or unsupervised settings \cite{tewari2017mofa} that do not require ground truth 3D data, increasing the convenience of training these models.
Another line of works produces personalized rigs for each generated face, which facilitates animation.
Some earlier works facilitate the facial animation by estimating a personalized set of blendshapes (e.g., a set containing different expressions) via deformation transfer \cite{sumner2004deformation,cao2016real}, which copies the deformations on a source mesh to a specified target mesh.
More recently, some works \cite{yang2020facescape} estimate the blendshapes via a deep-learning approach.
Notably, DECA \cite{feng2021learning} and ROME \cite{khakhulin2022realistic} propose approaches that produce a face with person-specific details from a single image, and can be trained with only in-the-wild 2D images.
On the other hand, some works \cite{hong2022headnerf} leverage an implicit NeRF representation to further increase the fidelity of generated images from novel views, and the NeRF representations can also be animatable and manipulatable
\cite{zhuang2022mofanerf}.
Note that such implicit approaches are often also able to photorealistically model hair and other complex textures through their implicit representation.

\subsubsection{Conditioned on Video (\textbf{Video-to-3D Human})}
Some other works take in an RGB video as input to generate a 3D avatar. This removes the need for posed 3D scans or posed 2D images (for multi-view settings) while providing better performance than using single images only.
We discuss the works that generate full avatars and those that generate avatar heads separately below.

\textit{Video-to-3D Avatar.} 
Some earlier works for video-to-3D avatar are based on human parametric models, and use video information to model the clothing and hair by learning the offsets to the mesh template (e.g., Alldieck et al. \cite{alldieck2018video}, Octopus \cite{alldieck2019learning}).
Subsequently, ICON \cite{xiu2022icon} produces an animatable avatar from an RGB video via an implicit SDF-based approach.
Specifically, ICON first performs 3D mesh recovery in a frame-wise manner to obtain the surface normals in each frame, and them combine them together to form a clothed and animatable avatar with realistic clothing deformations.
SelfRecon \cite{jiang2022selfrecon} adopts a hybrid approach which optimizes both an explicit mesh and SDF.
SCARF \cite{feng2022capturing} represents the body with a mesh and the clothing with a NeRF, which offers animator control over the avatar shape, pose hand articulations and body expressions while allowing extraction and transference of clothing.
Recently, Vid2Avatar \cite{guo2023vid2avatar} introduces a self-supervised way to reconstruct detailed 3D avatars from monocular videos in the wild, which does not require any ground truth supervision or other priors extracted from large clothed human datasets.

\textit{Video-to-3D Avatar Head.}
Some works aim to generate the 3D avatar head from RGB video, by leveraging multiple unposed and uncurated images to learn the 3D face structure.
Earlier works tend to be optimization-based \cite{garrido2013reconstructing,thies2016face2face}.
Subsequently, some works explore learning-based approaches \cite{tewari2019fml,tewari2018self}, which generalize better and can attain high speeds for avatar head reconstruction.
Some works also explore implicit SDF representation \cite{ramon2021h3d} for producing detailed geometry for the avatar heads.
In order to facilitate head animations, some works \cite{ichim2015dynamic} also generate morphable head avatars from input videos.
To further improve the quality (such as modeling of hair), many works explore implicit representations such as SDFs \cite{yenamandra2021i3dmm}, NeRFs \cite{gafni2021dynamic}, or deformable point-based representations \cite{zheng2023pointavatar}.

\subsection{Cross-Modality 3D Motion Generation}
\label{sec:crossmodal_3D_motion}
Beyond generating 3D human avatars, many AIGC methods also generate the desired 3D motions of the human avatar.
The motions are often using text or audio as input, which we discuss below.

\subsubsection{Conditioned on Text (\textbf{Text-to-3D Motion})}
The generated motions from text-to-3D motion models are represented using either 3D skeleton poses or 3D avatars, where the motion in skeleton poses often capture the movement of the human's major joints, while the movement of 3D avatars represent more fine-grained movements as well.
We discuss them separately below.

\textit{Text-to-3D Skeleton Motion.}
To generate 3D skeleton motions that correspond to an input text description,
JL2P \cite{ahuja2019language2pose} learns a joint text-pose embedding space via an autoencoder to map text to pose motions.
Text2Action \cite{ahn2018text2action} introduces a GAN-based model to synthesize pose sequences from language descriptions.
To generate multiple sequential and compositional actions, Ghosh et al. \cite{ghosh2021synthesis} adopt a GAN-based hierarchical two-stream approach.
Some works \cite{guo2022generating} also explore employing a VAE to synthesize diverse motions.
Following the development of CLIP, MotionCLIP \cite{tevet2022motionclip} leverages the rich semantic knowledge of CLIP \cite{radford2021learning} to align text with the synthesized motion.
T2M-GPT \cite{zhang2023generating} autoregressively generates motion tokens via GPT, and a VQ-VAE \cite{van2017neural} to decode the tokens to motions, which can handle challenging text descriptions and generate high-quality motion sequences.
Recently, DMs have also been explored for text-to-motion synthesis as well (e.g., MDM \cite{tevet2023human}, MoFusion \cite{dabral2023mofusion}) which have also shown to be capable of generating varied motions with many vivid and fine details.

\textit{Text-to-3D Avatar Motion.}
Several works explore motion generation for 3D avatars conditioned on a text description.
Some works (e.g., AvatarCLIP \cite{hong2022avatarclip}, MotionCLIP \cite{tevet2022motionclip}) rely on the rich text and semantic knowledge of CLIP to generate animations of avatars based on text input.
TBGAN \cite{canfes2023text} also leverages CLIP with a GAN-based generator to generate facial animations.

\subsubsection{Conditioned on Audio (\textbf{Audio-to-3D Motion})}
Many works also explore generating motion from audio. Here, we identify three main lines of work: music-to-3D motion, speech-to-3D gestures, and speech-to-head motions.

\textit{Music-to-3D Skeleton Motion.}
Some works aim to generate motion (e.g., dance) from music inputs.
Traditional methods tend to learn the statistical relationships between the music's loudness, rhythm, beat and style and the dance motions \cite{kim2003rhythmic}.
Early deep learning methods leverage the power of RNNs \cite{tang2018dance} to learn to generate dance motions with music input.
To generate diverse plausible motions, some works \cite{lee2019dancing} propose a GAN-based generator.
However, GANs can be unstable to train, thus some methods \cite{li2021ai} generate motions from music inputs via autoregressive Transformers, which allows for autoregressive generation of long sequences of realistic 3D dance motion.
Aristidou et al. \cite{aristidou2022rhythm} introduce a motion generation framework that generates long-term dance animations that are well aligned to the musical style and rhythm, which also enables high-level choreography control.
Recently, MoFusion \cite{dabral2023mofusion} adopts DMs for motion generation, which are effective at long-term and semantically accurate motions.
Some recent works \cite{tseng2023edge} also aim to edit the generated dance motions.

\textit{Speech-to-3D Body Gesture.}
Another line of works condition the generation process on an input speech to generate gestures.
This is also known as the co-speech gesture synthesis task, and is useful to help a talking avatar act more vividly.
Before the proliferation of deep learning, rule-based methods \cite{cassell1994animated} were proposed, where speech-gesture pairs were manually defined.
Subsequently, RNN-based approaches were proposed to automatically learn the speech-to-gesture synthesis \cite{hasegawa2018evaluation,kucherenko2019analyzing} from collected datasets.
In order to produce diverse motions, approaches based on VAEs \cite{li2021audio2gestures} and NFs \cite{henter2020moglow} have also been proposed.
Moreover, to further improve the realism of generated motions, GAN-based approaches \cite{ginosar2019learning,liu2022learning} were proposed.
Recently, DMs have been explored for co-speech gesture synthesis \cite{zhu2023taming}, which further improve the generation quality and avoid the training instability of GANs.

\textit{Speech-to-3D Head Motion.}
Instead of generating full body motions, some works also focus on generating head and lip movements based on the given speech. This is also known as ``talking head'' generation.
Some earlier works explore RNNs \cite{son2017lip} and CNNs \cite{wiles2018x2face} to directly synthesize videos with appropriate lip movements.
Some works \cite{chen2019hierarchical,prajwal2020lip} also employ GANs to generate the lip movements in videos from audio more realistically.
Some works directly perform animation on the 3D avatar heads.
Karras et al. \cite{karras2017audio} propose a CNN architecture to learn to map audio inputs and an emotional state into a animations of a 3D head mesh. 
Several works \cite{suwajanakorn2017synthesizing,thies2020neural} instead represent head movements with a head parametric model (i.e., a 3DMM), and drive the parametric model with speech input.
Recently, talking heads have also been synthesized via implicit representations.
AD-NeRF \cite{guo2021ad} presents an approach to map the audio features to dynamic NeRFs, which is a more powerful and flexible representation that can freely adjust the deformations of the head and model the fine details such as teeth and hair.
Subsequent works further improve the fidelity \cite{ye2023geneface} of the head motions, or reduce the human data required as input for each identity \cite{li2023one}.

\subsection{Summary of Cross-Modality 3D Generation}
To recap, in this section we reviewed the cross-modality AIGC methods to generate 3D outputs, covering 3D shapes, 3D scenes, 3D humans and 3D motions.
For cross-modality 3D shape generation (Section \ref{sec:crossmodal_3D_shape}), we first present a review of text-driven shape generation, where recent methods are often based on CLIP \cite{sanghi2022clip} and DMs \cite{poole2023dreamfusion}.
For image-driven shape generation, we present the developments across the various 3D representations, including voxels \cite{choy20163d}, point clouds \cite{fan2017point}, meshes \cite{kanazawa2018learning}, and neural fields \cite{xu2019disn}.
Then, for cross-modality 3D scene synthesis (Section \ref{sec:crossmodal_3D_scene}), we present the methods conditioned on text \cite{hwang2023text2scene}, image \cite{cai2022pix2nerf}, and video \cite{pumarola2021d}.
Next, in Section \ref{sec:crossmodal_3D_human} we review the methods for cross-modality 3D human generation, and cover the AIGC methods to generate 3D humans from text \cite{zhang2023dreamface}, image \cite{huang2020arch} and video \cite{alldieck2018video}. Note that the methods for human bodies and heads are often separate lines of work.
We survey the papers for 3D motion generation in Section \ref{sec:crossmodal_3D_motion}, where the methods are often conditioned on text \cite{ahn2018text2action} or audio in the form of music \cite{aristidou2022rhythm} or speech \cite{suwajanakorn2017synthesizing}.

\section{Cross-Modality Audio Generation}
\label{sec:crossmodal_audio}

While audio (e.g., speech and music) can be used to generate 3D motion in a cross-modality manner, the audio itself can also be generated from other modalities (e.g., text), which we discuss below.

\subsection{Conditioned on Text}
Two main research directions exist to generate audio from text: text-to-speech and text-to-music. We present them below.

\noindent
\textbf{Text-to-Speech.}
A popular research direction explores the generation of speech conditioned on text. 
Tacotron \cite{wang2017tacotron} introduces an RNN-based end-to-end trainable generative model that generates speech from characters, and is trained on audio-text pairs without phoneme-level alignment, while using the Griffin-Lim reconstruction algorithm \cite{griffin1984signal} to synthesize the final waveform to reduce artifacts.
DeepVoice 3 \cite{ping2018deep} is a fully convolutional architecture that is compatible for use with various vocoders (which convert the compact audio representations into audio waveforms) such as the Griffin-Lim \cite{griffin1984signal} and WaveNet \cite{oord2016wavenet} vocoders.
Tacotron 2 \cite{shen2018natural} extends Tacotron and uses a WaveNet-based vocoder to synthesize high-quality speech from a sequence of characters.
Transformer-TTS \cite{li2019neural} leverage a Transformer-based architecture which effectively models long-term dependencies and improves training efficiency.
FastSpeech \cite{ren2019fastspeech} greatly speeds up the audio synthesis through parallel generation of mel-spectrograms.
GAN-TTS \cite{binkowski2020high} introduces a GAN-based approach to produce speech, which is more efficient than autoregressive approaches.
Glow-TTS \cite{kim2020glow} combines the properties of a flow-based generative method \cite{kingma2018glow} and dynamic programming to learn its own alignment between speech and text via a technique called monotonic alignment search.
Grad-TTS \cite{popov2021grad} adopts a DM for text-to-speech synthesis, which generates high-quality mel-spectrograms and also aligns text and speech through monotonic alignment search, and provides control to trade-off the quality of the mel-spectrogram with inference speed.
As DMs are more stable and simpler to train than GANs while providing high-quality generated audio, many recent works adopt them for text-to-speech synthesis, e.g., Diffsound \cite{yang2023diffsound} and DiffSinger \cite{liu2022diffsinger}.

\noindent
\textbf{Text-to-Music}
Another direction also aims to generate music from text inputs.
Jukebox \cite{dhariwal2020jukebox} is an earlier work that allows generation of a singing voice conditioned on lyrics.
Recent works focus more on music generation based on provided text descriptions.
MusicLM \cite{agostinelli2023musiclm} relies on the joint text-audio embedding space of MuLan \cite{huang2022mulan} to produce audio representations from text input.
More recently, Noise2music \cite{huang2023noise2music} and Mo\^ usai \cite{schneider2023mo} adopt DMs with Transformer-based text encoders, which further improves performance.

\section{Summary of Cross-Modality AIGC Methods}
In summary, in Sections \ref{sec:crossmodal_image}-\ref{sec:crossmodal_audio}, we have covered the cross-modality AIGC methods that generate image, video, 3D, and audio modalities using information from other modalities. 
For each output modality, we further categorize the works according to the modality of the (input) conditioning information, further followed by the various settings that can result from that input-output modality combination.
Specifically, at the lowest levels, we focus on how the different cross-modality methods bridge the different modalities, or how they share a common latent representation.
A common theme is the usage of pre-trained text encoders (e.g., BERT \cite{devlin2019bert}), or operating upon a shared cross-modality embedding space (e.g., CLIP \cite{radford2021learning}).
See Figure \ref{figure:taxonomy_combined2} for the taxonomy of these cross-modality methods.

\section{Datasets}

\begin{wraptable}{r}{0.54\textwidth}
\vspace{-5mm}
  \tiny
  \centering
  \caption{\tiny
  Some representative benchmark datasets for AIGC throughout various data modalities. 
  We report the total number of samples in the full dataset, as well as the number of classes (if available).
  Note that, for the large web-crawled text datasets, we follow \cite{touvron2023llama,raffel2020exploring} to report the dataset size on disk, since it can be hard to compare between them (as each sample can be a sentence, a paragraph, or a document).  
  }
  \vspace{-0.25cm}
  \label{table:datasets}
  \scalebox{0.95}{
  \hspace{-0.4cm}
  \begin{tabular}{lcccccccc}
    \toprule
    Dataset & Year  & Modality  & \#Sample & \#Class     \\
    \midrule
    Cats \cite{zhang2008cat} &  2008  & image  &  10K  &  -   \\
    CIFAR-10 \cite{Krizhevsky09cifar10} &  2009  &  image  &   60K  &   10    \\
    ImageNet \cite{russakovsky2015imagenet}  &  2009  &  image  &  1.4M  &  1000    \\
    LSUN Cat \cite{yu2015lsun} &  2015  &  image  &  1.6M  &   -  \\
    LSUN Bedroom \cite{yu2015lsun} & 2015   &  image  &  3.0M  &  -   \\
    LSUN Horse \cite{yu2015lsun} & 2015   &  image  &  2.0M  &  -   \\        
    LSUN Churches \cite{yu2015lsun} & 2015   &  image  &  126K   &  -   \\
    Summer $\leftrightarrow$ Winter \cite{zhu2017unpaired} & 2017   &  image  &  2K  &   2  \\
    Photo $\leftrightarrow$ Art \cite{zhu2017unpaired} & 2017   &  image  &  10K  &   5  \\
    JFT-300M \cite{sun2017revisiting} & 2017  & image  & 300M & 18291  \\
    CelebA-HQ  \cite{karras2018progressive} &  2018  &  image  & 30K   &   -   \\
    FFHQ \cite{karras2019style}  &  2019  &  image   & 70K   &  -   \\
    AFHQ \cite{choi2020starganv2} & 2020 & image  & 15K & -  \\ 
    UCF-101 \cite{soomro2012ucf101}   &  2012  &  video  & 13K   &   101  \\
    Moving MNIST \cite{srivastava2015unsupervised} &  2015  & video  & 10K  &  -  \\
    Cityscapes  \cite{cordts2016cityscapes}  &  2016   &  video  &  3K  &  -   \\
    Youtube-8M \cite{abu2016youtube} & 2016 & video & 8M & 4800    \\
    Robotic pushing prediction \cite{finn2016unsupervised}  &  2016  & video    &  57K  &  -   \\
    BAIR robot pushing \cite{ebert2017self}  &  2017  & video    &  45K  &  -   \\
    Kinetics-600 \cite{carreira2018short} & 2017 & video &  490K  & 600  \\
    Sky Timelapse \cite{xiong2018learning}    &  2018  & video   &  38K  &  -   \\
    FaceForensics \cite{rossler2018faceforensics}  &   2018 & video   &  1K  &   -  \\
    Taichi-HD \cite{siarohin2019first}   &  2019  & video   &  3K  &  -   \\
    SNLI \cite{bowman2015large}  & 2015   &  text   &  570K  &   3  \\
    LAMBADA \cite{paperno2016lambada}   &  2016  &  text  &  10K  &  -   \\
    SQuAD \cite{rajpurkar2016squad} &  2016  & text   &  107K  & -    \\
    TriviaQA  \cite{joshi2017triviaqa}  &  2017  &  text  &  174K  &  -   \\
    OpenBookQA \cite{mihaylov2018can}     &  2018  &  text  &  5K   &  -  \\
    ARC-e \cite{clark2018think}   &  2018  &  text  &  5K  & -     \\
    ARC-c \cite{clark2018think}    &  2018  &  text  &  2K  & -    \\
    NaturalQuestions \cite{kwiatkowski2019natural}     &  2019  &  text  &  315K    &  -  \\
    CoQA  \cite{reddy2019coqa}  &  2019  &  text  &  127K   & -    \\
    BoolQ \cite{clark2019boolq}  &  2019  & text   &  12K  & -     \\
    GLUE \cite{wang2019glue} &  2019  &  text  & 1.4M   &   -  \\
    SuperGLUE \cite{wang2019superglue} & 2019   &  text  &  195K  &  -   \\
    HumanEval \cite{chen2021evaluating} &  2021  & text  & 164  &  -  \\  
    WinoGrande \cite{sakaguchi2021winogrande}   &  2021   & text    &  44K   &   -  \\
    Common Crawl\textsuperscript{1} \cite{touvron2023llama}     &  2017-2020  &  text  &  3.3TB  &   -  \\    
    C4  \cite{raffel2020exploring}   & 2020   &  text  &  783GB  &   -  \\
    English Wikipedia\textsuperscript{2}  &  2002-now  & text   &  20GB  &  -    \\    
    The Pile \cite{gao2020pile}  & 2020   &  text  &  825GB  &  -   \\
    ShapeNet \cite{chang2015shapenet} &  2015  & 3D shape   &  3M  & 3135     \\
    ShapeNet Core \cite{chang2015shapenet} &  2015  & 3D shape   &  51K  & 13    \\
    ShapeNet Chair \cite{chang2015shapenet} &  2015  & 3D shape   &  6778  & -   \\  
    ShapeNet Airplane \cite{chang2015shapenet} &  2015  & 3D shape   &  4045  & -   \\ 
    ShapeNet Car \cite{chang2015shapenet} &  2015  & 3D shape   &  7497  & -   \\
    ModelNet \cite{wu20153d} &  2015  & 3D shape   & 48K   &  660   \\
    CelebA \cite{liu2015deep}  &  2015  &  scene views &  200K &   10K identities \\
    CompCars \cite{yang2015large} &  2015  &   scene views  &  136K   &   1716 car models \\
    CARLA \cite{dosovitskiy2017carla,schwarz2020graf}  &  2020  &  scene views   &   10K &  16 car models  \\
    D-Faust \cite{bogo2017dynamic} &  2017  &  3D human sequences  &  129 sequences  &   10 subjects  \\
    CAPE \cite{ma2020learning} & 2020   &  3D clothed human sequences  &  $>$600 sequences  &  15 subjects   \\
    FaceScape \cite{yang2020facescape} & 2020   &  3D face  & 18K   &  938 subjects   \\
    DeepFashion \cite{liu2016deepfashion} &  2016  &  clothes image  &  800K  &  50   \\
    CelebAMask-HQ \cite{lee2020maskgan} & 2020   &  face image  &  30K  &  19   \\
    SHHQ \cite{fu2022stylegan} & 2022   &  full body human image  & 230K   &  -    \\
    CMU-Mocap\textsuperscript{3} \cite{li2018convolutional} &  2018  &  3D motion  &  249  &   8  \\
    AMASS \cite{mahmood2019amass}  &  2019  &  3D motion  &  11K  &   344 subjects  \\
    HumanAct12 \cite{guo2020action2motion} &  2020  &  3D motion  &  1K  &  12   \\
    Speech Commands \cite{warden2018speech} &  2018  &  audio  & 100K   &   35  \\
    \bottomrule
  \end{tabular}
  }
\vspace{-5mm}
\end{wraptable}

Many datasets have been created to train AIGC methods for various data modalities, and these datasets play a crucial role in the generation process.
We summarize a list of the representative single-modality and cross-modality datasets in Table \ref{table:datasets} and Table \ref{table:datasets_crossmodal} respectively. Note that, in both tables, the datasets are sorted according to their modality (or modalities), followed by the year they were introduced.~~~~~~~~~~~~
\hypersetup{linkcolor=white}
\begin{NoHyper}
\textcolor{white}{\footnote[1]{https://commoncrawl.org/}}
\textcolor{white}{\footnote[2]{https://dumps.wikimedia.org/}}
\textcolor{white}{\footnote[3]{http://mocap.cs.cmu.edu/}}
\end{NoHyper}
\hypersetup{linkcolor=black}

From the list of representative datasets shown in Table \ref{table:datasets} and Table \ref{table:datasets_crossmodal}, we can observe a few trends, which we discuss here.
Firstly, we can clearly observe that the size of datasets have quickly grown over time.
For image datasets, the scale of datasets has grown from tens of thousands of samples (e.g., Cats \cite{zhang2008cat} and CIFAR-10 \cite{Krizhevsky09cifar10} in the years 2008 and 2009) to hundreds of millions (e.g., JFT-300M \cite{sun2017revisiting}) about 10 years later.
At the same time, the size of text datasets have also increased tremendously, from datasets that are dedicated to specific text-based tasks such as SNLI \cite{bowman2015large} and LAMBADA \cite{paperno2016lambada} which have up to hundreds of thousands of labelled samples, to large web-scale datasets that are not labelled for any particular task, e.g., the whole English Wikipedia\textsuperscript{2} (with a size of 20 GB on disk), or the C4 dataset \cite{raffel2020exploring}, which is a cleaned version of the Common Crawl database \cite{touvron2023llama} that attempted to crawl the whole web.
Furthermore, the quality and resolution of the datasets have mostly improved over time as well. 
The image resolution of CIFAR-10 dataset \cite{Krizhevsky09cifar10} was only $32 \times 32$, but the image resolution of more recent datasets such as AFHQ \cite{choi2020starganv2} and FFHQ \cite{karras2019style} are much larger, at $512 \times 512$ and $1024 \times 1024$ respectively.
The improved quality of data over time also extends to the data diversity and labels.
For instance, improving upon the MS-COCO dataset \cite{chen2015microsoft} of image-text pairs, the more recent Conceptual 12M dataset \cite{changpinyo2021conceptual} contains a higher diversity of captured concepts (i.e., longer-tail distribution), as well as a much longer average caption length.
Moreover, besides collecting image-text pairs, the LN-COCO dataset \cite{pont2020connecting} also includes the data of the annotators hovering their mouse over the region that they are describing, further adding locality information to the text descriptions.
Overall, these observations indicate the significant improvements in size and quality of the datasets over recent years.

\noindent
\textbf{Limitations of Existing Datasets.}
Yet, despite significant improvements in size and quality of datasets, we also observe that there are certain limitations of existing datasets.
First, we observe that, although the datasets for several modalities (e.g., image, video and text) have already gotten quite large, certain other modalities (e.g., 3D humans, 3D motions, and audio) still lack a large unifying and common dataset that can be used for large-scale training of AIGC models for those modalities. This imbalance suggests that there is room for more works
\begin{wraptable}{r}{0.51\textwidth}
  \tiny
  \centering
  \caption{\tiny
  Some representative datasets for AIGC throughout various cross-modality settings.   
  We report the total number of samples in the full dataset, as well as the number of classes (if available) or subjects.
  Moreover, if the number of annotations for each modality is different, we list the number of annotations for each modality (separated by a comma).
  }
  \vspace{-0.3cm}
  \label{table:datasets_crossmodal}
  \scalebox{0.95}{
  \hspace{-0.4cm}
  \begin{tabular}{lcccccccc}
    \toprule
    Dataset & Year  & Modalities  & \#Sample & \#Class     \\
    \midrule
    MS-COCO \cite{lin2014microsoft,chen2015microsoft} & 2014   & image,text   & 123K,616K   &  80  \\
    CUB \cite{wah2011caltech,reed2016learningdeep}  &  2016  & image,text   &  11K,110K  &   200  \\
    Oxford-102 \cite{nilsback2008automated,reed2016learningdeep} & 2016   & image,text   &  8K,80K  &  102   \\
    Conceptual Captions \cite{sharma2018conceptual} &  2018  &  image,text   &  3.3M  &  -   \\
    LN-COCO \cite{pont2020connecting} &  2020  &  image,text  &  142K  & -    \\
    LAION-400M \cite{schuhmann2021laion} & 2021   &  image,text  &  400M  &  -   \\   
    Conceptual 12M \cite{changpinyo2021conceptual} &  2021  &  image,text  &  12.4M   & -    \\
    MSVD \cite{chen2011collecting} & 2011   &  video,text  &  2K,85K  & -     \\
    MSR-VTT \cite{xu2016msr}  &  2016  &  video,text  &  10K,200K  & -     \\
    KTH \cite{schuldt2004recognizing,marwah2017attentive} &  2017  &  video,text  &  2391  &  6   \\
    ANet Captions \cite{krishna2017dense} &  2017  & video,text   &  100K  &  -   \\
    HowTo100M \cite{miech2019howto100m} & 2019   &  video,text  & 136M   &  -   \\
    WebVid-10M \cite{bain2021frozen} &  2021  &  video,text  & 10.7M   &  -   \\
    PASCAL3D+ \cite{xiang2014beyond}  &  2014  &  3D shape,image   &  36K  &   12  \\
    ShapeNet (3D-R2N2) \cite{choy20163d} &  2016  & 3D shape,image   &  51K  & 13     \\
    Pix3D \cite{sun2018pix3d} &  2018  &  3D shape,image  &  10K,395   &  -   \\
    CO3Dv2 \cite{reizenstein2021common} & 2021  &  3D shape,video  &  19K   &  50   \\
    Text2Shape \cite{chen2019text2shape} & 2018 & 3D shape,text & 15K,75K  &  - \\
    Text2Shape++ \cite{fu2022shapecrafter} & 2022 &  3D shape,text & 369K  &  - \\
    Matterport3D \cite{chang2017matterport3d} &  2017  &  3D scene,image  &  90  &   40  \\
    ScanNet v2 \cite{dai2017scannet} & 2017   &  3D scene,video  &  1513  & 20    \\
    BU-3DFE \cite{yin20063d} &   2006 &  3D face,image  & 2.5K   &   100 subjects  \\
    FaceWarehouse \cite{cao2013facewarehouse} & 2013   &  3D face,image   &  150  &   47 facial expressions   \\
    VoxCeleb \cite{nagrani2017voxceleb} &  2017  &  face video,audio  &  153K  &  1251 subjects   \\
    VoxCeleb2 \cite{chung2018voxceleb2} &  2018  &  face video,audio  & 1.1M   &   6K  \\
    DeepHuman \cite{zheng2019deephuman} & 2019   & 3D human,image   &  7K   &  -   \\
    People-Snapshot \cite{alldieck2018video} &  2018  & 3D human,video   &  24 sequences  &   11 subjects  \\
    3DPW \cite{von2018recovering} & 2018   &   3D human,video &  51K frames  &   - \\
    MPI-INF-3DHP \cite{mehta2017monocular}  &  2017  &  3D human,video  & 1.3M frames    &  8 activities   \\
    TED Gesture \cite{yoon2020speech} &  2020  &  3D human,video,audio,text  & 1766 videos   &  -   \\ 
    DeepFashion-MultiModal \cite{jiang2022text2human}  & 2022  & human images,text & 11K & 24     \\      
    KIT \cite{plappert2016kit} &  2016  & 3D motion,text   &  3K,6K  &  -   \\
    HumanML3D \cite{guo2022generating} &  2022  &  3D motion,text  &  14K,44K  & -    \\
    Human3.6M \cite{ionescu2013human3} &  2013  &  3D motion,video  &  3.6M frames  &   15  \\
    NTU RGB+D \cite{shahroudy2016ntu} & 2016   &  3D motion,video  &  56K  &  60   \\
    NTU RGB+D 120  \cite{liu2019ntu}  &  2019  &  3D motion,video  &  114K  &  120   \\
    AIST++ \cite{li2021ai} &   2021 &  3D motion,music  &  1408 sequences  &  -   \\
    LJ Speech \cite{ito2017lj} &  2017  &  audio,text  &  13K  &   -  \\
    Audio Set \cite{gemmeke2017audio} & 2017 & audio,text &  1.7M & 632  \\     
    LibriTTS \cite{zen2019libritts} &  2019  & audio,text   &  585 hours  &  2456 speakers   \\
    GRID \cite{cooke2006audio} &  2006  &  audio,video  & 34K sentences   &   34 subjects  \\
    LRW \cite{chung2017lip} &  2017  & audio,video   &  1M words  &  1K subjects   \\
    Music Caps \cite{agostinelli2023musiclm} & 2023 & music,text & 5.5K  & -  \\
    \bottomrule
  \end{tabular}
  }
\vspace{-16mm}
\end{wraptable}
\noindent
 related to curating larger benchmarks for these under-explored modalities.
Notably, there are challenges involved in curating data for these modalities, such as the high cost of collecting high-quality 3D data, or the various privacy issues with collecting audio data, which may be some of the underlying issues behind the limited size of these datasets.
Furthermore, there is also a notable lack of standardized editing-based datasets/benchmarks.
In the earlier years of AIGC, many image-based editing works \cite{zhu2017unpaired,lee2018diverse} focused on style transfer, and these works generally experimented with common benchmarks such as the Summer $\leftrightarrow$ Winter dataset \cite{zhu2017unpaired} or the Photo $\leftrightarrow$ Art dataset \cite{zhu2017unpaired}.
However, the settings of recent works have become more widely varied, spanning over text-based editing, and controllable editing methods with various types of input information (e.g., clicks \cite{pan2023draggan}, desired objects \cite{li2023gligen}, desired movements \cite{hao2018controllable}).
These aspects lead to a difficulty in curating common benchmarks, especially in controllable editing, where the settings can widely vary.
Lastly, it can also often be difficult to find large datasets containing more than two modalities. 
This may hinder efforts for large-scale training of multi-modal AIGC models.

\section{Applications}
\label{sec:applications}
In this section, we discuss some of the typical application domains and practical application scenarios of AIGC models.
In line with the theme of our survey, we categorize the applications of the generative models according to the various modalities.

\subsection{Text Modality}
\noindent
\textbf{Chatbots.}
Text-based AI generative models are crucial for powerful conversational chatbots such as ChatGPT that can handle diverse tasks, ranging from customer service to personal assistants  \cite{kalla2023study}. 
Importantly, these text generative models can help to parse user queries and generate the relevant responses, which can enhance user experience, reduce the manpower costs, and allow for 24/7 customer support. 
These chatbots can also facilitate translation between multiple languages, and could potentially even translate sign languages as well \cite{gong2024llms}.

\noindent
\textbf{Education.}
Text-based AIGC models also have important applications in the education domain, through providing personalized learning experiences and enhancing educational tools \cite{baidoo2023education}.  
These text models can potentially automatically generate tailored content, such as quizzes, essays, and study materials, adapted to the needs and proficiency levels of individual students, which greatly facilitates learning.
They can also assist teachers by automating administrative tasks such as report generation and assignment grading, freeing up more time for teaching.

\noindent
\textbf{Gaming.}
In the gaming industry, text-based AIGC methods can also play a critical role in creating dynamic, immersive narratives and expanding gameplay possibilities. 
For instance, these models can be used to facilitate the generation of dialogues, storylines, and quest content, allowing developers to conveniently build more complex and interactive game worlds \cite{lanzi2023chatgpt}. 
In particular, text-based games benefit greatly from the ability of AI models to dynamically generate rich, adaptive narratives that evolve based on player actions, enhancing the depth, realism and immersion of the game.

\noindent
\textbf{Code Development.}
There is also a huge potential for text-based AIGC methods in the realm of code development \cite{chen2021evaluating}, where code generation tools can massively impact how software is written, debugged, and maintained. 
For instance, given a user-input text prompt, these models can automatically generate code snippets, functions, or detect potential errors, which can improve developer productivity and accelerate the software development process.

\subsection{Image and Video Modality}

\noindent
\textbf{Media and Entertainment.}
Image and video generative models are quickly revolutionizing the media and entertainment industry by enabling the rapid creation of visual and video content \cite{park2019generating}, greatly reducing the costs of content creation.
Specifically, these image and video models can greatly help with many aspects of media creation, including creating environments, adding textures or special effects, or creating interactive content.
Overall, the ability of these models to produce high-quality visuals quickly and efficiently is reshaping content creation across entertainment platforms.

\noindent
\textbf{Art.}
In the world of art, image and video generative models enable artists to explore new forms of expression.
In particular, artists may experiment with using algorithms to generate unique patterns, styles, and compositions, blending human creativity with AI capabilities to produce innovative works of digital art.
For example, artists may experiment with creating visual anagrams \cite{geng2024visualanagrams}, where images can contain multi-view optical illusions.

\noindent
\textbf{Advertising.}
In advertising, image and video-based AI generative models can play a key role in creating personalized and engaging visual content tailored to individual consumers \cite{guo2021vinci}, generating custom images or videos based on each consumer's preferences and behaviour.
At the same time, these AIGC methods can also enable the rapid creation of advertisements and marketing materials.
The ability to deliver highly targeted campaigns for specific demographics, and to quickly generate creative visuals at scale, make such AIGC methods greatly valuable for advertising, increasing the efficiency and efficacy of marketing strategies.

\noindent
\textbf{Film Production.}
Moreover, image and video generation models also have great potential in the film industry, through streamlining visual effects (VFX), animation, and even post-production editing processes \cite{singh2023artificial}.
Specifically, they can be used to generate realistic environments and virtual characters, reducing the need for expensive physical sets and manual animation work.
Furthermore, these generative models can be used for tasks such as de-aging actors, creating digital doubles, or enhancing scenes with effects like weather or lighting changes, which can aid filmmakers with create films which may be otherwise unfeasible (e.g., de-aging a famous actor for scenes involving a young character).
Overall, AIGC is not only speeding up film production by automating these labor-intensive processes, but also expanding creative possibilities for directors and artists.

\subsection{3D Modality}

\noindent
\textbf{VR/AR Applications.}
3D AIGC methods are crucial for creating immersive 3D environments for AR/VR applications, including the metaverse \cite{ratican2023proposed}.
These 3D generative models enable the rapid and automated generation of lifelike virtual worlds, avatars, and interactive objects, essential for building scalable and dynamic virtual environments. 
In the metaverse, generative AI can create personalized spaces and generate objects in real-time, enabling adaptive environments that respond to user behavior, enhancing immersion and user engagement.

\noindent
\textbf{E-commerce and Fashion.}
In e-commerce and fashion, 3D generative models allow for the creation of virtual try-on experiences, providing consumers with highly personalized shopping experiences.
For instance, by generating 3D models of clothing, accessories, and even virtual avatars that resemble customers \cite{gong2024laga}, 3D AIGC methods can simulate how products will fit and look in real life. 
These 3D generative models also enable fashion brands to create and showcase digital-only garments, opening up new markets for virtual fashion within the metaverse and online platforms.

\noindent
\textbf{Architecture and Interior Design.}
Furthermore, 3D generative models streamline the design process for for architecture and interior design, by automating the creation of complex structures and interior layouts \cite{gao2023scenehgn}. 
These models can quickly generate multiple design variations, offering architects and designers new ways to explore creative possibilities, and optimize spaces for functionality and aesthetics.

\subsection{Audio Modality}

\noindent
\textbf{Music Production.}
Audio generative models can facilitate musicians and composers to rapidly generate music content \cite{dhariwal2020jukebox}. These models can create original compositions, suggest harmonies or melodies, and even mimic the styles of specific genres or artists, drastically reducing the time and effort required for music creation.

\noindent
\textbf{Voice Synthesis and Speech Generation.}
In voice synthesis and speech generation, audio-based AIGC methods can potentially revolutionize the industry by creating natural-sounding, human-like voices for a wide range of applications \cite{borsos2023audiolm}. 
For example, AI-generated voices can provide accessibility solutions for people with speech impairments.
Furthermore, these audio generative models can generate expressive speech for applications ranging from virtual assistants and customer service bots to narration in videos or audiobooks.

\section{Challenges}

Despite the rapid progress of AIGC methods, there are still many challenges faced by researchers and users.
We explore some of these issues and challenges below.

\noindent
\textbf{Deepfakes.}
AIGC has the potential to be misused to create deepfakes, e.g., synthetically generated videos that look like real videos, which can be used to spread disinformation and fake news \cite{vaccari2020deepfakes}.
With the significant improvements in AIGC, where the generated outputs are of high quality and fidelity, it is getting more difficult to detect these synthetic media (e.g., images and videos) by the naked eye.
Thus, several algorithms have been proposed to detect them, especially for GAN-based CNN generators \cite{chandrasegaran2021closer,chandrasegaran2022discovering}.
With the rapid development of more recent architectures (e.g., Transformers) and techniques (e.g., DMs), how to detect emerging deepfakes deserves further investigation.

\noindent
\textbf{Adversarial Attacks.}
AIGC methods can also be vulnerable to adversarial attacks.
For instance, AIGC methods can be vulnerable to backdoor attacks, where the generated output can be manipulated by a malicious input trigger signal.
Specifically, these backdoor attacks can generate a pre-defined target image or an image from a specified class, when the trigger pattern is observed. 
Adversaries can insert such backdoors into models by training them on poisoned data before making them publicly available, where users might unknowingly use these backdoored models in their applications, which can be risky.
Some works investigate such backdoor attacks on GANs \cite{salem2020baaan} and DMs \cite{chou2023backdoor}.
Some defensive measures have been proposed \cite{wu2021adversarial}, but there still exists room for improvement in terms of defense efficacy while maintaining good generative performance.

\noindent
\textbf{Privacy Issues.}
There are also privacy issues with AIGC methods, since many models (e.g., chatbots such as ChatGPT) often handle sensitive user data, and such sensitive data can be collected to further train the model.
Such training data can potentially be memorized by large models \cite{carlini2023quantifying} which can be leaked by the model \cite{carlini2021extracting,carlini2023extracting}, causing privacy concerns.
How to mitigate such privacy issues in AIGC largely remains an open problem.

\noindent
\textbf{Legal Issues.}
There are also potential legal issues with AIGC, especially with regards to the intellectual property rights of generated outputs.
For instance, recent text-to-image DMs \cite{ramesh2022hierarchical,rombach2022high} are often trained on extremely large datasets with billions of text-caption pairs that are obtained from the web.
Due to the size of the dataset, it is often difficult to carefully curate and check for the intellectual property rights of each image.
Furthermore, due to the memorization ability of large models \cite{carlini2023extracting,carlini2023quantifying}, they might directly replicate images from the training set, which might leads to copyright infringement.
To overcome this, some works \cite{ramesh2022hierarchical} have pre-training mitigations \cite{nichol2022dalle} such as deduplicating the training data, while some works investigate training factors to encourage diffusion models to generate novel images \cite{somepalli2023diffusion}.
Further investigations in this direction are required for more effective and efficient ways of tackling this issue.

\noindent
\textbf{Domain-specialized Applications.}
A potential challenge for AIGC is their application to specialized domains which might have less data, such as medical areas \cite{nori2023capabilities}, or laws \cite{xiao2021lawformer}.
When using large language models such as ChatGPT, they also can hallucinate make factual errors \cite{ji2023survey}, which can be problematic in these specialized domains. 
A promising direction is to find ways to apply domain-specific knowledge to reduce and mitigate these errors when large language models are applied to specific domains \cite{agrawal2022large}, or to enhance the factuality of the generated text using a text corpus \cite{lee2022factuality}.
We further note that LLMs have also been successfully applied in non-generative tasks such as segmentation \cite{zhu2024llafs}, action recognition \cite{qu2024llms} and long-tail recognition \cite{zhao2024ltgc}, showing their versatility even in these scenarios.

\noindent
\textbf{Multi-modal Combinations.}
Generative models with multi-modal inputs are very useful as they can gain understanding of the context through different modalities, and can better capture the complexity of the real world.
Such multi-modal models are important for many practical applications such as robotics \cite{driess2023palme} and dialogue systems \cite{OpenAI2023GPT4}.
However, when we aim to have several modalities as input, it can be difficult to curate large-scale datasets with all modalities provided for each sample, which makes training such models challenging.
To overcome this, one line of works \cite{tumanyan2023plug,nair2023unite} allow for multiple input modalities by collaborating multiple pre-trained DMs, each with different input modalities.
Further advancements in this domain are necessary to devise more efficient strategies for attaining multi-modality.

\noindent
\textbf{Updating of Knowledge.}
As human knowledge, pop cultures, and word usage keep evolving, large pre-trained AIGC models should also update their knowledge accordingly.
Importantly, such updating of knowledge should be done in an efficient and effective manner, without requiring a total re-training from scratch with all the previous and updated data, which can be very costly.
To learn new information, some works perform continual learning \cite{zhai2019lifelong}, where models can learn new information in the absence of the previous training data, without suffering from catastrophic forgetting.
Another direction of model editing \cite{mitchell2022fast,meng2022locating} aims to inject updated knowledge and erase incorrect knowledge or undesirable behaviours, e.g., editing factual associations in language models \cite{meng2022locating}.
However, as pointed out in \cite{mitchell2022fast}, existing methods are limited in their enforcing of the locality of edits, and also do not assess the edits in terms of their generality (i.e., editing of indirect associations and implications as well).
Thus, the are many opportunities for further investigations into this area.

\begin{figure*}[t]
    \center
    \includegraphics[width=\textwidth]{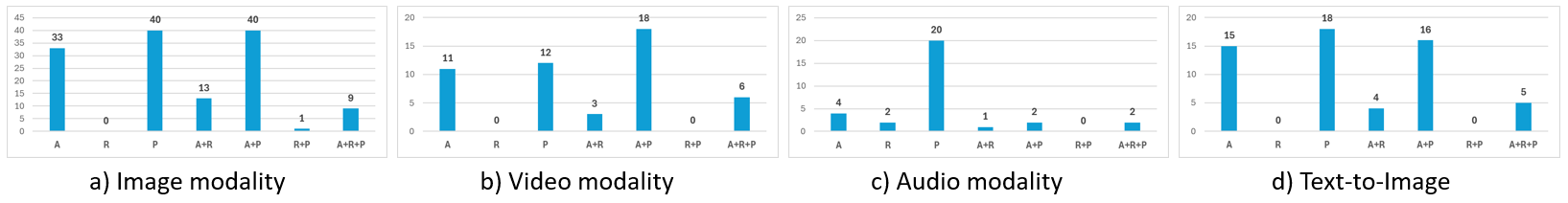}
    \vspace{-5mm}
    \caption{
    Affiliated organizations of the surveyed papers.
    Here, we show the number of papers published by each type of organization:
    Academia (A), Research Institution (R), Private Company (P). If multiple types of organizations were affiliated with the paper, we indicate the collaboration with a plus sign (+), e.g., if a paper is affiliated with both academia and a private company, it is in the 'A+P' category.
    The subplots each show the tally for the number of papers in one field:
    (a) image modality; (b) video modality; (c) audio modality; (d) text-to-image.
    }
    \label{figure:affilation_counts}
    \vspace{-5mm}
\end{figure*}

\section{Discussion}
\label{sec:papers_discussion}

In this survey, we seek to find and include the most representative works of each modality and each sub-field. To do this, we select papers considering the following aspects: 1) We searched through all the papers published at top conferences (e.g., CVPR, ICCV, ECCV, NeurIPS, ICLR, ICML) and journals (e.g., TPAMI, ToG) of each sub-field to find representative fields and works. 2) For works that are not in these venues, we also selected highly cited works. 3) For each representative technical line, we cite the pioneering works, and we also cite milestone ones. By using this methodology, we find that selected works are generally representative papers in the field that represent notable advancements, which have remained influential over time and have laid the foundations for recent advancements and emerging directions.
In this section, we collate some information regarding the organizations affiliated with these representative papers, and discuss some observations.

Specifically, for several selected modalities, we tallied the number of papers affiliated with academia (A) vs private companies (P) vs research institutions (R), also noting down the collaborations between these types of organizations.
These analyses have been visualized in Figure \ref{figure:affilation_counts}.
From these analyses, we observe a trend that academic institutions, private companies, as well as collaborations between academic institutions and private companies make up a large proportion of the representative papers selected for this survey.
For all the plots in Figure \ref{figure:affilation_counts}, about 90\% of the papers involve at least an academic institution, a private company, or even a collaboration between the two.

Furthermore, we observe a significant anomaly in the audio modality, where the number of papers from academia is abnormally low, 
For the image, video modalities and the text-to-image setting, approximately 70\% of the representative works in these fields involved academic institutions.
Yet, for the audio modality, only 29\% of the surveyed works involved academic institutions, and even fewer papers involved only academic institutions (13\%).
This disparity may be due to the lack of large-scale open-source audio datasets (e.g., large music datasets), which may pose bigger issues for academic researchers.
This may also explain why the number of audio generation papers (in Section \ref{sec:audio}) is lower than the other modalities.

\section{Conclusion}

AIGC is an important topic that has garnered significant research attention across diverse modalities, each presenting different characteristics and challenges. 
In this survey, we have comprehensively reviewed AIGC methods across different data modalities, including single-modality and cross-modality methods.
Moreover, we have organized these methods based on the conditioning input information, affording a structured and standardized overview of the landscape in terms of input modalities. 
We highlight the various challenges, representative works, and recent technical directions in each setting. 
Furthermore, we make performance comparisons between representative works, and also review the representative datasets and benchmarks throughout the modalities.
We also discuss the challenges and potential future research directions.

\renewcommand*{\bibfont}{\small}


\end{document}